\documentclass[letterpaper,journal]{IEEEtran}
\usepackage[T1]{fontenc}
\usepackage{amsmath,amsfonts}
\usepackage{algorithmic}
\usepackage{algorithm}
\usepackage{array}
\usepackage[caption=false,font=normalsize,labelfont=sf,textfont=sf]{subfig}
\usepackage{textcomp}
\usepackage{stfloats}
\usepackage{url}
\usepackage{verbatim}
\usepackage{graphicx}
\usepackage{multirow}
\usepackage{cite}
\usepackage{amssymb}
\usepackage{xcolor}
\usepackage{makecell}
\usepackage{tabularx}
\usepackage{needspace}
\newcolumntype{Y}{>{\centering\arraybackslash}X}
\definecolor{ao}{rgb}{0.0,0.5,0.0}
\newcommand{\hcell}[2]{\begin{tabular}[t]{@{}c@{}}#1\\#2\end{tabular}}
\providecommand{\xmark}{\ensuremath{\times}}
\newcommand{\greencheck}{{\color{ao}\ensuremath{\checkmark}}}
\newcommand{\redcross}{{\color{red}\ensuremath{\xmark}}}
\begin{document}
\title{GAAT: Geometry-Aware Alignment Transformer for Multimodal UAV Perception}

\author{
Jingpu Yang$^{1,2}$,~\IEEEmembership{Student Member, IEEE},
Debin Tang$^{3}$,
Yilin Sun$^{1}$,
Fengxian Ji$^{2}$,
Jiahua Zhu$^{1}$,
Wenrui Ding$^{1}$,
and Yufeng Wang$^{1}$,~\IEEEmembership{Member,~IEEE}
\thanks{Jingpu Yang, Debin Tang, and Yilin Sun are co-first authors.}
\thanks{$^{1}$Beihang University, Beijing, China.}
\thanks{$^{2}$Zhongguancun Academy, Beijing, China.}
\thanks{$^{3}$Northeastern University, Shenyang, China.}
\thanks{E-mail: Jingpu Yang: jingpuyang@buaa.edu.cn; Debin Tang:
lihuaijiao@gmail.com; Yilin Sun: simon.sun@buaa.edu.cn; Fengxian Ji:
jifengxian1224@gmail.com; Jiahua Zhu: zhujiahua23@mails.ucas.ac.cn;
Wenrui Ding: ding@buaa.edu.cn; Yufeng Wang: wyfeng@buaa.edu.cn.}
\thanks{Yufeng Wang is the corresponding author.}
}

\maketitle

\begin{abstract}
Unmanned aerial vehicle (UAV) multimodal perception integrates heterogeneous sensors---visible (RGB), infrared (IR), synthetic aperture radar (SAR), or depth---to support scene understanding under diverse conditions. However, UAV-mounted sensors differ in optics, resolution, and mounting, so practical systems often provide only global or image-center alignment. Once such pairs are tokenized into patches, parallax, platform motion, and lens distortion can shift corresponding patch centers across modalities, weakening the spatial correspondence assumption underlying dense contrastive learning and cross-modal fusion. We propose GAAT (Geometry-Aware Alignment Transformer), an alignment-first multimodal pretraining framework that estimates local reliability before the modalities interact. GAAT introduces syncPATC, which learns transformation-consistent patch-center reliability under synchronized view transformations without manually annotated correspondences. It uses token and query confidence to rank trustworthy local anchors, together with query centers and sub-token offsets to guide local sampling in the presence of residual misalignment. Guided by these priors, MG-Sparse-MMA performs query-mediated sparse fusion over top-$K_s$ reliable regions, replacing dense all-patch interaction with geometry-calibrated local updates. RA-QCGCL aligns the pretraining supervision with this sparse query bottleneck through reliable patch-to-patch, patch-to-query, and query-to-query contrastive branches. We also introduce UAVMeta for paired RGB--IR pretraining and in-domain multi-task evaluation, together with StateBench for paired-input prediction of acquisition-state proxy scores derived from platform telemetry and image statistics. Across six primary downstream tasks and multiple UAV datasets, GAAT demonstrates strong multi-task transfer, attains state-of-the-art results on multiple benchmarks, and remains competitive in task-specific settings. It also achieves the highest reported aggregate MSPA-4D score on StateBench.
\end{abstract}

\begin{IEEEkeywords}
  UAV Perception, Multimodal Pretrained Model, Patch-Center Alignment, Geometry-Aware Alignment, UAVMeta Dataset, StateBench.
\end{IEEEkeywords}

\section{Introduction}
\label{sec:intro}

\IEEEPARstart{U}{nmanned} aerial vehicle (UAV) multimodal perception, in which heterogeneous on-board sensors observe the same scene from a moving aerial platform, supports applications ranging from urban scene understanding to disaster response, search-and-rescue, and night-time surveillance. UAV payloads often combine RGB cameras with thermal infrared (IR) cameras, multispectral sensors, SAR, depth sensors, or LiDAR, whose signals are complementary: optical channels carry fine texture and color, whereas non-optical channels remain informative when visible information is degraded by illumination, weather, or occlusion \cite{ying2025visiblethermal,zhu2025wavemamba,ouyang2025kust4k}. This motivates multimodal pretraining that accounts for the geometry of aerial capture. Here, we develop and evaluate the Geometry-Aware Alignment Transformer (GAAT) and UAVMeta in the paired RGB--IR setting.

Unlike satellite or stationary multi-sensor systems, UAV-mounted heterogeneous cameras typically differ in optics, resolution, field of view, exposure, mounting, and acquisition time \cite{suo2023hituav,jia2021llvip,nie2025m3ot}. Pixel-level alignment of the modality streams is therefore often infeasible at scale; in our setting, the data pipeline typically provides coarse global alignment, usually around the image center, after on-board cropping, resizing, or coarse warping. The cost of this practical compromise becomes evident in transformer-style pretraining \cite{dosovitskiy2021vit,liu2021swin,liu2022swinv2}. After paired images are split into patches, same-index tokens can still cover different physical regions across modalities. Parallax, oblique viewpoints, lens distortion, and platform motion coupled with acquisition-time offsets can make this residual misregistration large enough to invalidate same-index correspondence in some scenes.

\begin{figure*}[!t]
  \centering
  \includegraphics[width=\textwidth]{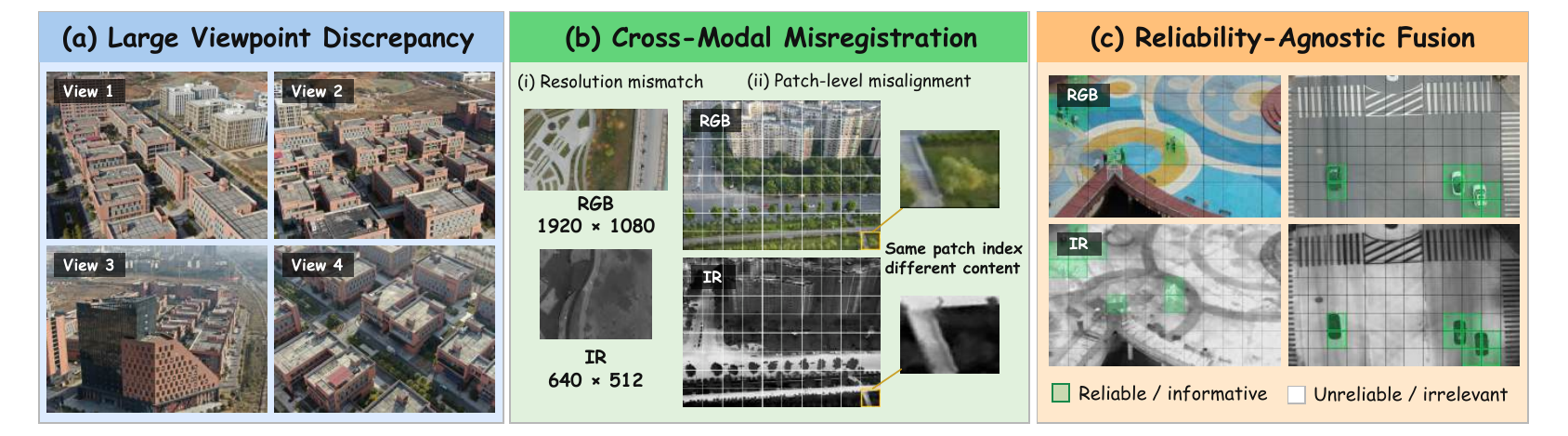}
  \caption{Geometry and reliability challenges in multimodal UAV perception: (a) viewpoint variation; (b) cross-modal misregistration from resolution mismatch and patch displacement; and (c) reliability-agnostic fusion of reliable or informative (green) and unreliable or irrelevant (white) regions.}
  \label{fig:uav-challenges}
\end{figure*}

This residual patch-center misalignment creates three coupled challenges. \emph{First}, without dense correspondence annotations, the model must estimate which local regions remain reliable enough to interact: same-index tokens may depict different physical regions, and using them as positives can corrupt contrastive learning \cite{chen2020simclr,he2020moco,caron2021dino}. \emph{Second}, cross-modal fusion must concentrate on reliable, informative neighborhoods. Dense all-patch interaction otherwise mixes mismatched content and allocates computation to the many low-information patches common in UAV frames. \emph{Third}, pretraining supervision must match the granularity of the resulting sparse interaction: global objectives overlook the selected patch and query relations, whereas indiscriminate same-index supervision can reinforce unreliable positives. Fig.~\ref{fig:uav-challenges}(a) and (b) illustrate the geometric variation and local displacement underlying the first challenge, while Fig.~\ref{fig:uav-challenges}(c) motivates reliability-aware fusion; together, these failure modes motivate supervision targeted at the relations actually used for interaction.

\begin{figure*}[!t]
  \centering
  \includegraphics[width=\textwidth]{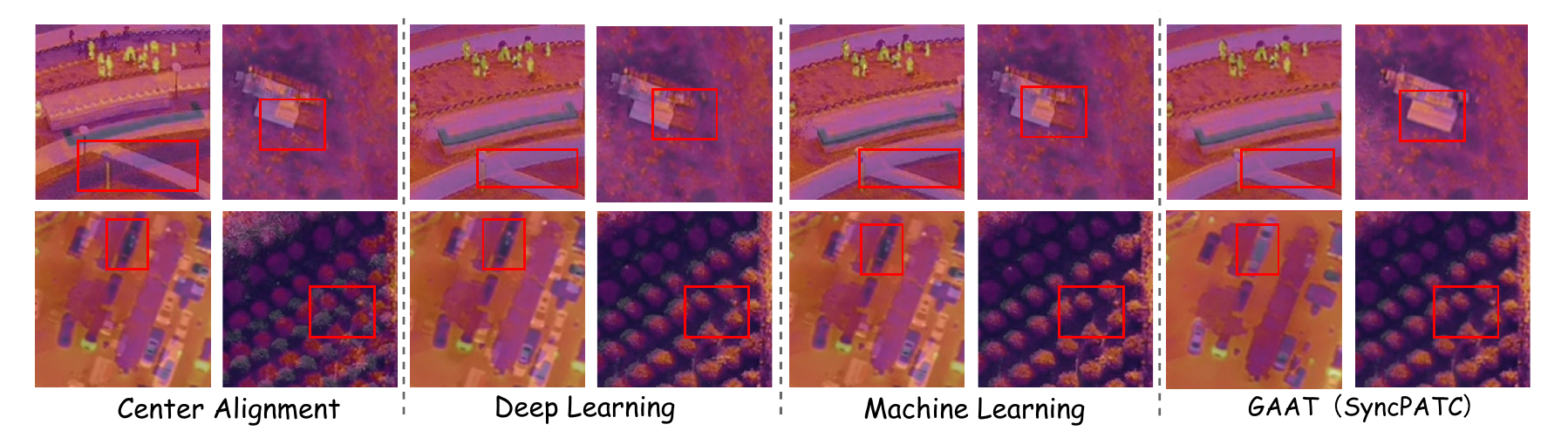}
  \caption{Qualitative illustration of local-region localization under coarsely aligned RGB--IR inputs. Red boxes indicate candidate regions produced by the four strategies shown; this figure is illustrative and is not a quantitative registration benchmark.}
  \label{fig:alignment-comparison}
\end{figure*}

GAAT addresses these challenges through a shared reliability pathway rather than three independent additions. \emph{First}, Synchronized Patch-Center Alignment (syncPATC) learns transformation-consistent reliability priors from synchronized synthetic view transformations, without dense correspondence labels or explicit regression of an RGB--IR registration warp. It produces token and query confidence for ranking reliable anchors, together with query centers and sub-token offsets for locating and refining local sampling. \emph{Second}, Modality-Guided Sparse Multimodal Attention (MG-Sparse-MMA) uses these priors to route bidirectional fusion through reliability-ranked queries and geometry-calibrated local neighborhoods. \emph{Third}, Reliability-Aware Query-Guided Cross-Granularity Contrastive Learning (RA-QCGCL) uses the same priors and queries to supervise patch-to-patch, patch-to-query, and query-to-query relations. The alignment assumption, fusion mechanism, and learning objective are therefore coupled through one reliability signal. Fig.~\ref{fig:alignment-comparison} qualitatively illustrates local-region localization before fusion; it is not a quantitative registration evaluation.

Table~\ref{tab:rsfm-task-coverage} situates GAAT by sensing platform, modality setting, and reported task coverage; it is a scope comparison rather than an alignment-performance ranking. Representative Earth-observation foundation models establish broad transfer across satellite imagery and heterogeneous observations \cite{szwarcman2025prithvi,cong2022satmae,reed2023scalemae,sun2023ringmo,bastani2023satlas,fuller2023croma,xiong2024dofa,astruc2024omnisat,astruc2025anysat,guo2024skysense,zhang2025skysensev2,wu2025skysensepp,wang2024mtp,wang2025ringmogalaxy}, while RingMo-Aerial extends foundation-model transfer to single-modal aerial imagery \cite{diao2025ringmoaerial}. GAAT has a complementary focus: residual multimodal misalignment in coarsely aligned paired UAV streams. Registration-first pipelines estimate an explicit image-space transform; GAAT instead estimates transformation-consistent interaction reliability and reuses it to control both sparse fusion and cross-granularity supervision.

\begin{table*}[!t]
\caption{Scope comparison of representative remote-sensing foundation models by sensing platform, multimodal setting, and reported task coverage. The table reports evaluation breadth; alignment assumptions, supervision signals, and fusion mechanisms are distinguished in Sec.~\ref{sec:related-work}. \greencheck~and \redcross~denote the capabilities reported for the corresponding model and column, respectively.}
\label{tab:rsfm-task-coverage}
\centering
\footnotesize
\setlength{\tabcolsep}{3.4pt}
\renewcommand{\arraystretch}{1.12}
\begin{tabular}{@{}l l c c c c c c c@{}}
\hline
\textbf{RS Type} 
& \textbf{Model} 
& \makecell{\textbf{Multi-}\\\textbf{modal}} 
& \makecell{\textbf{Semantic}\\\textbf{Segmentation}} 
& \makecell{\textbf{Object}\\\textbf{Detection}} 
& \makecell{\textbf{Multi-Object}\\\textbf{Tracking}} 
& \makecell{\textbf{Scene}\\\textbf{Classification}} 
& \makecell{\textbf{Change}\\\textbf{Detection}} 
& \makecell{\textbf{3D}\\\textbf{Reconstruction}} \\
\hline
\multirow{11}{*}{Satellite / EO}
& Prithvi-EO              & \redcross   & \greencheck & \redcross   & \redcross   & \greencheck & \redcross   & \redcross \\
& SatMAE / SatMAE++       & \redcross   & \greencheck & \redcross   & \redcross   & \greencheck & \redcross   & \redcross \\
& Scale-MAE               & \redcross   & \greencheck & \redcross   & \redcross   & \greencheck & \redcross   & \redcross \\
& RingMo                  & \redcross   & \greencheck & \greencheck & \redcross   & \greencheck & \redcross   & \redcross \\
& SatlasPretrain          & \greencheck & \greencheck & \greencheck & \redcross   & \greencheck & \greencheck & \redcross \\
& CROMA                   & \greencheck & \greencheck & \redcross   & \redcross   & \greencheck & \redcross   & \redcross \\
& DOFA                    & \greencheck & \greencheck & \greencheck & \redcross   & \greencheck & \redcross   & \redcross \\
& OmniSat                 & \greencheck & \greencheck & \redcross   & \redcross   & \greencheck & \redcross   & \redcross \\
& AnySat                  & \greencheck & \greencheck & \redcross   & \redcross   & \greencheck & \greencheck & \redcross \\
& SkySense / SkySense V2  & \greencheck & \greencheck & \greencheck & \redcross   & \greencheck & \greencheck & \redcross \\
& MTP                     & \redcross   & \greencheck & \greencheck & \redcross   & \greencheck & \greencheck & \redcross \\
\hline
\multirow{2}{*}{Aerial / UAV}
& RingMo-Aerial           & \redcross   & \greencheck & \greencheck & \greencheck & \greencheck & \greencheck & \greencheck \\
& \textbf{GAAT (Ours)}    & \greencheck & \greencheck & \greencheck & \greencheck & \greencheck & \greencheck & \greencheck \\
\hline
\end{tabular}
\end{table*}

To support model development and evaluation in the same acquisition domain, we introduce UAVMeta and StateBench with distinct roles. UAVMeta supplies paired RGB--IR imagery, multi-task annotations, and acquisition metadata for pretraining and in-domain task evaluation; its schema can additionally accommodate SAR, depth, and multispectral streams. StateBench converts platform telemetry and image statistics into four acquisition-state proxy scores---Camera Acquisition Reliability Score (CARS), Observation Scale and Ground Sampling Score (OSGS), Viewpoint Stability Score (VSS), and Flight Maneuver Complexity Score (FMCS)---and evaluates paired-input state prediction through the MSPA-4D protocol.

Across six primary downstream tasks and multiple UAV datasets, GAAT shows strong multi-task transfer, reaches state-of-the-art performance on multiple benchmarks, and remains competitive in other task-specific settings. On StateBench, it obtains the highest reported aggregate MSPA-4D score.

\Needspace{7\baselineskip}
The main contributions of this work are summarized as follows.
\begin{itemize}
\item We formulate residual \emph{patch-center misalignment} after coarse RGB--IR registration as a reliability problem and introduce \textbf{syncPATC}, which learns transformation-consistent token and query priors from synchronized synthetic views without dense correspondence labels.
\item We introduce \textbf{MG-Sparse-MMA}, which converts these priors into bidirectional, query-mediated fusion over reliability-ranked, geometry-calibrated local neighborhoods instead of dense all-patch interaction.
\item We introduce \textbf{RA-QCGCL}, which matches pretraining supervision to the sparse interaction structure through reliable patch-to-patch, patch-to-query, and query-to-query contrastive relations.
\item We construct \textbf{UAVMeta} for paired RGB--IR pretraining and multi-task evaluation and \textbf{StateBench} for paired-input prediction of acquisition-state proxy scores. Together, these resources support pretraining, in-domain task evaluation, and acquisition-state evaluation in a common paired UAV setting.
\end{itemize}

\section{Related Work}
\label{sec:related-work}
\subsection{UAV-Oriented Aerial Perception and Foundation Models}
UAV imagery has become an important source for fine-grained aerial perception due to its flexible deployment, high spatial resolution, and real-time acquisition capability. Unlike satellite imagery, UAV imagery is usually captured from lower altitudes and more variable trajectories, leading to distinctive visual characteristics such as oblique viewpoints, large-scale variations, dense small objects, platform motion, and frequent occlusion. These properties make UAV perception more challenging than conventional nadir-view remote sensing interpretation, especially for object detection, semantic segmentation, tracking, scene classification, and 3D reconstruction \cite{zhu2021visdrone,du2018uavdt,lyu2020uavid,cheng2017resisc45,long2021millionaid}.

Recent aerial and remote-sensing foundation models adapt masked modeling and contrastive learning to viewpoint, scale, and modality variation \cite{sun2023ringmo,diao2025ringmoaerial,yang2026mars,he2022mae,bachmann2022multimae,chen2020simclr,manas2021seco,fan2025movingdrone,yang2026geometricgating}. RingMo-Aerial introduces affine-transformation contrastive learning for single-modal aerial transfer, whereas MaRS develops cross-granularity meta-modality learning for multimodal very-high-resolution imagery \cite{diao2025ringmoaerial,yang2026mars}. Multimodal satellite-oriented models such as CROMA and SkySense emphasize transferable representations across heterogeneous observations \cite{fuller2023croma,guo2024skysense,zhang2025skysensev2,wu2025skysensepp}. These studies establish view-aware augmentation, multimodal pretraining, and cross-granularity supervision as effective components. GAAT addresses a distinct coupling: residual local misregistration can simultaneously corrupt positive construction and fusion in paired UAV cameras. Its novelty lies in estimating reliability before interaction and using the same signal to control both fusion neighborhoods and patch- and query-level supervision.

UAV platforms often carry multiple sensors, including RGB cameras, IR cameras, multispectral sensors, SAR, depth sensors, and LiDAR. These sensors observe the same scene through different imaging mechanisms and provide complementary cues. However, sensor baselines, view changes, field-of-view mismatch, and acquisition-time offsets under platform motion can cause local spatial displacement. Recent multimodal aerial models and benchmarks illustrate both the value of fusion and the persistence of this alignment problem \cite{zhu2025wavemamba,zhang2023cmx,fuller2023croma,guo2024skysense,wu2025skysensepp,yang2026mars}. Therefore, UAV multimodal pretraining should model local interaction reliability rather than assume that paired images or same-index tokens are already aligned.

\subsection{Multimodal Fusion for UAV Perception}
Multimodal fusion is widely used in UAV perception to integrate complementary information from heterogeneous sensors. Existing fusion methods can be roughly divided into input-level fusion, feature-level fusion, and decision-level fusion. Input-level methods directly concatenate or transform raw modality signals; these methods are simple but highly sensitive to calibration errors, resolution differences, and field-of-view mismatch. Decision-level methods combine predictions from separate modality-specific models; this improves robustness to modality degradation but limits early cross-modal interaction. Feature-level fusion has therefore become the dominant strategy, where modality-specific encoders extract features and cross-modal modules exchange information through concatenation, gating, attention, or transformer-based interaction \cite{ha2017mfnet,wang2022tokenfusion,zhang2023cmx,guo2024skysense,wu2025skysensepp,zhu2025wavemamba,wan2025sigma}.

Recent UAV multimodal fusion methods usually adopt dual-branch networks, cross-attention modules, modality-aware weighting, uncertainty estimation, or missing-modality training. These designs improve perception robustness under illumination changes, night-time scenes, weather variation, and sensor degradation. However, most of them fuse features according to the same spatial index or through dense global attention. This design implicitly assumes that tokens at the same location across modalities correspond to the same physical region. In UAV imagery, this assumption is fragile because parallax, sensor displacement, view-angle changes, and platform motion coupled with timing offsets can shift locally corresponding regions away from each other.

Under local misalignment, dense fusion may introduce semantically inconsistent interactions. For example, a vehicle token in one modality may be fused with a road or background token in another modality. Attention-based fusion can further propagate this problem, since visually salient but geometrically mismatched tokens may receive high attention weights. Moreover, global dense interaction is computationally expensive for high-resolution UAV imagery and does not explicitly distinguish reliable and unreliable cross-modal regions. These limitations motivate alignment-aware fusion strategies that use local correspondence, confidence estimation, and sparse region selection \cite{dai2017dcn,zhu2020deformabledetr}. GAAT differs from saliency weighting or sparse attention alone by ranking interaction regions with transformation-consistent reliability priors and confining bidirectional updates to geometry-calibrated neighborhoods around a sparse query set.

\subsection{Cross-Modal Registration and Representation Alignment}
Cross-modal registration and representation alignment address different levels of multimodal UAV perception. Calibration-based projection and image registration estimate explicit geometric correspondences or spatial transforms, whereas representation alignment constrains latent features without necessarily warping either input. Existing geometric approaches commonly use sensor calibration, time synchronization, global positioning system/inertial measurement unit (GPS/IMU) metadata, or learning-based correspondence estimation. Calibration-based methods use intrinsic and extrinsic sensor parameters to project modalities into a common coordinate system. Registration-based methods estimate affine, homography, or deformable transformations using handcrafted features, structural descriptors, mutual information, optical flow, or deep matching networks \cite{lowe2004sift,rublee2011orb,fischler1981ransac,teed2020raft,sun2021loftr,detone2018superpoint,sarlin2020superglue,lindenberger2023lightglue}. In many UAV datasets, paired images are instead cropped or resized around a common image center and then treated as aligned samples.

These methods have several limitations in UAV scenarios. First, hardware calibration and metadata-based projection require accurate sensor parameters and time synchronization, which are often unavailable or unstable in large-scale UAV data collection. Second, global affine or homography transforms cannot fully handle height-dependent parallax, oblique-view deformation, lens distortion, and local displacement from timing offsets under platform motion. Third, classical registration methods rely on repeatable textures or structural keypoints, while heterogeneous modalities may have very different appearances. For example, RGB imagery emphasizes texture and color, whereas IR or other physical modalities respond to temperature, material, or geometry. This appearance gap makes robust local matching difficult.

Self-supervised and contrastive methods instead align image- or token-level representations, often defining positives from paired images or identical patch indices \cite{chen2020simclr,he2020moco,caron2021dino,cong2022satmae,he2022mae}. Their supervision therefore inherits the assumption that coarse image pairing remains valid after tokenization. syncPATC does not estimate an RGB--IR registration warp; it learns transformation-consistent confidence and local query-offset priors from known synchronized transformations. MG-Sparse-MMA uses these priors to select local fusion neighborhoods, while RA-QCGCL uses them to select patch- and query-level contrastive relations. The resulting novelty is a shared reliability signal that links the alignment assumption, supervision signal, and fusion mechanism.

\section{UAVMeta Dataset and StateBench}
\subsection{Dataset Construction}
\label{subsec:dataset}
UAVMeta is a metadata-enriched multimodal UAV corpus built around paired RGB and IR cameras, the sensing setting studied here. Its schema and benchmark protocol also accommodate SAR, depth, multispectral, and other sensing streams without changing the task definitions or state-evaluation logic. The data were collected using UAV platforms equipped with paired sensors, covering diverse scenes and flight conditions. UAVMeta consists of 2,575 synchronized image pairs spanning daytime and nighttime, with 1,715/572/288 pairs in the train/validation/test splits. The benchmark is partitioned by acquisition rather than by frame, so temporally adjacent samples cannot straddle the train/validation/test boundary. The RGB and IR streams are time-synchronized but retain residual displacement from viewpoint, optics, and platform motion, a setting reflected in existing paired visible--infrared benchmarks \cite{jia2021llvip,suo2023hituav,nie2025m3ot,ouyang2025kust4k,ying2025visiblethermal,sun2020dronevehicle,zhang2022vtuav}.

Compared with existing UAV perception datasets (Table~\ref{tab:uav_dataset_comparison}), UAVMeta offers paired RGB--IR streams that are coarsely image-center aligned while retaining realistic local displacement, creating a setting where patch-level correspondence remains uncertain. It also provides several task annotations within the same acquisition domain. VisDrone and UAVDT provide multi-task annotations but are RGB-only \cite{zhu2021visdrone,du2018uavdt}, whereas KUST4K and other visible--IR benchmarks generally focus on one task \cite{ouyang2025kust4k}. UAVMeta 1.0.0-rc1 provides paired RGB--IR semantic-segmentation, object-detection, and scene-classification annotations. UAVMeta also includes acquisition-state CARS, OSGS, VSS, and FMCS scores, allowing StateBench to evaluate visual predictions alongside acquisition-state scores.

Raw RGB--IR pairs are aligned using local matches and a global affine transform: (i) LoFTR estimates cross-modal dense correspondences on structure-rich regions for daytime scenarios, while a combination of SuperPoint and LightGlue delivers sparse correspondences under weak visual conditions \cite{sun2021loftr,detone2018superpoint,lindenberger2023lightglue}; (ii) a global affine transform is fitted to the matches with RANSAC, with homography disabled to avoid overfitting \cite{lowe2004sift,fischler1981ransac}; (iii) the RGB frame is warped to the IR coordinate system without local deformation; (iv) patch-level quality scores filter low-confidence regions and output diagnostic maps for manual inspection. The pair-level registration-quality score is normalized to a pair-level quality prior $q_b\in(0,1]$ for contrastive weighting. This preprocessing yields coarse image-center alignment while retaining the local displacement that motivates GAAT.

\subsection{Multi-Task Annotation}

\begin{figure*}[!t]
\centering
\includegraphics[width=\textwidth]{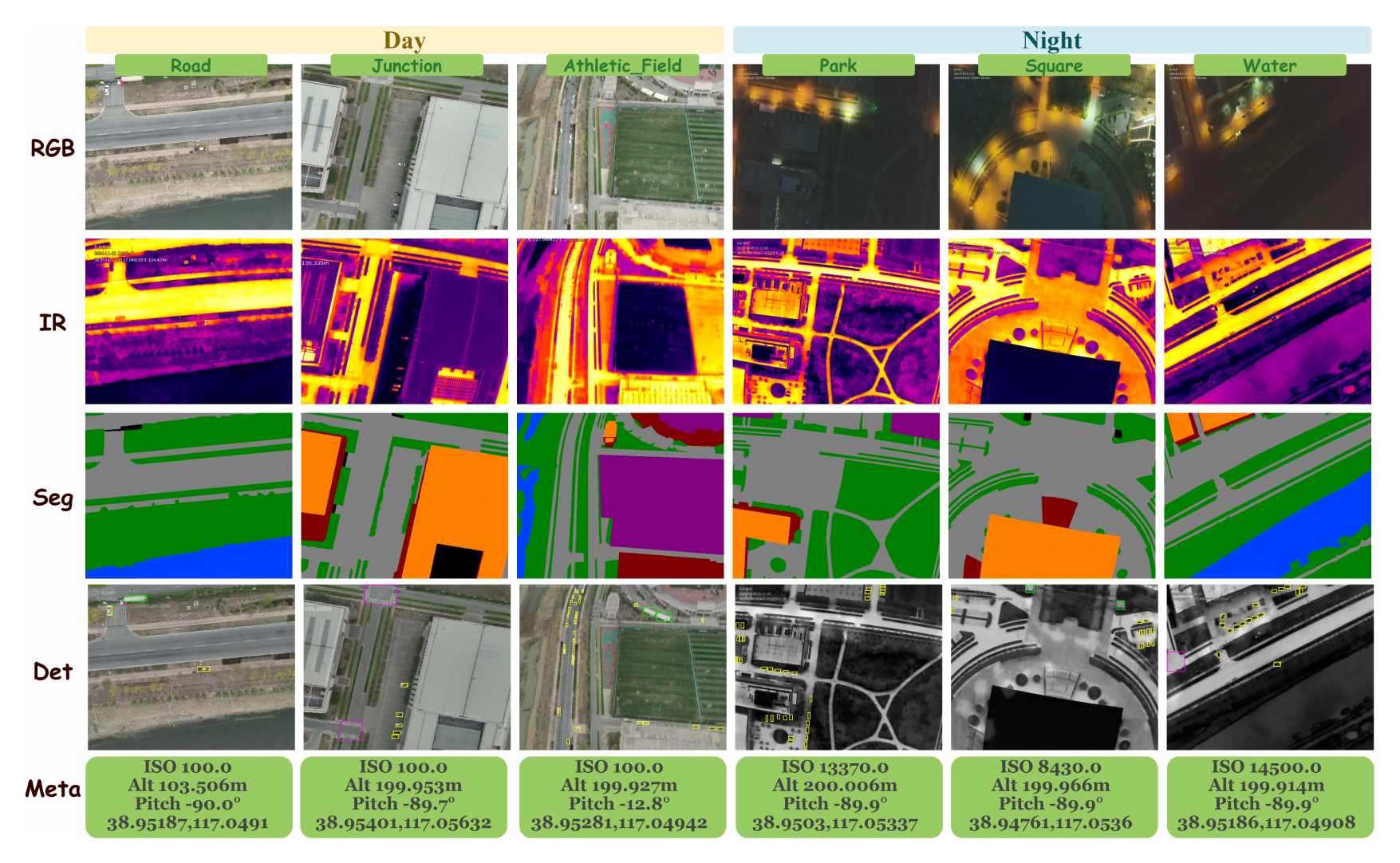}
  \caption{Representative UAVMeta samples. Columns span daytime and nighttime scene classes; rows show synchronized RGB and IR images, segmentation masks, detection annotations, and acquisition metadata.}
\label{fig:uavmeta}
\end{figure*}
  
As shown in Fig.~\ref{fig:uavmeta}, UAVMeta 1.0.0-rc1 supports object detection (bounding boxes), semantic segmentation (pixel-wise labels), and scene classification (per-frame categories) across synchronized RGB--IR streams. Detailed protocols for the released tasks are provided in Appendix~\ref{sec:appendix}. Existing UAV benchmarks usually evaluate one or two related tasks. They are useful for focused evaluation, but do not provide a consistent view of a pretrained model across task abilities. They also rely on task-specific metrics (mIoU for segmentation, mAP for detection, and overall accuracy for classification) that measure \emph{what} the model predicts rather than \emph{under what acquisition conditions} each prediction is made. Altitude, viewpoint angle, camera settings, sensor exposure, illumination, and flight maneuver intensity all affect task difficulty, yet existing protocols rarely quantify them. StateBench addresses this gap through acquisition-state metrics (Sec.~\ref{subsec:state-metrics}) and the MSPA-4D protocol (Sec.~\ref{subsec:mspa-4d}).

\begin{table}[!t]
\caption{Comparison of representative UAV perception datasets by paired modality, annotation coverage, multi-task support, and acquisition-state scores.}
\label{tab:uav_dataset_comparison}
\centering
\scriptsize
\setlength{\tabcolsep}{1.5pt}
\renewcommand{\arraystretch}{1.08}
\resizebox{\columnwidth}{!}{%
\begin{tabular}{@{}l c c c c c@{}}
\hline
\textbf{Dataset} 
& \textbf{Size} 
& \makecell{\textbf{Paired}\\\textbf{Multimodal}} 
& \makecell{\textbf{Fully Annotated}\\\textbf{Samples}} 
& \makecell{\textbf{Multi-Task}\\\textbf{Labels}} 
& \makecell{\textbf{Acquisition-State}\\\textbf{Scores}} \\
\hline
UAVid
& 300 images
& \redcross 
& \greencheck 
& \redcross 
& \redcross \\

VisDrone
& \makecell{10K images}
& \redcross 
& \greencheck 
& \greencheck 
& \redcross \\

UAVDT
& 80K frames
& \redcross 
& \greencheck 
& \greencheck 
& \redcross \\

AU-AIR
& 32K frames
& \redcross
& \greencheck 
& \redcross 
& \redcross \\

HIT-UAV
& 2.9K images
& \redcross 
& \greencheck 
& \redcross 
& \redcross \\

DroneRGBT 
& 3.6K pairs 
& \greencheck 
& \greencheck 
& \redcross 
& \redcross \\

DroneVehicle 
& 28K pairs 
& \greencheck 
& \greencheck 
& \redcross 
& \redcross \\

KUST4K 
& 4.0K pairs 
& \greencheck 
& \greencheck 
& \redcross 
& \redcross \\

VTUAV
& 1.7M pairs 
& \greencheck 
& \redcross 
& \redcross 
& \redcross \\

UAVScenes
& 120K samples 
& \greencheck 
& \redcross 
& \redcross  
& \redcross \\
\hline

\textbf{UAVMeta (Ours)} 
& \textbf{2.5K pairs} 
& \greencheck 
& \greencheck 
& \greencheck 
& \greencheck \\
\hline
\end{tabular}%
}
\end{table}
\subsection{UAV Acquisition-State Metrics}
\label{subsec:state-metrics}

We extract latitude/longitude $(\mathit{lat}_i, \mathit{lon}_i)$, height $h_i$, focal length $f_i$, zoom $z_i$, gimbal pitch/roll $(\theta_i, \phi_i)$, and yaw $\psi_i$ from platform metadata. Speed $v_i$, acceleration $a_i$, height rate $\dot h_i$, and yaw rate $\omega_i$ are derived via finite differences with a 3-to-5-frame rolling median to suppress GPS jitter. All quantities are normalized via $N(x; Q_5, Q_{95}) = \mathrm{clip}((x-Q_5)/(Q_{95}-Q_5+\epsilon), 0, 1)$ using training-set percentiles.

\textbf{CARS: Camera Acquisition Reliability Score.} CARS quantifies the reliability with which the RGB--IR pair is acquired. Because both streams are acquired on the same platform, CARS combines RGB photometric components --- luminance balance $q^{\text{hist}}$, clipping ratio $q^{\text{clip}}$, and sharpness $q^{\text{sharp}}$ --- with IR thermal components --- dynamic range $q^{\text{dyn}}$, hot/cold saturation $q^{\text{sat}}$, and local thermal contrast $q^{\text{con}}$ --- as well as a motion-blur risk $q^{\text{mot}}$ derived from the shutter time, speed, and yaw rate when available:
\begin{equation}
\text{CARS}_i = 100 \cdot \!\sum_{k} w^{\text{c}}_k\,q^{(k)}_i, \quad \textstyle\sum_k w^{\text{c}}_k = 1.
\label{eq:cars}
\end{equation}
Sub-components without valid inputs are masked out and the remaining weights are re-normalized.

\textbf{OSGS: Observation Scale and Ground Sampling Score.} OSGS quantifies the spatial scale of the ground footprint observed by the UAV. When camera calibration is available, the field-of-view (FOV)-based footprint dimensions $W^g_i = 2h_i\tan(\mathit{FOV}_x/2)$ and $H^g_i = 2h_i\tan(\mathit{FOV}_y/2)$ are combined with the image width and height to compute the mean per-pixel ground sampling distance (GSD) $\mathrm{GSD}_i$, and we set $g_i=\log(\mathrm{GSD}_i+\epsilon)$. Otherwise, the calibration-free proxy $g_i = \log(h_i / (f_i z_i + \epsilon))$ is used.
\begin{equation}
\text{OSGS}_i = 100 \cdot N(g_i;\,Q_5,\,Q_{95}).
\label{eq:osgs}
\end{equation}

\textbf{VSS: Viewpoint Stability Score.} VSS captures both the gimbal's proximity to a nominal nadir viewpoint and the stability of the heading. With nominal pitch $\theta_0 = -90^\circ$ and roll $\phi_0 = 0^\circ$,
\begin{equation}
\text{VSS}_i = 100 \cdot \exp\!\Big(\!-\tfrac{|\theta_i-\theta_0|}{\sigma_\theta}-\tfrac{|\phi_i-\phi_0|}{\sigma_\phi}-\tfrac{|\omega_i|}{\sigma_\omega}\Big),
\label{eq:vss}
\end{equation}
with default tolerances $(\sigma_\theta, \sigma_\phi, \sigma_\omega) = (8^\circ, 8^\circ, 25^\circ\!/\text{s})$.

\textbf{FMCS: Flight Maneuver Complexity Score.} FMCS encodes a state \emph{intensity} rather than a quality score: a higher value indicates a more aggressive maneuver, not a better image. The score combines the normalized speed, acceleration, yaw rate, and height rate as
\begin{equation}
\resizebox{0.88\linewidth}{!}{$\displaystyle
\text{FMCS}_i =
100\!\cdot\!\bigl(
w_v N(v_i)+w_a N(a_i)+w_\omega N(|\omega_i|)+w_h N(|\dot h_i|)
\bigr)
$}
\label{eq:fmcs}
\end{equation}
where $w_v, w_a, w_\omega, w_h$ are hyperparameters that balance the contribution of each motion component (see Tab.~\ref{tab:hparams}). When training-set statistics are unavailable, the upper percentiles of the normalizer fall back to the physical limits $(15\,\text{m}/\text{s},\,4\,\text{m}/\text{s}^2,\,45^\circ\!/\text{s},\,1\,\text{m}/\text{s})$.

The four target scores $S_i$ (namely CARS$_i$, OSGS$_i$, VSS$_i$, and FMCS$_i$) are stored as a per-pair vector together with a component-validity mask $m_{ij} \in \{0,1\}$ that flags missing inputs.

\subsection{StateBench: Unified Multimodal State Evaluation}
\label{subsec:mspa-4d}

To benchmark a model's ability to predict the four state scores from a single multimodal pair, we propose \textbf{StateBench}, a unified state prediction benchmark evaluated via the \emph{Multi-State Prediction Accuracy} (MSPA-4D) protocol. StateBench evaluates prediction of metadata- and image-derived acquisition-state proxy scores as a complement to downstream task metrics. Let $\hat s_{ij}$ be the prediction and $s_{ij}$ the ground truth for frame $i$ and metric $j$. With a per-metric tolerance $\tau_j$, the unit-bounded error is
\begin{equation}
e_{ij} = \mathrm{clip}\!\Big(\tfrac{|\hat s_{ij} - s_{ij}|}{\tau_j},\, 0,\, 1\Big).
\label{eq:mspa-error}
\end{equation}
The dataset-level MSPA-4D score is then the validity-masked and activity-weighted accuracy
\begin{equation}
\text{MSPA-4D} = 100 \cdot \frac{\sum_{i,j} w_j r_j m_{ij}(1 - e_{ij})}{\sum_{i,j} w_j r_j m_{ij} + \epsilon},
\label{eq:mspa-4d}
\end{equation}
where the activity weight $r_j = \mathrm{clip}((Q_{95}(S_j)-Q_5(S_j))/30,\,0.25,\,1.0)$ down-weights near-constant components, reducing the influence of quasi-static dimensions. The component weights $w_j$ and tolerances $\tau_j$ for $(\text{CARS}, \text{OSGS}, \text{VSS}, \text{FMCS})$ are hyperparameters listed in Tab.~\ref{tab:hparams}.

Because every multimodal pair in UAVMeta is coarsely image-center aligned and time-synchronized but may retain residual local misalignment, the four state scores are derived from shared platform metadata and, where applicable, modality-specific image statistics (Sec.~\ref{subsec:state-metrics}). StateBench uses the paired input; the main table reports the scalar MSPA-4D score and per-metric mean absolute error $\text{MAE}_j$.

\section{Methodology}
\label{sec:method}

\subsection{Framework Overview}
\label{subsec:framework}

\begin{figure*}[!t]
  \centering
  \includegraphics[width=\textwidth]{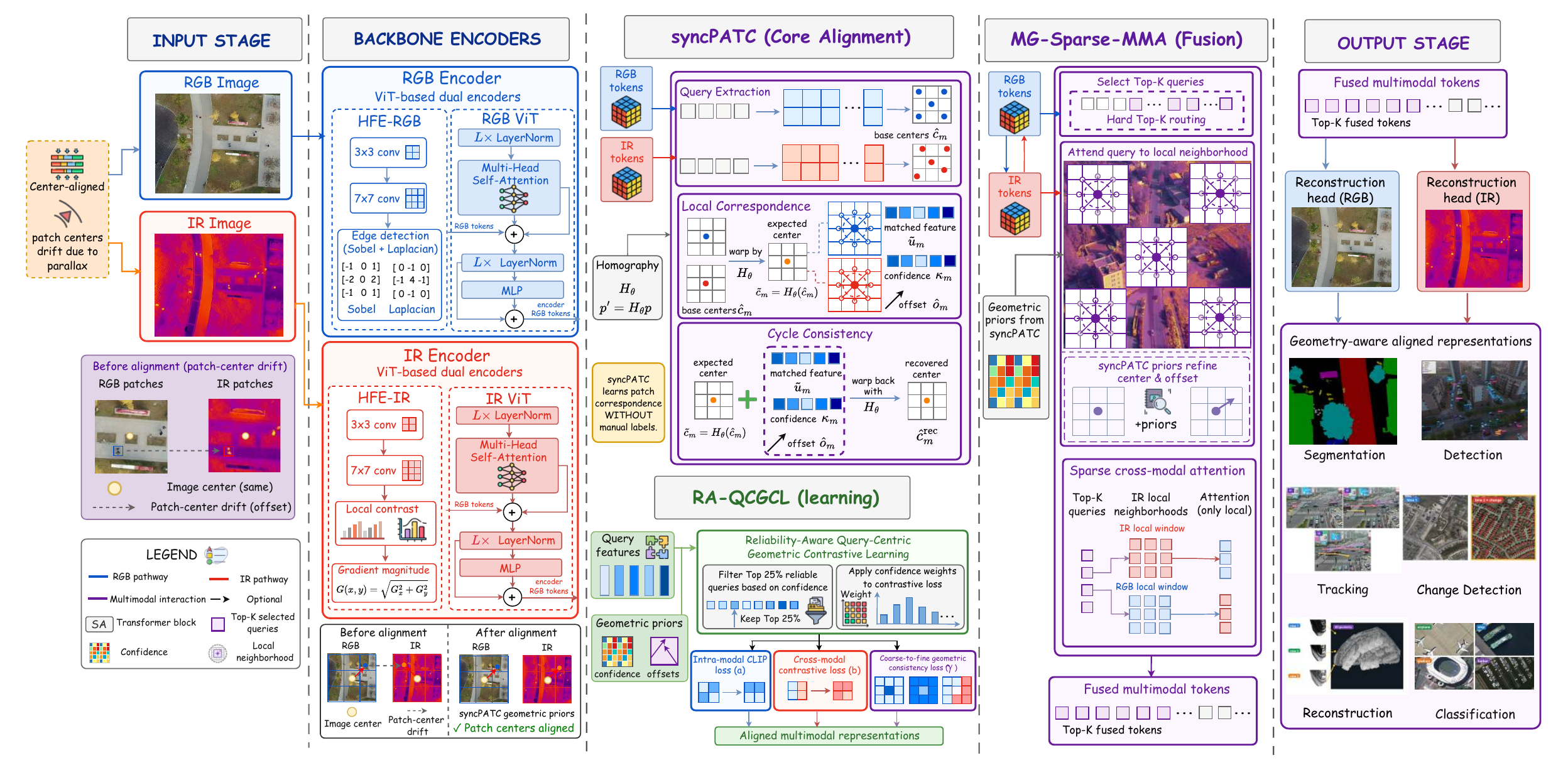}
  \caption{GAAT architecture. Modality-specific HFE blocks augment the RGB and IR encoders; syncPATC, MG-Sparse-MMA, and RA-QCGCL provide geometric priors, sparse fusion, and cross-granularity contrastive supervision, respectively.}
  \label{fig:pipeline}
  \end{figure*}

The GAAT framework implements three linked design choices introduced in Sec.~\ref{sec:intro}; the resulting information flow is summarized in Fig.~\ref{fig:pipeline}. First, \emph{estimate local correspondence reliability before fusion}: the model estimates patch-center reliability in a dedicated self-supervised module and uses it to define positives and attention anchors. Second, \emph{fuse only reliable regions}: reliable UAV cross-modal anchors are sparse, so fusion is routed through a small set of semantically salient, reliability-ranked queries. Third, \emph{supervise at the same granularity as the interaction}: the contrastive objective follows the reliable patches and query bottleneck used by the fusion module. syncPATC implements the first choice, MG-Sparse-MMA implements sparse query-mediated fusion, and RA-QCGCL provides the matched cross-granularity supervision. The remainder of this subsection introduces the input formulation, the dual encoders, the information flow between the three modules, and the overall pretraining objective.

Let $(I_R, I_T)$ be a coarsely center-aligned UAV image pair with $I_R\!\in\!\mathbb{R}^{3\times H_0\times W_0}$ (RGB) and $I_T\!\in\!\mathbb{R}^{1\times H_0\times W_0}$ (IR). For each pair, the synchronized-view pipeline samples an affine parameter set $\theta$ and applies the resulting warp $\mathcal{W}_\theta$ jointly to RGB and IR, producing a synchronized counterpart view $(I_R', I_T')$ together with the affine transform matrix $H_\theta\!\in\!\mathbb{R}^{3\times 3}$, an out-of-bounds validity mask $M_v$, and a normalized view gap $g\!\in\![0,1]$. The view gap $g$ quantifies the magnitude of the transformation as a weighted combination of rotation, translation, scale, and shear components, each normalized by its respective maximum range and combined as $g\!=\!\eta_{\text{rot}}\,g_{\text{rot}}\!+\!\eta_{\text{trans}}\,g_{\text{trans}}\!+\!\eta_{\text{scale}}\,g_{\text{scale}}\!+\!\eta_{\text{shear}}\,g_{\text{shear}}$. Because $\mathcal{W}_\theta$ acts jointly on the two modalities, the relation between each base view and its synthetically warped counterpart is exactly known for this synthetic pair and provides the self-supervision signal that drives syncPATC.

The two modalities are encoded by a pair of architecturally identical but independently parameterized Swin Transformer V2 encoders $f_R, f_T$, following the hierarchical transformer designs of ViT, Swin, and Swin V2 \cite{vaswani2017transformer,dosovitskiy2021vit,liu2021swin,liu2022swinv2}. After patch embedding and stages 1 and 2 of each encoder, we insert a modality-specific high-frequency/contrast enhancement (HFE) block that injects modality-specific image cues into the token representation. Both modalities first apply depthwise $3\!\times\!3$ and $7\!\times\!7$ convolutions for multi-scale context, after which the hand-crafted branch differs across modalities: the RGB side appends a single \emph{high-frequency} branch that fuses Sobel and Laplacian responses on the channel-averaged map and captures the texture and edge structure characteristic of visible imagery, while the IR side appends two \emph{contrast} branches --- a Sobel magnitude and a $5\!\times\!5$ local-contrast map --- which respond respectively to thermal boundaries and homogeneous-region heat differentials. The concatenated branches are projected by a $1\!\times\!1$ convolution $P_m$, modulated by a parallel sigmoid gate $\sigma(G_m(\cdot))$, and added back to the token map as a gated residual:
\begin{equation}
Y_m = X_m + \sigma(G_m(B_m)) \odot P_m(B_m),
\label{eq:hfe}
\end{equation}
where $B_m$ denotes the concatenated multi-scale and hand-crafted branches. We refer to this convention as \emph{HF for RGB, contrast for IR}: each modality receives the image cue to which its imaging mechanism is most sensitive. Each encoder thus emits a multi-scale token set $\{X_m^{(s)}\}$ with $m\!\in\!\{R,T\}$.

Two distinct but coupled streams flow through the encoders. The \emph{clean} (unmasked, no-dropout) mid-stage tokens $X_m^{(s_p)}$ enter syncPATC (Sec.~\ref{subsec:syncpatc}), which estimates transformation-consistent patch-center reliability under the synchronized view pair. It emits four geometric priors---token-level confidence $\pi_m^{\text{tok}}$, query-level confidence $\pi_{m,k}^q$, query centers $\bar c_m^k$, and sub-token offsets $\bar o_m^k$---all rescaled to the fusion grid before downstream use.

\noindent The masked top-stage tokens $X_m^{(4)}$, obtained under random patch masking $M$ and stochastic modality dropout, enter MG-Sparse-MMA\footnote{The prefix MG- denotes the modality-guided sparse interaction implemented through query-selected local cross-modal updates.} (Sec.~\ref{subsec:mgmma}). Guided by the same priors, MG-Sparse-MMA performs sparse bidirectional cross-modal fusion. It emits fused tokens $\tilde X_R, \tilde X_T$, cross-modal queries $Q_m$, attention maps $A^q_m$, and per-query saliencies $\tilde s_m$. RA-QCGCL (Sec.~\ref{subsec:raqcgcl}) combines these priors, image-domain reliability maps $\rho_m$, and the MG-Sparse-MMA queries into a reliability-weighted three-granularity contrastive objective. Lightweight $1\!\times\!1$ Conv + PixelShuffle decoders $D_m$ then reconstruct $\hat I_m\!=\!D_m(\tilde X_m)$ for the masked image modeling target.

The full pretraining objective couples five terms. The asymmetric masked image modeling loss $\mathcal{L}^{\text{mim}}_m$ augments the standard pixel + Sobel-gradient base with a Haar high-frequency term on the RGB side and a $5\!\times\!5$ local-contrast term on the IR side, matching the modality-specific image cues injected by HFE. This follows masked-image-modeling objectives used in general vision and remote-sensing foundation models \cite{he2022mae,sun2023ringmo,cong2022satmae,reed2023scalemae}. A visibility-weighted cosine distillation $\mathcal{L}^{\text{kd}}$ matches the masked student tokens from the default teacher modality (RGB) to the output of a frozen, architecturally identical teacher, following the teacher--student formulation of knowledge distillation \cite{hinton2015distill}, with the per-sample loss defined as the mean cosine distance across student and teacher tokens. The RA-QCGCL term $\mathcal{L}^{\text{ra-qcgcl}}$ enforces cross-modal contrastive alignment on the top-$K$ reliability-ranked patch subset using symmetric InfoNCE-style objectives \cite{chen2020simclr,he2020moco,caron2021dino}; a modality-completion term $\mathcal{L}^{\text{mod}}$ reuses $\mathcal{L}^{\text{mim}}_m$ on the modality-dropout mask so that fully or partially erased modalities can still be reconstructed; and the syncPATC term $\mathcal{L}^{\text{patc}}_m$ supervises the geometric learning itself. Combining the five terms gives
\begin{equation}
\resizebox{0.90\linewidth}{!}{$\displaystyle
\mathcal{L}
= \lambda_r\!\sum_{m}\mathcal{L}^{\text{mim}}_m
+ \lambda_t\mathcal{L}^{\text{kd}}
+ \lambda_c\mathcal{L}^{\text{ra-qcgcl}}
+ \lambda_d\mathcal{L}^{\text{mod}}
+ \lambda_p\!\sum_m\mathcal{L}^{\text{patc}}_m
$}
\label{eq:total}
\end{equation}
where the five weights $\{\lambda_r,\lambda_t,\lambda_c,\lambda_d,\lambda_p\}$ balance the corresponding terms; their default values are listed in Tab.~\ref{tab:hparams}.
The HFE blocks, distillation, and modality-completion loss support pretraining, whereas syncPATC, MG-Sparse-MMA, and RA-QCGCL carry the alignment design.

\subsection{syncPATC: Synchronized Patch-Center Alignment}
\label{subsec:syncpatc}

\begin{figure}[!t]
  \centering
  \includegraphics[width=\columnwidth]{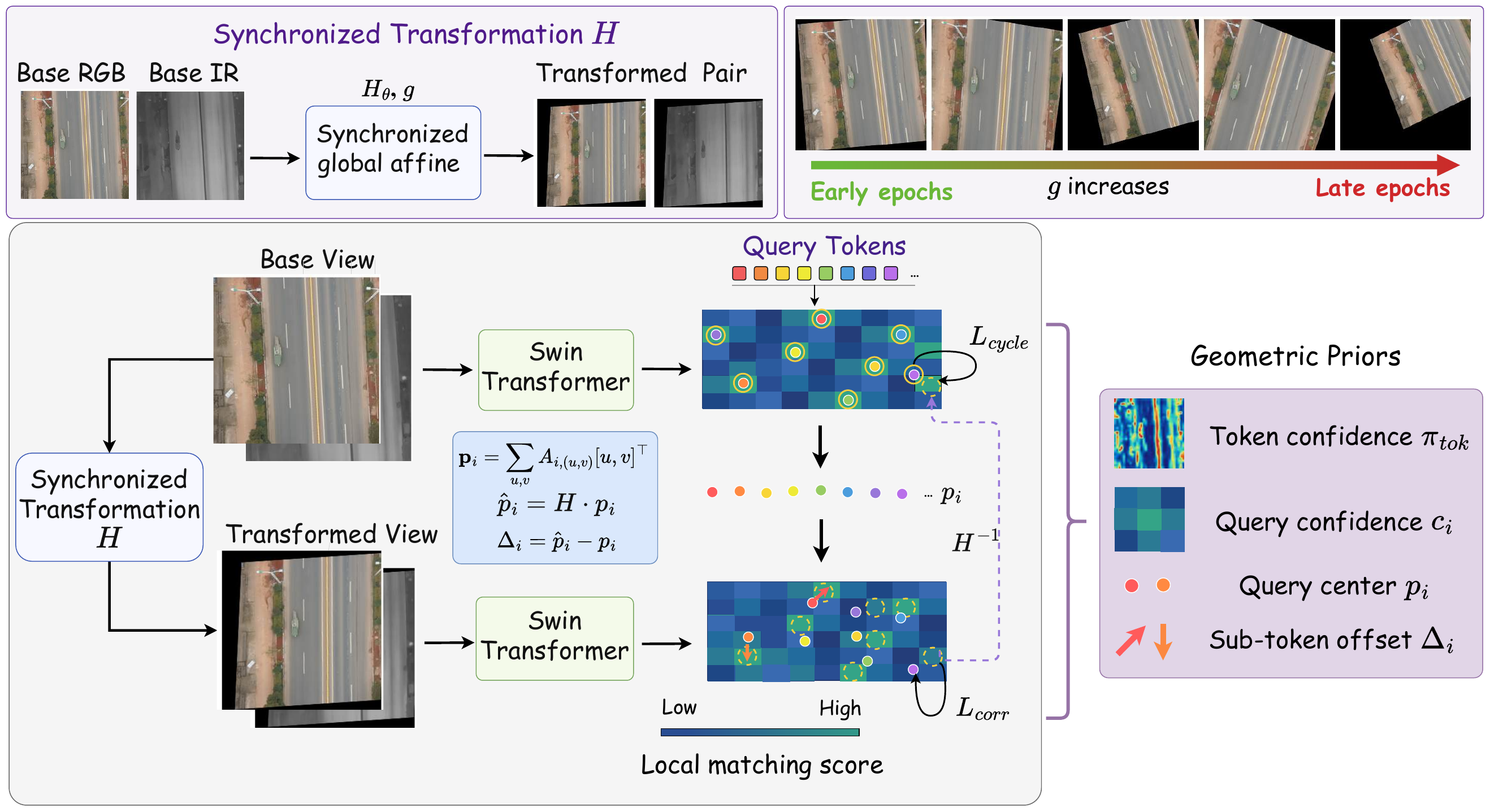}
  \caption{Overview of syncPATC. A shared affine transform $H_\theta$ generates synchronized counterpart views under a view-gap curriculum. Transform-guided matching and cycle consistency estimate local reliability, producing token and query confidence, query centers, and sub-token offsets.}
  \label{fig:syncpatc}
\end{figure}

\begin{figure}[!t]
  \centering
  \includegraphics[width=\columnwidth]{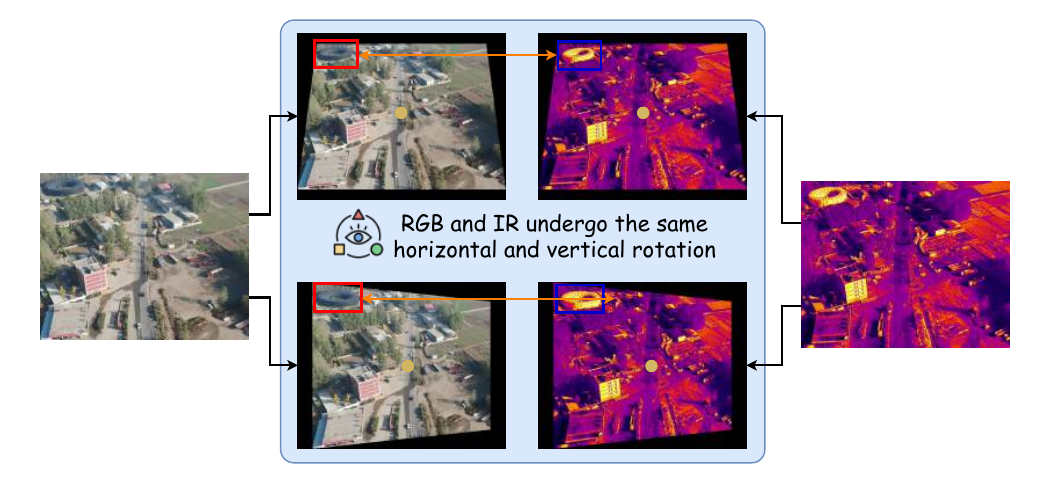}
  \caption{Expanded synchronized transformation in syncPATC. The same sampled $H_\theta$ is applied to RGB and IR; its known geometry guides within-modality query-offset estimation. Boxes and arrows indicate corresponding regions, not RGB--IR registration offsets.}
  \label{fig:sync-transform}
\end{figure}

Fig.~\ref{fig:sync-transform} expands the synchronized-transformation block in Fig.~\ref{fig:syncpatc}: the same known $H_\theta$ is applied to the RGB and IR streams, while base-to-counterpart matching and inverse-warp cycle consistency are computed separately within each modality. The resulting token and query confidence priors prioritize the top-$K_s$ regions for fusion, while residual sub-token offsets refine the local sampling locations of MG-Sparse-MMA; these offsets are query offsets rather than RGB--IR registration offsets.

\textbf{Query centers.} Let $Z_m, Z_m'\!\in\!\mathbb{R}^{B\times HW\times D}$ denote the clean base- and counterpart-view tokens at stage $s_p$. We learn $N_q$ query tokens $\Phi_m$ per modality, run multi-head cross-attention $\mathrm{MHCA}(\Phi_m, Z_m, Z_m)$ with head-averaged outputs $(Q_m, A_m)$, and similarly obtain $(Q'_m, A'_m)$ from $Z_m'$. With $p_n\!\in\!\mathbb{R}^2$ the grid coordinate of token $n$, the $k$-th query center is the attention-weighted average $\hat c_m^k\!=\!\sum_n A_{m,k,n}\,p_n/\!\sum_j\!A_{m,k,j}$.

\textbf{Affine-transform-guided local correspondence.} Mapping $\hat c_m^k$ through $H_\theta$ in the pixel domain and back to the token grid (accounting for the stride-induced half-pixel offset) yields the expected counterpart-view center
\begin{equation}
\tilde c_m^k = \Pi_{s_p}\!\big(H_\theta\,\Pi_{s_p}^{-1}(\hat c_m^k)\big).
\label{eq:patc-warp}
\end{equation}
Because the same warp is applied to both modalities, the synthetic pair shares a known geometric perturbation; this supplies the self-supervision signal for transformation consistency. syncPATC learns transformation-consistent, modality-specific reliability priors, which are then used by the paired fusion and contrastive branches.
We bilinearly sample $Z_m'$ on a discrete neighborhood $\{\Delta_l\}_{l=1}^{L}$ of radius $r$ around $\tilde c_m^k$, with a per-sample validity indicator $\nu_{m,k,l}\!\in\!\{0,1\}$ derived by warping $M_v$ to the same grid. With $\ell_2$-normalized projections $\hat q\!=\!\mathcal{N}(W_qQ_m^k)$ and $\hat k_l\!=\!\mathcal{N}(W_k u_{m,k,l})$ and a Gaussian neighborhood prior of bandwidth $\sigma$, the masked correspondence distribution is
\begin{equation}
\resizebox{0.89\linewidth}{!}{$\displaystyle
\alpha_{m,k,l} =
\frac{\nu_{m,k,l}\,e^{s_{m,k,l}}}
 {\sum_{l'}\nu_{m,k,l'}\,e^{s_{m,k,l'}}+\epsilon},
\quad
s_{m,k,l}=
\frac{\langle\hat q,\hat k_l\rangle}{\sqrt D}
-\frac{\|\Delta_l\|_2^2}{2\sigma^2}
$}
\label{eq:patc-attn-local}
\end{equation}
where $\epsilon>0$ is a small numerical constant. If a query has no valid
neighbor ($N_v^k=0$), we set $\alpha_{m,k,l}=0$ for every $l$,
$\tilde u_m^k=0$, and $\hat o_m^k=0$, and exclude that query from the
normalized correspondence and cycle losses. The matched token, sub-token offset,
and entropy-normalized confidence for valid neighborhoods are
\begin{equation}
\tilde u_m^k\!=\!\mathrm{LN}\!\Big(\!\sum_l\!\alpha_{m,k,l}W_v u_{m,k,l}\!\Big),\;\;\hat o_m^k\!=\!\sum_l\!\alpha_{m,k,l}\Delta_l,
\label{eq:patc-match}
\end{equation}
\begin{equation}
\kappa_m^k = \mathbf{1}[N_v^k\!>\!0]\!\cdot\!\Big(1-\tfrac{\mathcal{H}_k}{\log\max(N_v^k,2)}\Big)_{[0,1]},
\label{eq:patc-conf}
\end{equation}
where $N_v^k\!=\!\sum_l\nu_{m,k,l}$ and $\mathcal{H}_k\!=\!-\!\sum_l\alpha_{m,k,l}\log\alpha_{m,k,l}$, so that a more peaked distribution receives a higher confidence.

\textbf{Loss with view-gap curriculum.} Let $d(a,b)\!=\!1\!-\!\cos(a,b)$. Both view-side queries are anchored to the matched token, and a cycle term re-enters the base view through $H_\theta^{-1}$ to give a recovered feature $\check u_m^k$ with confidence $\check\kappa_m^k$. The correspondence loss pulls both base-view and counterpart-view queries toward the matched token:
\begin{equation}
\mathcal{L}^{\text{corr}}_m = \mathbb{E}_w\!\big[\tfrac{1}{2}(d(Q_m^k,\tilde u_m^k)\!+\!d(Q'^k_m,\tilde u_m^k))\big],
\label{eq:patc-corr}
\end{equation}
and the cycle-consistency loss enforces that the base-view query remains consistent after a round-trip transformation:
\begin{equation}
\mathcal{L}^{\text{cycle}}_m = \mathbb{E}_{w\check\kappa}\!\big[d(Q_m^k,\check u_m^k)\big].
\label{eq:patc-cycle}
\end{equation}
The total syncPATC loss combines both terms as $\mathcal{L}^{\text{patc}}_m\!=\!\mathcal{L}^{\text{corr}}_m\!+\!\lambda_{\text{cyc}}\mathcal{L}^{\text{cycle}}_m$, with $\lambda_{\text{cyc}}$ balancing the two objectives, and $\mathbb{E}_w[\cdot]$ denoting the weight-normalized mean over $(b,k)$ with $w_{m,b}^k\!=\!\kappa_{m,b}^k\,\gamma_b$. The view-gap curriculum coefficient
\begin{equation}
\gamma_b = (1-p_b)(\gamma_0-\gamma_1\,g_b)+p_b(\gamma_2+g_b)
\label{eq:patc-curric}
\end{equation}
emphasizes reliable small-gap samples when training progress $p_b\!\to\!0$ and increases the loss weight for large-gap samples when $p_b\!\to\!1$, defining an easy-to-hard weighting schedule.

\textbf{Geometric priors.} syncPATC outputs four quantities, rescaled to the fusion grid before downstream consumption:
\begin{equation}
\begin{aligned}
\pi_m^{\text{tok}}(n)&\!\propto\!\sum_k A_{m,k,n}\kappa_m^k,\qquad
\pi_{m,k}^q\!=\!\tfrac{1}{2}(\kappa_m^k\!+\!\check\kappa_m^k),\\
\bar c_m^k&\!=\!\mathrm{Rescale}(\hat c_m^k),\qquad
\bar o_m^k\!=\!\mathrm{Rescale}(\hat o_m^k).
\end{aligned}
\label{eq:patc-prior}
\end{equation}
Here $\mathrm{Rescale}(\cdot)$ denotes the deterministic coordinate conversion
from the syncPATC stage grid to the fusion grid.
To prevent contrastive shortcuts from reshaping the geometry module, the confidence priors used by RA-QCGCL are detached by default. The priors used by MG-Sparse-MMA can retain a lightweight gradient path for fusion calibration.

\subsection{MG-Sparse-MMA: Modality-Guided Sparse Multimodal Attention}
\label{subsec:mgmma}

\begin{figure*}[!t]
  \centering
  \includegraphics[width=\textwidth]{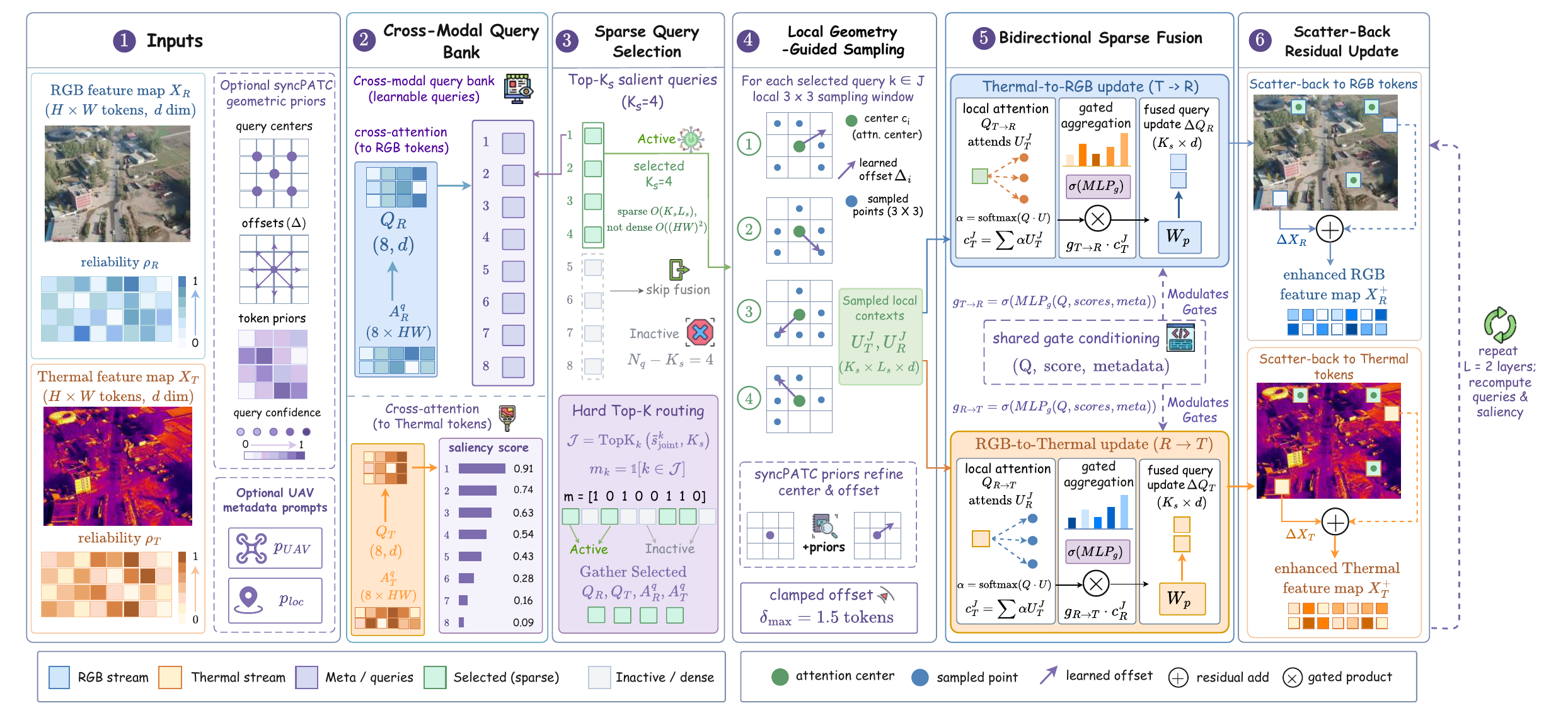}
  \caption{MG-Sparse-MMA architecture. syncPATC reliability reweights candidate RGB and IR queries before top-$K_s$ selection. Query centers and sub-token offsets guide deformable local sampling, followed by bidirectional gated fusion and residual scatter-back.}
  \label{fig:mg-sparse-mma}
\end{figure*}

\textbf{Module organization.} Fig.~\ref{fig:mg-sparse-mma} expands MG-Sparse-MMA as the sparse fusion stage of GAAT. Rather than letting every token attend to every cross-modal token, this module is inserted into the forward path and updates the two modality streams only through a small set of reliable query slots, drawing on deformable convolution, deformable attention, and cross-modal transformer fusion \cite{dai2017dcn,zhu2020deformabledetr,zhang2023cmx}. The query bank first exposes candidate regions in each modality, the reliability-weighted saliency score selects the top-$K_s$ slots, and the selected queries drive geometry-calibrated local attention before their updates are scattered back to the RGB and IR token grids.

\textbf{Cross-modal query bank.} MG-Sparse-MMA stacks $L$ identical layers. Each layer first applies per-modality token self-attention on $X_m\!\in\!\mathbb{R}^{B\times HW\times D}$. A learnable query bank $\Phi^{\mathrm{MMA}}\!\in\!\mathbb{R}^{N_q\times D}$, optionally concatenated with a UAV metadata prompt $\phi_v\!=\!\mathrm{MLP}(v_{\text{meta}})$, extracts modality-specific queries $Q_m$ and head-averaged cross-attention maps $A^q_m$ via $\mathrm{MHCA}(\Phi^{\mathrm{MMA}}\!\cup\!\phi_v, X_m, X_m)$, with only the first $N_q$ slots retained. Because the bank is shared across modalities, slot $k$ retains a shared learned identity while its output remains modality-conditioned. When syncPATC is on, the token prior modulates and re-normalizes the attention so that geometric consistency and semantic importance are coupled:
\begin{equation}
A^q_m \leftarrow \mathrm{Renorm}\!\big(A^q_m \odot \pi_m^{\text{tok}}\big).
\label{eq:cmq-prior}
\end{equation}
The per-query saliency is the reliability-weighted attention mass, further weighted by the query prior:
\begin{equation}
\resizebox{0.88\linewidth}{!}{$\displaystyle
\tilde{s}_m^k =
\Bigl(\sum_n A^q_{m,k,n}\rho_m(n)\Bigr)
\bigl(\lambda_1+\lambda_2\pi_{m,k}^{q}\bigr),
\quad
\tilde{s}_{\mathrm{joint}}^k =
\frac{1}{2}\bigl(\tilde{s}_R^k+\tilde{s}_T^k\bigr)
$}
\label{eq:cmq-score}
\end{equation}
where $\lambda_1$ and $\lambda_2$ balance the base attention score and the prior term (see Tab.~\ref{tab:hparams}).

\textbf{Sparse selection.} The top $K_s$ queries by $\tilde s_{\text{joint}}$ form the ordered selected-query list $\mathcal{J}=(\mathcal{J}_1,\ldots,\mathcal{J}_{K_s})$, where $\mathcal{J}_k$ is the original query index of selected slot $k$. Cross-modal interaction is then confined to a $K_s\!\times\!L_s$ neighborhood with $L_s\!=\!(2r_s\!+\!1)^2$, reducing the cross-modal interaction cost from $\mathcal{O}((HW)^2)$ to $\mathcal{O}(K_s L_s)$.

\textbf{Geometry-calibrated local sampling.} For each selected slot $k=1,\ldots,K_s$, the local sampling center is the mean of the two-modality attention centers, refined by the syncPATC center prior:
\begin{equation}
\hat c^k = \tfrac{1}{2}\!\big(\mu(A^{q,\mathcal{J}_k}_R)\!+\!\mu(A^{q,\mathcal{J}_k}_T)\big),\;\;\hat c^k\leftarrow\tfrac{1}{2}\!\big(\hat c^k\!+\!\tfrac{1}{2}(\bar c_R^{\mathcal{J}_k}\!+\!\bar c_T^{\mathcal{J}_k})\big).
\label{eq:mgmma-center}
\end{equation}
For each direction $m\!\to\!m'$ (target $m'$, source $m$), a learnable offset is predicted from the query pair and metadata prompt, then summed with the syncPATC offset prior and hard-clipped:
\begin{equation}
\delta_{m\to m'}^k = \delta_{\max}\tanh\!\big(\mathrm{MLP}_o([Q_{m'}^{\mathcal{J}_k};Q_m^{\mathcal{J}_k};\bar\phi_v])\big),
\label{eq:mgmma-offset-learn}
\end{equation}
\begin{equation}
\Delta_{m\to m'}^k = \mathrm{clip}_{\delta_{\max}}\!\Big(\delta_{m\to m'}^k + \tfrac{1}{2}(\bar o_R^{\mathcal{J}_k}\!+\!\bar o_T^{\mathcal{J}_k})\Big),
\label{eq:mgmma-offset}
\end{equation}
where $\bar\phi_v$ is the query-axis mean of $\phi_v$ (omitted when no prompt is used).

\textbf{Gated fusion and scatter.} Bilinearly sampling the source feature map $X_m$ at $\{\hat c^k\!+\!\Delta_{m\to m'}^k\!+\!\Delta_l\}_l$ produces the local context $U^k$. With the direction-specific query projection $\tilde Q_{m\to m'}^k\!=\!W^q_{m\to m'}[Q_{m'}^{\mathcal{J}_k};Q_m^{\mathcal{J}_k}]$ and a saliency-aware modality gate $g^k\!=\!\sigma(\mathrm{MLP}_g([Q_{m'}^{\mathcal{J}_k};Q_m^{\mathcal{J}_k};\bar\phi_v;\tilde s_{m'}^{\mathcal{J}_k};\tilde s_m^{\mathcal{J}_k};\tilde s_{\text{joint}}^{\mathcal{J}_k}]))$ that takes its inputs in (target, source, joint) order to match the directional update, the per-query update is
\begin{equation}
\hat y^k
= W_p^{m\to m'}\!\Big(
g^k \sum_l
\mathrm{softmax}_l\!\Big(
\tfrac{\langle \tilde Q_{m\to m'}^k, U_l^k\rangle}{\sqrt D}
\Big)
U_l^k
\Big),
\label{eq:mgmma-query-update}
\end{equation}
and the residual write-back is
\begin{equation}
X_{m'}^{(\ell+1)}
= X_{m'}^{(\ell)}
+ \mathrm{DropPath}\!\Big(
\sum_{k=1}^{K_s} r_n^k \hat y^k
\Big),
\label{eq:mgmma-residual}
\end{equation}
where $r_n^k\!=\!A^{q,\mathcal{J}_k}_{R,n}A^{q,\mathcal{J}_k}_{T,n}/\!\sum_{n'}\!A^{q,\mathcal{J}_k}_{R,n'}A^{q,\mathcal{J}_k}_{T,n'}$ is the joint attention assignment that scatters the sparse update across the target-modality token grid. The two directions IR$\!\to\!$RGB and RGB$\!\to\!$IR are executed in parallel within each layer; we refer to this symmetric scheme as \emph{modality-guided}.

\textbf{Role of syncPATC priors.} The token prior couples geometric reliability into the query attention and hence into the top-$K_s$ selection (Eq.~\eqref{eq:cmq-prior}); the centers and offsets calibrate the local sampling positions (Eqs.~\eqref{eq:mgmma-center}--\eqref{eq:mgmma-offset}); the query prior promotes reliable queries in the saliency ranking (Eq.~\eqref{eq:cmq-score}). Together with the symmetric bidirectional update, these priors help reduce the failure mode in which a salient but spatially mismatched region receives heavy cross-modal attention.

\subsection{RA-QCGCL: Reliability-Aware Query-Guided Cross-Granularity Contrastive Learning}
\label{subsec:raqcgcl}

Let $Z_m\!\in\!\mathbb{R}^{B\times HW\times D}$ denote the masked top-stage tokens refined by MG-Sparse-MMA, with $\ell_2$-normalized form $\hat Z_m$, and let $\hat Q_m$, $A^q_m$, $\tilde s_m$ be the normalized queries, attention maps, and per-query saliencies emitted by MG-Sparse-MMA (Sec.~\ref{subsec:mgmma}). The query notation is module-local: syncPATC queries are computed from clean mid-stage tokens, whereas these queries are computed from masked top-stage tokens.

\begin{figure}[!t]
  \centering
  \includegraphics[width=0.49\textwidth]{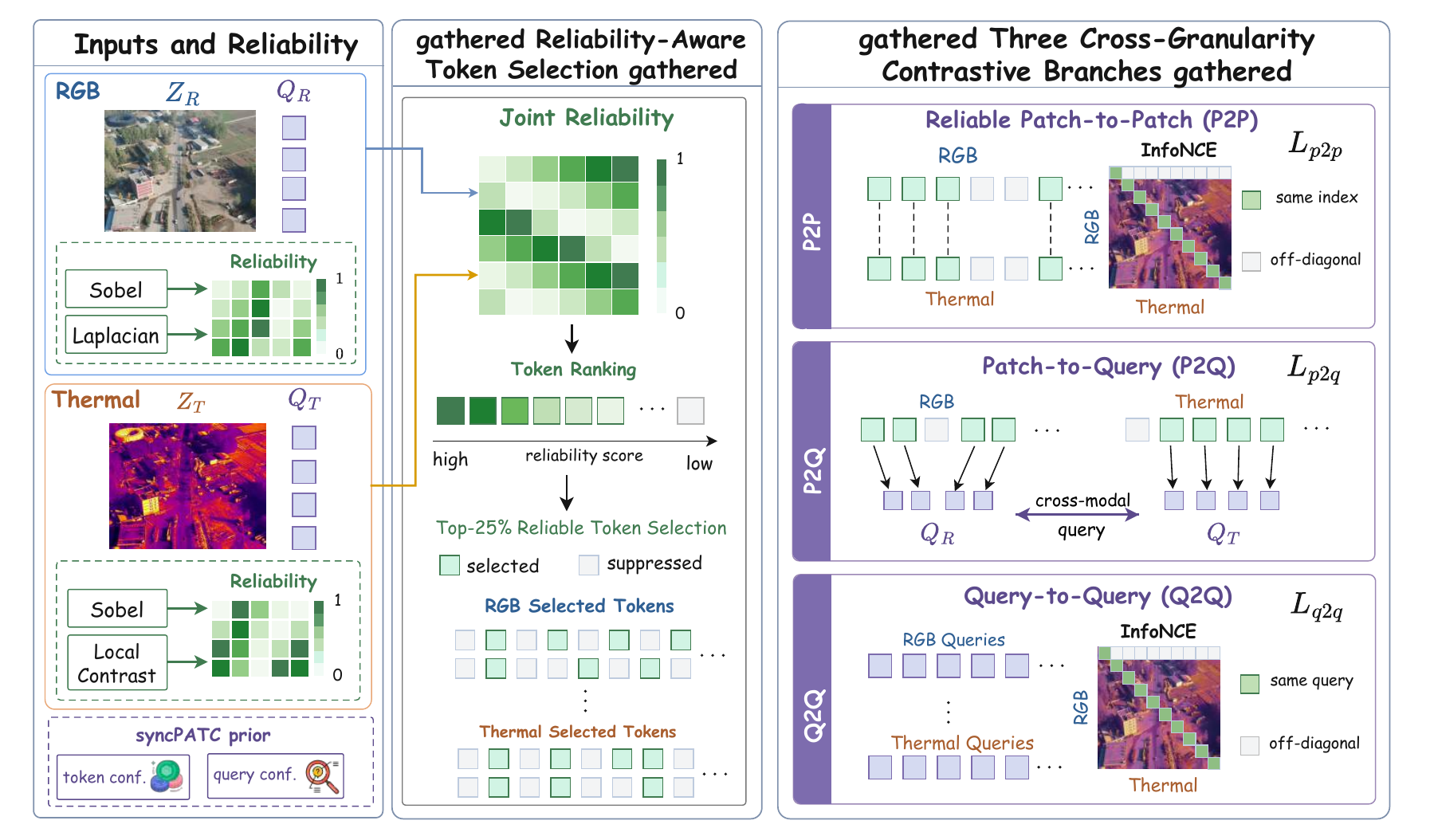}
\caption{RA-QCGCL architecture. Joint geometric and modality-specific reliability selects the top-25\% tokens. P2P aligns spatially anchored patches, P2Q assigns selected patches to semantic queries, and Q2Q aligns cross-modal query slots.}
  \label{fig:raqcgcl}
\end{figure}

\textbf{Module organization.} As detailed in Fig.~\ref{fig:raqcgcl}, RA-QCGCL avoids returning to the dense patch-index assumption after the model has adopted sparse query-mediated fusion. The first stage constructs a joint reliability score from syncPATC priors and modality-specific image evidence, so that token-index positives are used only where local correspondence is credible. The second stage applies complementary contrastive branches with different spatial assumptions: P2P retains strict patch-index supervision on reliable tokens, P2Q transfers selected patches to cross-modal semantic queries, and Q2Q aligns query slots across the batch when patch-level geometry is weak.

 \textbf{Joint reliability.} On the fusion grid, the per-modality reliability map is derived from the RGB luminance proxy $I^{\text Y}_R\!=\!\tfrac{1}{3}\!\sum_c\!I_R^{(c)}$ as a combined Sobel/Laplacian response, and from $I_T$ using a combination of a Sobel magnitude and a $5\!\times\!5$ local-contrast response. Each map undergoes per-sample min--max normalization, point-wise attenuation by $1\!-\!M_m^{\text{drop}}$ (the modality-dropout mask), and adaptive average pooling. When syncPATC is on, the geometric token confidence modulates the maps point-wise and the joint reliability is
\begin{equation}
\bar\rho = \tfrac{1}{2}\big(\rho_R\!\odot\!\pi_R^{\text{tok}} + \rho_T\!\odot\!\pi_T^{\text{tok}}\big).
\label{eq:rho-fuse}
\end{equation}
A top-$K$ subset with $K\!=\!\mathrm{round}(\tau_K HW)$ is selected by $\bar\rho$, yielding the per-sample index set $\mathcal{I}_b$ and weights $w_b^k\!=\!\bar\rho_b(\mathcal{I}_b^k)\,q_b$; $q_b\!\in\!(0,1]$ is a normalized pair-level registration-quality prior retained from alignment preprocessing, so lower-quality registrations contribute less to the patch-level contrastive terms.

\textbf{Reliable patch-to-patch (P2P).} Here and below, $\mathbb{E}_w$ denotes the weight-normalized mean over selected sample-index pairs $(b,k)$ using $w_b^k$. Using the selected token indices as positive labels in a temperature-$\tau_p$ cross-modal InfoNCE,
\begin{equation}
\mathcal{L}^{\text{p2p}} = \tfrac{1}{2}\!\Big(\mathbb{E}_w\!\big[\mathrm{CE}(\tfrac{\hat Z_R^{\mathcal{I}}\hat Z_T^\top}{\tau_p},\mathcal{I})\big] + \mathbb{E}_w\!\big[\mathrm{CE}(\tfrac{\hat Z_T^{\mathcal{I}}\hat Z_R^\top}{\tau_p},\mathcal{I})\big]\Big),
\label{eq:p2p}
\end{equation}
which simultaneously enforces semantic consistency and spatial uniqueness on the selected subset.

\textbf{Patch-to-query (P2Q).} Each selected token is assigned to a query slot in the same modality by $\hat k_m^n\!=\!\arg\max_k A^q_{m,k,n}$ and pulled towards the cross-modal counterpart query,
\begin{equation}
\mathcal{L}^{\text{p2q}} = \tfrac{1}{2}\!\Big(\mathbb{E}_w\!\big[d(\hat Z_R^{\mathcal{I}}, \hat Q_T^{\hat k_R(\mathcal{I})})\big] + \mathbb{E}_w\!\big[d(\hat Z_T^{\mathcal{I}}, \hat Q_R^{\hat k_T(\mathcal{I})})\big]\Big).
\label{eq:p2q}
\end{equation}

\textbf{Query-to-query (Q2Q).} For each query slot $k$ we form an in-batch similarity matrix $S^k_{b,b'}\!=\!\langle\hat Q_R^{b,k},\hat Q_T^{b',k}\rangle/\tau_q$ and run a symmetric InfoNCE with the diagonal labels $\mathbf{y}\!=\!(0,1,\dots,B\!-\!1)$:
\begin{equation}
\mathcal{L}^{\text{q2q}} = \tfrac{1}{2}\!\Big(\mathbb{E}_{w^q}\!\big[\mathrm{CE}(S^k,\mathbf{y})\big] + \mathbb{E}_{w^q}\!\big[\mathrm{CE}((S^k)^\top,\mathbf{y})\big]\Big),
\label{eq:q2q}
\end{equation}
where $\mathbb{E}_{w^q}[\cdot]$ averages over both $b$ and $k$ with weight $w^q_{b,k}\!=\!\tfrac{1}{2}(\tilde s_{R,b}^k\pi_{R,b,k}^{q}\!+\!\tilde s_{T,b}^k\pi_{T,b,k}^{q})\,q_b$, and the per-query saliency $\tilde s_m^k$ is given in Eq.~\eqref{eq:cmq-score}. When $B\!=\!1$, $\mathcal{L}^{\text{q2q}}$ degenerates to $\mathbb{E}_{w^q}[d(\hat Q_R,\hat Q_T)]$.

The training objective balances three contrastive granularities:
\begin{equation}
\mathcal{L}^{\text{ra-qcgcl}} = \alpha\mathcal{L}^{\text{p2p}}+\beta\mathcal{L}^{\text{p2q}}+\gamma\mathcal{L}^{\text{q2q}},
\label{eq:raqcgcl}
\end{equation}
with $\alpha,\beta,\gamma$ balancing the three granularities. The reliability factor concentrates the contrastive signal on tokens and queries whose cross-modal correspondence is supported by image evidence and syncPATC geometry. When patch-level reliability is low, the Q2Q branch provides an additional alignment signal because it operates on semantic query slots rather than token indices.

\section{Experiments}
\subsection{Experimental Settings}
\label{subsec:settings}

All GAAT task variants are initialized, where applicable, from the same UAVMeta-pretrained dual Swin-V2-Base RGB--IR encoder; each benchmark activates the branch or branches matching its input modality \cite{liu2022swinv2}. GAAT(U) attaches UPerNet, whereas GAAT(C) uses a CMX/SegFormer-style multimodal decoder for semantic segmentation; GAAT(D) uses DDQ-DETR for object detection; and GAAT(G) applies a Global MLP head to mean-pooled top-stage tokens for scene classification \cite{xiao2018upernet,zhang2023cmx,xie2021segformer,zhang2023ddq}. ChangeFormer and ChangerEx provide the change-detection heads (GAAT(CF/CX)), a ByteTrack-style tracker is attached to the GAAT detector (GAAT(B)), and GAAT+ThermalGS couples the IR branch to a gsplat-based renderer for 3D reconstruction \cite{bandara2022changeformer,zhang2021bytetrack,kerbl2023gaussian}. Default training and module settings are summarized in Tab.~\ref{tab:hparams}; task-specific input modalities and augmentations follow the corresponding benchmark protocols.

\begin{table}[!t]
\caption{Default training, module, and StateBench configuration. Task-specific input modalities and augmentations follow the corresponding benchmark protocols.}
\label{tab:hparams}
\centering
\scriptsize
\setlength{\tabcolsep}{2.5pt}
\renewcommand{\arraystretch}{1.00}
\resizebox{\columnwidth}{!}{%
\begin{tabular}{@{}c c c@{}}
\hline
\textbf{Group} & \textbf{Symbol} & \textbf{Value} \\
\hline
\multirow{7}{*}{Pretraining}
& Backbone & Dual Swin-V2-Base \\
& Data / crop / epochs & UAVMeta / $256\!\times\!256$ / $12$ \\
& Optimizer / $(\beta_1,\beta_2)$ & AdamW~\cite{loshchilov2019adamw} / $(0.9,0.95)$ \\
& Learning rate / schedule & $1\!\times\!10^{-4}$ / cosine decay \\
& Mask ratio & $0.6$ \\
& Sync. affine: rotation / translation & $\pm20^\circ$ / $\pm24$ px \\
& Sync. affine: scale / shear & $[0.85,1.15]$ / $\pm8^\circ$ \\
\hline
Fine-tuning
& Optimizer / lr / weight decay & AdamW / $1\!\times\!10^{-4}$ / $0.05$ \\
\hline
\multirow{4}{*}{Loss weights}
& $\lambda_r$, $\lambda_t$, $\lambda_c$ & $1.0$ \\
& $\lambda_d$ & $0.25$ \\
& $\lambda_p$ & $0.2$ \\
& $\lambda_{\text{cyc}}$ & $0.1$ \\
\hline
\multirow{5}{*}{syncPATC}
& $s_p$ (encoder stage) & $2$ \\
& $N_q$ (queries) & $8$ \\
& $r$ (neighborhood radius) & $2$ \\
& $\sigma$ (Gaussian bandwidth) & $1.25$ \\
& $\gamma_0,\gamma_1,\gamma_2$ (curriculum) & $1.25,\,0.75,\,0.5$ \\
\hline
\multirow{5}{*}{MG-Sparse-MMA}
& $L$ (layers) & $2$ \\
& $K_s$ (sparse Top-$K$) & $4$ \\
& $r_s$ (neighborhood radius) & $1$ \\
& $\delta_{\max}$ (offset cap) & $1.5$ \\
& $\lambda_1,\lambda_2$ (saliency weights) & $0.5,\,0.5$ \\
\hline
\multirow{3}{*}{RA-QCGCL}
& $\alpha,\beta,\gamma$ & $0.5,\,0.3,\,0.2$ \\
& $\tau_K$ (Top-$K$ ratio) & $0.25$ \\
& $\tau_p,\tau_q$ (temperatures) & $0.1,\,0.07$ \\
\hline
View gap weights
& $\eta_{\text{rot}},\eta_{\text{trans}},\eta_{\text{scale}},\eta_{\text{shear}}$ & $0.35,\,0.35,\,0.2,\,0.1$ \\
\hline
FMCS weights
& $w_v,w_a,w_\omega,w_h$ & $0.35,\,0.25,\,0.25,\,0.15$ \\
\hline
\multirow{2}{*}{MSPA-4D}
& $w_1,w_2,w_3,w_4$ & $0.25,\,0.20,\,0.25,\,0.30$ \\
& $\tau_1,\tau_2,\tau_3,\tau_4$ (tolerances) & $12,\,10,\,12,\,15$ \\
\hline
\end{tabular}%
}
\end{table}

\subsection{Semantic Segmentation}
\label{subsec:seg}

The main text presents KUST4K, an established visible--IR benchmark, and UAVMeta, our in-domain RGB--IR benchmark \cite{ouyang2025kust4k}. KUST4K evaluates transfer under an external annotation protocol, whereas UAVMeta evaluates the same pretrained representation on the official acquisition-disjoint split. We report mIoU and the accuracy measures defined by each benchmark. GAAT(C) uses the geometry-aware encoder with a CMX/SegFormer-style multimodal decoder \cite{zhang2023cmx,xie2021segformer}. The comparison spans convolutional, multispectral-fusion, lightweight, foundation-model, and state-space segmentation families, including U-Net, MFNet, RTFNet, ICAFusion, SGFNet, MobileNetV3, DOFA, and Sigma \cite{ronneberger2015unet,ha2017mfnet,sun2019rtfnet,shen2024icafusion,wang2023sgfnet,howard2019mobilenetv3,xiong2024dofa,wan2025sigma}. Additional segmentation comparisons on UAVid, UDD5, UDD6, and UAVM are reported in Appendix~\ref{subsec:appendix-seg}.

On KUST4K (Tab.~\ref{tab:seg-kust4k}), GAAT(C) obtains 82.14 mIoU and 92.41 mean class accuracy (mAcc), the highest values in the table. The mIoU exceeds SGFNet by 1.07 points and the mAcc exceeds DOFA by 2.93 points, showing transfer of the geometry-aware representation to visible--IR prediction.

\begin{table}[!t]
\caption{Semantic-segmentation results on KUST4K under the official evaluation protocol. Higher values are better.}
\label{tab:seg-kust4k}
\centering
\footnotesize
\renewcommand{\arraystretch}{1.1}
\setlength{\tabcolsep}{1pt}
\begin{tabular}{@{}l l c c c@{}}
\hline
\textbf{Method} & \textbf{Backbone} & \textbf{Input} & \textbf{mIoU} & \textbf{mAcc} \\
\hline
UNet & ResNet-50 & RGB--IR & 52.70 & 22.02 \\
UPerNet & ResNet-50 & RGB--IR & 55.40 & 54.06 \\
RTFNet & ResNet-50 & RGB--IR & 60.00 & 78.82 \\
FEANet & ResNet-50 & RGB--IR & 67.60 & 78.80 \\
EAEFNet & ResNet-50 & RGB--IR & 70.10 & 83.13 \\
CMNeXt-B2 & CMNeXt & RGB--IR & 70.00 & 68.68 \\
CMNeXt-B4 & CMNeXt & RGB--IR & 72.37 & 79.53 \\
CMX-B2 & MiT-B2 & RGB--IR & 72.00 & 76.45 \\
CMX-B4 & MiT-B4 & RGB--IR & 71.71 & 78.90 \\
Sigma & Mamba & RGB--IR & 75.30 & 82.82 \\

SegFormer-RGB & MiT-B5 & RGB & 59.84 & 75.80 \\
DeepLabV3+-RGB & ResNet-50 & RGB & 67.40 & 74.20 \\
SegFormer-RGB & MiT-B5 & RGB & 60.26 & 65.85 \\
PIDNet-RGB & PIDNet-L & RGB & 55.95 & 62.43 \\
SegFormer-IR & MiT-B5 & IR & 63.20 & 77.45 \\
ICAFusion & ICAFusion & RGB--IR & 66.17 & 78.89 \\
SHIFNet (SAM2) & SAM2 & RGB--IR & 65.30 & 74.61 \\
SGFNet & ResNet-50 & RGB--IR & 81.07 & 88.02 \\
FGFNet & ResNet-50 & RGB--IR & 72.12 & 84.51 \\

DOFA & ViT-L & RGB & 80.83 & 89.48 \\
SatMAE++ & ViT-L & RGB & 67.89 & 75.67 \\
Prithvi-EO-2.0 & Prithvi-EO & RGB & 66.10 & 76.18 \\
\hline
\textbf{GAAT(C) (Ours)} & GAAT & \textbf{RGB--IR} & \textbf{82.14} & \textbf{92.41} \\
\hline
\end{tabular}
\end{table}

\textbf{UAVMeta protocol and results.} UAVMeta uses the official acquisition-disjoint split of 1,715/572/288 synchronized RGB--IR pairs for training/validation/test. All seven stored mask classes, including background, are included in the evaluation. Results are the mean and standard deviation over seeds 42--44; checkpoints are selected by validation mIoU and evaluated once on the test set. The reported GAAT configuration uses the common pretrained encoder with multiscale RGB--IR augmentation and a SegFormer-style decoder.

As shown in Tab.~\ref{tab:uavmeta-seg}, GAAT reaches $67.79\pm1.18$ mIoU, $78.58\pm1.32$ frequency-weighted intersection over union (FWIoU), and $87.44\pm0.93$ pixel accuracy. Relative to SegFormer-B5, the strongest listed baseline on all three aggregate metrics, the corresponding mean improvements are 4.27, 4.05, and 2.58 points. GAAT uses paired RGB--IR input, whereas SegFormer-B5 is RGB-only; the ablation study further examines the contribution of the alignment modules.

\begin{table}[!t]
\caption{Semantic-segmentation results on UAVMeta across RGB-only, IR-only, and RGB--IR configurations; input modality is listed in the table.}
\label{tab:uavmeta-seg}
\centering
\footnotesize
\renewcommand{\arraystretch}{1.15}
\setlength{\tabcolsep}{2pt}
\resizebox{\columnwidth}{!}{
\begin{tabular}{@{}l l l c c c@{}}
\hline
\textbf{Method} & \textbf{Backbone/Decoder} & \textbf{Input} &
\textbf{mIoU} $\uparrow$ & \textbf{FWIoU} $\uparrow$ &
\textbf{Pixel Acc.} $\uparrow$ \\
\hline
DeepLabV3 & ResNet-50 & RGB & 57.54\ensuremath{\pm}2.39 & 68.02\ensuremath{\pm}1.71 & 80.35\ensuremath{\pm}1.35 \\
DeepLabV3 & ResNet-50 & IR & 58.91\ensuremath{\pm}0.65 & 70.63\ensuremath{\pm}0.67 & 81.88\ensuremath{\pm}0.59 \\
LR-ASPP & MobileNetV3-L & RGB & 49.24\ensuremath{\pm}1.08 & 60.46\ensuremath{\pm}0.63 & 74.55\ensuremath{\pm}0.56 \\
Global-IR FiLM & ResNet-50 & RGB--IR & 54.88\ensuremath{\pm}1.50 & 64.81\ensuremath{\pm}1.42 & 77.92\ensuremath{\pm}1.10 \\
DeepLabV3 & ResNet-101 & RGB & 57.16\ensuremath{\pm}1.40 & 68.00\ensuremath{\pm}0.98 & 80.05\ensuremath{\pm}0.78 \\
SegFormer-B5 & MiT-B5 & RGB & 63.52\ensuremath{\pm}0.98 & 74.53\ensuremath{\pm}1.17 & 84.86\ensuremath{\pm}0.82 \\
\hline
\textbf{GAAT(C) (Ours)} & GAAT & RGB--IR & \textbf{67.79\ensuremath{\pm}1.18} & \textbf{78.58\ensuremath{\pm}1.32} & \textbf{87.44\ensuremath{\pm}0.93} \\
\hline
\end{tabular}
}
\end{table}

\subsection{Object Detection}
\label{subsec:det}

The main text presents DroneVehicle and UAVMeta. DroneVehicle provides an established cross-modal vehicle-detection benchmark \cite{sun2020dronevehicle}, while UAVMeta measures in-domain transfer on synchronized RGB--IR pairs under an acquisition-disjoint split. We report COCO-style mAP, mAP$_{50}$, and mAP$_{75}$ \cite{lin2014coco} and use DDQ-DETR as the principal detection head for GAAT(D). The listed baselines cover two-stage, one-stage, oriented, and real-time transformer detectors \cite{ren2017fasterrcnn,cai2018cascadercnn,liu2016ssd,lin2017focalloss,redmon2016yolo,wang2023yolov7,ding2019roitransformer,han2022s2anet,xie2021orientedrcnn,tian2019fcos,zhao2024rtdetr}. Additional detection comparisons on HIT-UAV, LLVIP, UAVDT, VisDrone-DET, and IndraEye are reported in Appendix~\ref{subsec:appendix-det}.

On DroneVehicle (Tab.~\ref{tab:det-dronevehicle}), GAAT(D) achieves 56.59 mAP, 80.12 mAP$_{50}$, and 67.16 mAP$_{75}$. Relative to the strongest listed baseline, DDQ-DETR, these values improve by 6.49, 7.72, and 7.66 points, respectively, with gains across all three reported metrics.

\begin{table}[!t]
\caption{Object-detection results on DroneVehicle across RGB, IR, and paired RGB--IR configurations; the input modality is indicated by the modality group in the table.}
\label{tab:det-dronevehicle}
\centering
\footnotesize
\renewcommand{\arraystretch}{1.15}
\begin{tabular}{@{}l l c c c@{}}
\hline
\textbf{Method} & \textbf{Backbone} & \textbf{mAP} & \textbf{mAP$_{50}$} & \textbf{mAP$_{75}$} \\
\hline
\multicolumn{5}{c}{\textit{RGB modality}} \\
\hline
Faster R-CNN & ResNet-101 & 45.00 & 70.10 & 50.50 \\
RoITransformer & ResNet-50 & 38.59 & 65.04 & 40.37 \\
YOLOv7 & EfficientNet-B0 & 47.24 & 71.56 & 53.57 \\
S2ANet & ResNet-50 & 34.35 & 62.04 & 34.51 \\
Oriented R-CNN & ResNet-50 & 38.20 & 64.00 & 41.25 \\
Cascade R-CNN & ResNet-101 & 47.10 & 71.00 & 53.60 \\
\hline
\multicolumn{5}{c}{\textit{IR modality}} \\
\hline
Faster R-CNN & ResNet-101 & 41.70 & 68.30 & 46.00 \\
RoITransformer & ResNet-50 & 41.26 & 65.94 & 47.00 \\
S2ANet & ResNet-50 & 38.91 & 65.76 & 42.16 \\
Oriented R-CNN & ResNet-50 & 40.89 & 66.13 & 46.62 \\
Cascade R-CNN & ResNet-101 & 43.80 & 69.40 & 48.90 \\
\hline
\multicolumn{5}{c}{\textit{RGB--IR modality}} \\
\hline
Cascade R-CNN & ResNet-50 & 45.60 & 69.40 & 51.70 \\
DDQ-DETR & Swin-Base & 50.10 & 72.40 & 59.50 \\
Faster R-CNN & ResNet-50 FPN-v2 & 46.12 & 70.46 & 52.97 \\
\hline
\textbf{GAAT(D) (Ours)} & GAAT & \textbf{56.59} & \textbf{80.12} & \textbf{67.16} \\
\hline
\end{tabular}
\end{table}

\textbf{UAVMeta protocol and results.} The UAVMeta detector is trained and evaluated on the official acquisition-disjoint split, with model selection performed on the validation set. Tab.~\ref{tab:ours-det} reports the mean and standard deviation over three fine-tuning seeds. The reported GAAT(D) configuration combines the common GAAT encoder with a DDQ-DETR multimodal detection head.

GAAT(D) obtains $44.86\pm3.76$ mAP, $62.64\pm4.40$ mAP$_{50}$, and $52.00\pm5.13$ mAP$_{75}$. The corresponding DDQ-DETR baseline means are 43.33, 61.93, and 50.37, yielding absolute improvements of 1.53, 0.71, and 1.63 points. The DDQ-DETR baseline is RGB-only, whereas GAAT(D) uses RGB--IR input.

\begin{table*}[!t]
\caption{Object-detection results on UAVMeta across RGB, IR, and RGB--IR configurations under the official acquisition-disjoint split; input modality is listed in the table.}
\label{tab:ours-det}
\centering
\footnotesize
\renewcommand{\arraystretch}{1.15}
\setlength{\tabcolsep}{4pt}
\begin{tabular}{@{}l l c c c c@{}}
\hline
\textbf{Method} & \textbf{Backbone} & \textbf{Input} & \textbf{mAP} & \textbf{mAP$_{50}$} & \textbf{mAP$_{75}$} \\
\hline
Faster R-CNN & ResNet-50 & RGB & $37.88 \pm 2.39$ & $56.58 \pm 1.71$ & $44.76 \pm 3.63$ \\
Cascade R-CNN & ResNet-50 & RGB & $37.84 \pm 3.61$ & $55.23 \pm 2.78$ & $43.53 \pm 5.34$ \\
YOLOX & YOLOX-X & RGB & $39.56 \pm 6.53$ & $60.21 \pm 3.22$ & $44.23 \pm 9.73$ \\
DDQ-DETR & Swin-Base & RGB & $43.33 \pm 3.67$ & $61.93 \pm 3.91$ & $50.37 \pm 5.52$ \\
DINO & ResNet-50 & RGB & $42.37 \pm 1.99$ & $57.90 \pm 2.33$ & $48.53 \pm 2.25$ \\
Swin-V2 Faster R-CNN & Swin-V2-B & RGB & $22.71 \pm 6.34$ & $44.21 \pm 6.58$ & $20.24 \pm 9.66$ \\
Faster R-CNN & ResNet-50 & IR & $18.95 \pm 0.43$ & $34.68 \pm 1.29$ & $19.33 \pm 1.62$ \\
\hline
\textbf{GAAT(D) (Ours)} & GAAT & RGB--IR & $\mathbf{44.86 \pm 3.76}$ & $\mathbf{62.64 \pm 4.40}$ & $\mathbf{52.00 \pm 5.13}$ \\
\hline
\end{tabular}
\end{table*}

\subsection{Scene Classification}
\label{subsec:cls}

We evaluate scene classification on AID and RESISC45 using the RGB branch and on UAVMeta using paired RGB--IR inputs. A Global MLP classification head is applied to mean-pooled top-stage tokens. For AID and RESISC45, the backbone is fully fine-tuned under the standard training-ratio protocols; AID, RESISC45, Million-AID, and EuroSAT are established aerial or remote-sensing scene-classification benchmarks \cite{xia2017aid,cheng2017resisc45,long2021millionaid,helber2019eurosat}. The comparison includes remote-sensing pretraining approaches such as SeCo, SatMAE, CMID, and RVSA \cite{manas2021seco,cong2022satmae,muhtar2023cmid,wang2023rvsa}. UAVMeta uses the official acquisition-disjoint RGB--IR split and its per-frame scene labels.

On AID (TR=20\%/50\%), the GAAT classification configuration records 95.90/97.38 accuracy; on RESISC45 (TR=10\%/20\%), it records 92.84/94.49. These results show that the pretrained encoder transfers to both scene-classification benchmarks. SkySense is stronger on AID at the 20\% ratio and on both RESISC45 ratios, while RVSA is strongest on AID at the 50\% ratio \cite{guo2024skysense,wang2023rvsa}.

\begin{table}[!t]
\caption{Scene-classification overall accuracy (OA, \%) on AID and RESISC45; OA denotes overall accuracy. GAAT(G) denotes GAAT with the Global MLP classification head applied to mean-pooled top-stage tokens.}
\label{tab:cls}
\centering
\footnotesize
\renewcommand{\arraystretch}{1.12}
\setlength{\tabcolsep}{3pt}
\resizebox{\columnwidth}{!}{%
\begin{tabular}{@{}l l c c@{}}
\hline
\textbf{Method} & \textbf{Backbone} & \textbf{AID (OA)} & \textbf{RESISC45 (OA)} \\
 & & \textbf{TR=20\%/50\%} & \textbf{TR=10\%/20\%} \\
\hline
GASSL & ResNet-50 & 93.55/95.92 & 90.86/93.06 \\
SeCo & ResNet-50 & 93.47/95.99 & 89.64/92.91 \\
SatMAE & ViT-L & 95.02/94.98 & 91.72/91.84 \\
RingMo & Swin-B & 96.90/98.34 & 94.25/95.67 \\
RVSA & ViT-B & 97.03/\textbf{98.50} & 93.93/95.69 \\
TOV & ViT-B & 95.16/97.09 & 90.97/93.79 \\
SSL4EO & ResNet-50 & 91.06/94.74 & 87.60/91.27 \\
CMID & ResNet-50 & 96.11/97.79 & 91.84/94.07 \\
CACo & ResNet-50 & 90.88/95.05 & 88.28/91.94 \\
CROMA & ViT-B & 96.44/97.58 & 92.63/95.04 \\
SatLas & Swin-B & 94.96/97.38 & 92.16/94.70 \\
Scale-MAE & ViT-L & 96.44/97.58 & 92.63/95.04 \\
SkySense & Swin-H & \textbf{97.56}/98.42 & \textbf{94.96/96.20} \\
RingMo-Aerial & RingMo-Aerial & 95.81/96.46 & 92.28/95.65 \\
\hline
\textbf{GAAT(G) (Ours)} & GAAT & 95.90/97.38 & 92.84/94.49 \\
\hline
\end{tabular}%
}
\end{table}

\textbf{UAVMeta protocol and results.} For UAVMeta scene classification, label 0 (\texttt{Void}) is excluded and classes 1--8 are evaluated. Results are the mean and standard deviation over seeds 42--44; checkpoints are selected by validation macro-F1 and evaluated once on the test split. The reported GAAT(G) configuration is initialized from the common pretrained checkpoint and uses the Global MLP head with the IR branch enabled.

Tab.~\ref{tab:uavmeta-cls} shows that GAAT(G) achieves the highest reported mean accuracy ($63.66\pm1.78$) and balanced accuracy ($51.97\pm0.56$), improving over EfficientNetV2-S by 0.23 and 0.72 points, respectively. Its macro-F1 is $48.12\pm1.23$, while the IR-only ResNet-50 baseline reaches $48.72\pm5.85$. The comparison uses the established ResNet, ConvNeXt, and EfficientNetV2 architecture families \cite{he2016resnet,liu2022convnext,tan2021efficientnetv2}. GAAT uses both modalities, whereas the strongest baseline entries use a single modality.

\begin{table}[!t]
\caption{Scene-classification results on UAVMeta across RGB, IR, and RGB--IR configurations; input modality is listed in the table.}
\label{tab:uavmeta-cls}
\centering
\footnotesize
\renewcommand{\arraystretch}{1.15}
\setlength{\tabcolsep}{3.5pt}
\begin{tabular}{@{}l l c c c@{}}
\hline
\textbf{Method} & \textbf{Input} &
\textbf{Acc.} $\uparrow$ & \textbf{Bal. Acc.} $\uparrow$ &
\textbf{Macro-F1} $\uparrow$ \\
\hline
ResNet-50 & RGB & 49.65\ensuremath{\pm}2.50 & 39.25\ensuremath{\pm}0.99 & 37.39\ensuremath{\pm}0.28 \\
ResNet-50 & IR & 62.73\ensuremath{\pm}4.14 & 50.51\ensuremath{\pm}4.97 & \textbf{48.72\ensuremath{\pm}5.85} \\
Swin-T & RGB & 47.80\ensuremath{\pm}0.72 & 37.76\ensuremath{\pm}1.67 & 35.29\ensuremath{\pm}0.77 \\
Late-Fusion ResNet-50 & RGB--IR & 54.40\ensuremath{\pm}1.60 & 44.11\ensuremath{\pm}1.88 & 42.25\ensuremath{\pm}3.47 \\
ConvNeXt-B & IR & 59.95\ensuremath{\pm}4.02 & 50.23\ensuremath{\pm}4.15 & 47.08\ensuremath{\pm}6.19 \\
EfficientNetV2-S & IR & 63.43\ensuremath{\pm}1.91 & 51.25\ensuremath{\pm}0.75 & 47.86\ensuremath{\pm}1.67 \\
\hline
\textbf{GAAT(G) (Ours)} & RGB--IR & \textbf{63.66\ensuremath{\pm}1.78} & \textbf{51.97\ensuremath{\pm}0.56} & 48.12\ensuremath{\pm}1.23 \\
\hline
\end{tabular}
\end{table}

\subsection{Change Detection}
\label{subsec:cd}

We next examine bi-temporal transfer on CDD (Change Detection Dataset) and LEVIR-CD (Large-scale building change detection) \cite{shi2021cdd,chen2021levircd}. The table reports F1 and IoU scores. We use ChangeFormer and ChangerEx as the change-detection heads, denoted GAAT(CF) and GAAT(CX), respectively; the comparison spans fully convolutional Siamese, densely connected Siamese, and transformer-based change detectors \cite{cayedaudt2018fcsiam,fang2022snunetcd,bandara2022changeformer,chen2021bit,codegoni2022tinycd}.

Tab.~\ref{tab:cd} reports the two change-detection datasets in one compact table. On CDD, GAAT(CF) obtains 97.85 F1 and 95.79 IoU, within 0.03 and 0.06 points of ScratchFormer, respectively. On LEVIR-CD, GAAT(CX) achieves the highest listed F1 (95.96) and IoU (92.47), showing that GAAT transfers across both change-detection benchmarks.

\begin{table}[!t]
\caption{Change-detection results on the bi-temporal CDD and LEVIR-CD benchmarks. Higher values are better.}
\label{tab:cd}
\centering
\footnotesize
\renewcommand{\arraystretch}{1.15}
\begin{tabular}{@{}l c c c | c c@{}}
\hline
\multirow{2}{*}{\textbf{Method}} & \multirow{2}{*}{\textbf{Backbone}} & \multicolumn{2}{c}{\textbf{CDD}} & \multicolumn{2}{c}{\textbf{LEVIR-CD}} \\
\cline{3-4} \cline{5-6}
  &  & F1 & IoU & F1 & IoU \\
\hline
ChangeFormer & Transformer & 97.29 & 94.72 & 90.40 & 88.80 \\
BIT & ResNet-18 & 88.90 & 80.01 & 89.31 & 89.37 \\
RingMo & RingMo & 90.90 & 84.09 & 89.53 & 82.38 \\
ScratchFormer & Transformer & \textbf{97.88} & \textbf{95.85} & 91.68 & 84.63 \\
Changer & ResNet-18 & 67.81 & 57.72 & 90.70 & 82.99 \\
TinyCD & CNN & 62.50 & 52.37 & 91.05 & 83.57 \\
Change3D & ResNet-18 & 96.25 & 92.77 & 91.82 & 84.87 \\
GeSANet & ResNet-18 & 95.14 & 90.73 & 90.05 & 83.67 \\
TransUNetCD & ResNet-50 & 97.17 & 94.50 & 91.11 & 83.67 \\
\hline
\textbf{GAAT(CF) (Ours)} & GAAT & 97.85 & 95.79 & 86.24 & 75.81 \\
\textbf{GAAT(CX) (Ours)} & GAAT & 97.77 & 95.64 & \textbf{95.96} & \textbf{92.47} \\
\hline
\end{tabular}
\end{table}

\subsection{Multi-Object Tracking}
\label{subsec:mot}

We then assess temporal association on the official M3OT validation sequences, which contain paired RGB--IR UAV videos \cite{nie2025m3ot}. The metrics are Higher Order Tracking Accuracy (HOTA), Multiple Object Tracking Accuracy (MOTA), and ID F1 Score (IDF1), following the HOTA and CLEAR MOT evaluation frameworks \cite{luiten2021hota,bernardin2008clearmot}. We adopt a ByteTrack-style tracking framework on top of the GAAT detector, denoted GAAT(B). The comparison covers the SORT/Deep SORT association lineage and representative joint detection-and-tracking methods, including CenterTrack, FairMOT, OC-SORT, and Deep OC-SORT \cite{bewley2016sort,wojke2017deepsort,zhang2021bytetrack,zhou2020centertrack,zhang2021fairmot,cao2023ocsort,maggiolino2023deepocsort}.

Tab.~\ref{tab:mot} reports RGB, IR, and pooled M3OT results on the official validation sequences. Drone~1 corresponds to sequence 1-08 and Drone~2 to sequence 2-08. The baseline entries use the backbones listed in the table: CenterTrack and FairMOT use DLA-34 \cite{yu2018dla}, whereas the other baselines use YOLOv8-n; GAAT(B) uses the GAAT-B backbone. GAAT(B) has the highest pooled HOTA, IDF1, and MOTA on Drone~1. On Drone~2 it has the highest pooled HOTA and MOTA, while CenterTrack retains the highest pooled IDF1 (65.78 versus 60.84).

\begin{table*}[!t]
  \caption{Multi-object tracking results on the official M3OT validation sequences. Overall values are pooled RGB--IR summaries computed by the benchmark protocol (reported values are rounded); higher values are better.}
\label{tab:mot}
\centering
\footnotesize
\setlength{\tabcolsep}{0pt}
\renewcommand{\arraystretch}{1.05}
\begin{tabular*}{\textwidth}{@{\extracolsep{\fill}}l l l c c c c c c c c c@{}}
\hline
\multirow{2}{*}{\textbf{UAV}} & \multirow{2}{*}{\textbf{Tracker}} & \multirow{2}{*}{\textbf{Backbone}} & \multicolumn{3}{c}{\textbf{RGB}} & \multicolumn{3}{c}{\textbf{IR}} & \multicolumn{3}{c}{\textbf{Overall}} \\
\cline{4-12}
& & & \textbf{HOTA$\uparrow$} & \textbf{IDF1$\uparrow$} & \textbf{MOTA$\uparrow$} & \textbf{HOTA$\uparrow$} & \textbf{IDF1$\uparrow$} & \textbf{MOTA$\uparrow$} & \textbf{HOTA$\uparrow$} & \textbf{IDF1$\uparrow$} & \textbf{MOTA$\uparrow$} \\
\hline
\multirow{7}{*}{Drone 1}
& CenterTrack & DLA-34 & 39.28 & 56.84 & 25.39 & 42.55 & 60.80 & 24.72 & 40.89 & 58.89 & 25.06 \\
& FairMOT & DLA-34 & 35.46 & 47.78 & 14.66 & 43.23 & 56.51 & 31.22 & 39.30 & 52.10 & 22.84 \\
& ByteTrack & YOLOv8-n & 37.70 & 59.00 & 38.90 & 34.90 & 62.10 & 43.60 & 36.30 & 60.60 & 41.30 \\
& OC-SORT & YOLOv8-n & 36.80 & 54.30 & 34.80 & 31.80 & 55.30 & 35.30 & 34.30 & 54.80 & 35.10 \\
& Deep OC-SORT & YOLOv8-n & 31.10 & 47.70 & 27.80 & 31.70 & 55.80 & 36.20 & 31.40 & 51.80 & 32.00 \\
& ImprAsso & YOLOv8-n & 29.60 & 40.10 & 28.80 & 28.20 & 43.00 & 40.00 & 28.90 & 41.60 & 34.40 \\
& \textbf{GAAT(B) (Ours)} & \textbf{GAAT-B} & \textbf{43.09} & \textbf{62.07} & \textbf{40.50} & \textbf{59.01} & \textbf{73.86} & \textbf{55.11} & \textbf{51.05} & \textbf{67.96} & \textbf{47.80} \\
\hline
\multirow{7}{*}{Drone 2}
& CenterTrack & DLA-34 & 39.18 & \textbf{59.86} & 29.35 & 48.73 & \textbf{71.84} & 49.80 & 43.91 & \textbf{65.78} & 39.47 \\
& FairMOT & DLA-34 & 36.95 & 41.39 & 14.82 & 37.68 & 39.67 & 26.27 & 37.31 & 40.55 & 20.48 \\
& ByteTrack & YOLOv8-n & 36.00 & 54.80 & 36.40 & 43.50 & 62.80 & 51.20 & 39.80 & 58.80 & 43.80 \\
& OC-SORT & YOLOv8-n & 33.90 & 49.70 & 31.80 & 37.70 & 50.10 & 41.90 & 35.80 & 49.90 & 36.90 \\
& Deep OC-SORT & YOLOv8-n & 32.10 & 47.10 & 28.50 & 38.30 & 54.20 & 44.70 & 35.20 & 50.70 & 36.60 \\
& ImprAsso & YOLOv8-n & 28.90 & 37.10 & 29.60 & 40.10 & 48.30 & 48.70 & 34.50 & 42.70 & 39.20 \\
& \textbf{GAAT(B) (Ours)} & \textbf{GAAT-B} & \textbf{43.75} & 55.71 & \textbf{40.26} & \textbf{55.83} & 65.97 & \textbf{58.83} & \textbf{49.79} & 60.84 & \textbf{49.54} \\
\hline
\end{tabular*}
\end{table*}

\subsection{3D Reconstruction}
\label{subsec:3d}

Finally, we assess geometry-sensitive transfer on the IR stream of TSDN (Thermal Scene Dataset for novel-view synthesis) using the SingleTime-AM00 protocol. It measures per-scene reconstruction quality through Peak Signal-to-Noise Ratio (PSNR), Structural Similarity Index (SSIM), and Learned Perceptual Image Patch Similarity (LPIPS) \cite{zhang2018lpips}. We integrate GAAT features with a gsplat-based 3D Gaussian Splatting renderer, denoted GAAT+ThermalGS. The comparison is situated among neural radiance fields, multiresolution hash-encoded neural graphics primitives, tensorial radiance fields, and 3D Gaussian Splatting \cite{mildenhall2020nerf,muller2022instantngp,chen2022tensorf,kerbl2023gaussian}.

Tab.~\ref{tab:3d-tsdn} presents results on the TSDN SingleTime-AM00 setting. GAAT+ThermalGS achieves 26.92 PSNR, 0.89 SSIM, and 0.11 LPIPS. GAAT+ThermalGS delivers the best SSIM and LPIPS, while Thermal3D-GS \cite{chen2024thermal3dgs} and ThermalGS tie for the highest PSNR.

\begin{table}[!t]
\caption{Novel-view synthesis results on TSDN under the SingleTime-AM00 protocol. Higher PSNR and SSIM, and lower LPIPS, are better.}
\label{tab:3d-tsdn}
\centering
\footnotesize
\renewcommand{\arraystretch}{1.15}
\setlength{\tabcolsep}{3pt}
\resizebox{\columnwidth}{!}{%
\begin{tabular}{@{}l l c c c@{}}
\hline
\textbf{Method} & \textbf{Backbone} & \textbf{PSNR} $\uparrow$ & \textbf{SSIM} $\uparrow$ & \textbf{LPIPS} $\downarrow$ \\
\hline
TensoRF & Tensorial RF & 10.10 & 0.64 & 0.62 \\
SfM-Based & SfM/MVS & 23.97 & 0.77 & 0.19 \\
3DGS & 3DGS & 27.16 & 0.83 & 0.18 \\
GaussianShader & 3DGS & 23.43 & 0.77 & 0.25 \\
Thermal3D-GS & 3DGS & \textbf{27.21} & 0.84 & 0.17 \\
ThermalGS & 3DGS & \textbf{27.21} & 0.82 & 0.13 \\
\hline
\textbf{GAAT+ThermalGS} & \textbf{GAAT} & 26.92 & \textbf{0.89} & \textbf{0.11} \\
\hline
\end{tabular}%
}
\end{table}

\subsection{StateBench Evaluation on UAVMeta}
\label{subsec:ours-eval}

StateBench complements the supervised UAVMeta tasks by evaluating acquisition-state prediction from visual input.

We benchmark single-frame state prediction from synchronized RGB--IR pairs against generic visual backbones and the public SkySense foundation baselines. The four prediction targets are the acquisition-state scores defined in Sec.~\ref{subsec:state-metrics}; models receive the paired images, while telemetry fields and precomputed image statistics are used to construct the targets. The reported GAAT configuration uses a frozen GAAT predictor; the optional metadata prompt $\phi_v$ is disabled. Tab.~\ref{tab:mspa} reports the mean over optimization seeds 42--44 on a fixed clip split shared by all methods. Temporal neighbours are not used.

GAAT obtains an aggregate MSPA-4D score of 52.69, close to SkySense-S2 (52.26), and the lowest FMCS MAE (12.66). SkySense-S2 is best on CARS MAE, SkySense-HR on OSGS MAE, and ViT-Large on VSS MAE.

\begin{table*}[!t]
\caption{Single-frame acquisition-state prediction results on UAVMeta StateBench. All methods use the synchronized paired RGB--IR input; no metadata or temporal neighbours are provided as model inputs.}
\label{tab:mspa}
\centering
\footnotesize
\renewcommand{\arraystretch}{1.15}
\setlength{\tabcolsep}{4pt}
\begin{tabular}{@{}l l c c c c c@{}}
\hline
\textbf{Method} & \textbf{Backbone} & MSPA-4D$\uparrow$ & CARS MAE$\downarrow$ & OSGS MAE$\downarrow$ & VSS MAE$\downarrow$ & FMCS MAE$\downarrow$ \\
\hline
ResNet-50           & ResNet-50      & 42.03 & 3.57 & 22.19 & 12.05 & 13.53 \\
ResNet-101          & ResNet-101     & 36.33 & 7.63 & 20.75 & 9.05 & 17.75 \\
ViT-Base            & ViT-B/16       & 40.13 & 7.30 & 20.53 & 7.68 & 17.68 \\
ViT-Large           & ViT-L/16       & 40.83 & 7.33 & 20.65 & \textbf{7.61} & 17.72 \\
Swin-Base           & Swin-B         & 39.85 & 7.24 & 20.55 & 7.81 & 17.68 \\
Swin-Large          & Swin-L         & 40.66 & 7.21 & 20.55 & 7.88 & 17.68 \\
Swin-V2             & Swin-V2-B      & 42.27 & 7.34 & 20.59 & 9.06 & 13.24 \\
SkySense-HR         & Swin-H         & 45.82 & 4.16 & \textbf{20.45} & 7.88 & 15.97 \\
SkySense-S2         & ViT-L/4        & 52.26 & \textbf{2.24} & 20.70 & 8.16 & 12.92 \\
\hline
\textbf{GAAT (Ours)} & GAAT & \textbf{52.69} & 2.39 & 21.04 & 8.05 & \textbf{12.66} \\
\hline
\end{tabular}
\end{table*}

\subsection{Ablation Study}
\label{subsec:ablation}

We isolate the contribution of each component on downstream benchmarks spanning segmentation, detection, classification, change detection, and tracking. The default configuration corresponds to the full GAAT pipeline; variants are obtained by removing or replacing individual components.

Here, ``Baseline'' denotes the common pretrained encoder and task heads with the three alignment modules disabled; the benchmark-specific input modalities and optimization protocols remain unchanged.

We compare the baseline, the three module ablations (without syncPATC, MG-Sparse-MMA, or RA-QCGCL), and the full configuration while keeping the remaining components and pretraining schedule unchanged. Tab.~\ref{tab:abl-modules} reports these rows. The full configuration gives the strongest overall performance across the representative tasks. The individual removals reveal the roles of syncPATC, MG-Sparse-MMA, and RA-QCGCL.

\begin{table*}[!t]
\caption{Ablation of GAAT components across representative downstream tasks. Higher is better for all metrics. Benchmark-specific protocols, input modalities, and training ratios follow the corresponding main-text definitions; M3OT ``Overall'' values use the benchmark's pooled RGB--IR protocol.}
\label{tab:abl-modules}
\centering
\scriptsize
\renewcommand{\arraystretch}{1.35}
\setlength{\tabcolsep}{1.8pt}
\begin{tabularx}{\textwidth}{@{}l|YYY|YYY|YY|YY|YY@{}}
\hline
\multirow{3}{*}{\textbf{Variant}}
& \multicolumn{3}{c|}{\textbf{Semantic Segmentation}}
& \multicolumn{3}{c|}{\textbf{Object Detection}}
& \multicolumn{2}{c|}{\textbf{Scene Classification}}
& \multicolumn{2}{c|}{\textbf{Change Detection}}
& \multicolumn{2}{c}{\textbf{Multi-Object Tracking}} \\
\cline{2-13}
& \hcell{UAVid}{mIoU}
& \hcell{UDD6}{mIoU}
& \hcell{KUST4K}{mIoU}
& \hcell{DroneVeh.}{mAP$_{50}$}
& \hcell{VisDrone}{mAP$_{50}$}
& \hcell{UAVDT}{mAP$_{50}$}
& \hcell{AID}{Acc}
& \hcell{RESISC45}{Acc}
& \hcell{LEVIR-CD}{F1}
& \hcell{CDD}{F1}
& \hcell{M3OT}{HOTA}
& \hcell{M3OT}{IDF1} \\
\hline
Baseline            & 55.40 & 67.98 & 80.19 & 64.72 & 53.63 & 79.65 & 97.32 & 94.37 & 85.47 & 92.07 & 49.77 & 61.03 \\
~~w/o syncPATC      & 67.47 & 77.11 & 82.07 & 78.94 & 56.60 & 77.40 & 95.25 & 92.43 & 83.19 & 93.37 & 48.85 & 58.92 \\
~~w/o MG-Sparse-MMA & 55.55 & 68.45 & 80.09 & 53.05 & 57.16 & 79.85 & 96.60 & 94.49 & 85.62 & 93.39 & 48.82 & 58.50 \\
~~w/o RA-QCGCL      & 55.08 & 68.67 & 80.60 & 74.18 & 53.63 & 81.31 & 97.19 & 94.49 & 85.56 & 95.30 & 44.84 & 54.10 \\
\textbf{Full GAAT}  & 69.98 & 78.71 & 82.14 & 80.12 & 66.75 & 84.26 & 97.48 & 94.49 & 95.96 & 97.79 & 50.42 & 64.40 \\
\hline
\end{tabularx}
\end{table*}

For the reported configuration, the syncPATC priors $\pi_m^{\text{tok}}, \pi_{m,k}^q$ are detached for RA-QCGCL and retained for MG-Sparse-MMA. We sweep the sparse-query count $K_s\!\in\!\{2,4,8\}$, query count $N_q\!\in\!\{4,8,16\}$, neighborhood radius $r\!\in\!\{1,2,3\}$, and reliable-token ratio $\tau_K\!\in\!\{0.10,0.25,0.50\}$. The default settings are $K_s\!=\!4$, $N_q\!=\!8$, $r\!=\!2$, and $\tau_K\!=\!0.25$.

\subsection{Qualitative Analysis}
\label{subsec:qualitative}

\begin{figure*}[!t]
  \centering
  \includegraphics[width=\textwidth]{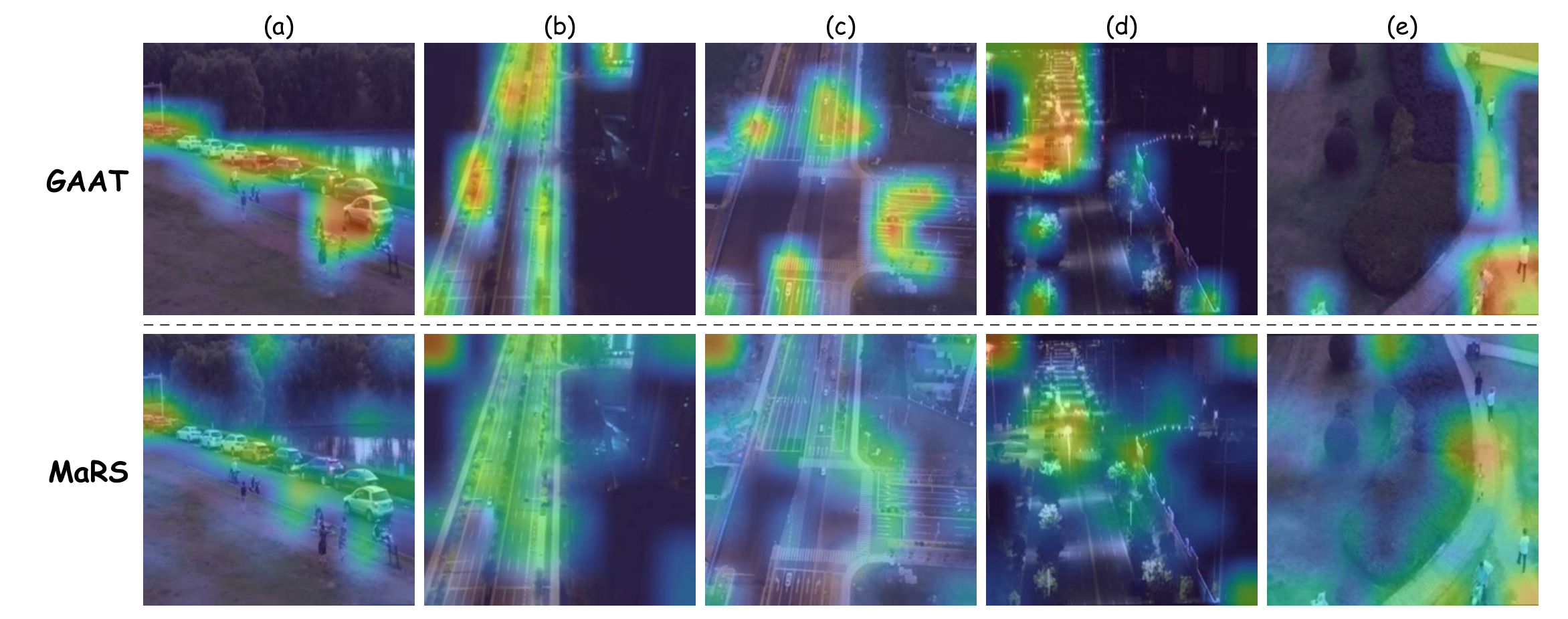}
  \caption{Feature-response heatmaps for GAAT (top) and MaRS (bottom) across five UAV scenes (a)--(e); brighter colors indicate stronger responses. Each column shows the same scene for both models, so the two rows can be compared directly. GAAT responses stay compact around vehicles, road structures, and pedestrians, while MaRS spreads activation across larger background regions.}
  \label{fig:gaat-attention}
\end{figure*}

Fig.~\ref{fig:gaat-attention} compares GAAT and MaRS heatmaps on five representative UAV scenes. In these five examples, GAAT concentrates responses near vehicles, road structures, and pedestrians, whereas MaRS activates broader regions. This pattern follows the top-$K_s$ query routing and geometry-calibrated local sampling of MG-Sparse-MMA, with the syncPATC offset prior $\bar o_m^k$ refining the sampling locations. The compactness is expected from the reliable-token ratio $\tau_K$, which keeps only the top quarter of patches before fusion, so query updates stay confined to the selected neighborhoods while RA-QCGCL supplies query-level alignment where patch-level correspondence is weak. These maps are qualitative, and they are consistent with Table~\ref{tab:abl-modules}, where removing syncPATC or MG-Sparse-MMA lowers detection accuracy on DroneVehicle and VisDrone.

\section{Discussion}
\subsection{Cross-Task Impact of Geometry-Aware Alignment}
GAAT shows that local geometric reliability can serve as a shared organizing principle for multimodal UAV representation learning. Across six primary downstream tasks, GAAT achieves state-of-the-art results on multiple benchmarks and remains competitive in settings where geometry is less central. The strongest performance pattern appears in segmentation, detection, change detection, and tracking, where predictions depend directly on spatially precise features. Strong results on both established external benchmarks and the acquisition-disjoint UAVMeta split show that the gains are not confined to one dataset or task head. This breadth supports GAAT as a transferable multimodal pretrained model rather than a task-specific fusion architecture.

The ablation results in Tab.~\ref{tab:abl-modules} further support the alignment-first design. The full model provides the strongest overall performance pattern across the representative tasks, while removing any core module weakens transfer on several benchmarks. syncPATC supplies transformation-consistent reliability priors before cross-modal interaction. MG-Sparse-MMA converts these priors into geometry-calibrated sparse updates, and RA-QCGCL aligns the supervision with the same patch-and-query structure. Their complementary behavior indicates that reliability estimation, sparse fusion, and cross-granularity supervision operate as one coordinated pipeline. This ablation pattern supports the joint contribution of the three components without attributing the gains to any one module in isolation.

\subsection{Interpreting Geometry-Aware Transfer}
The cross-task pattern also clarifies when geometry-aware pretraining is most valuable. Dense prediction and association tasks require local structures to remain spatially coherent across modalities, so incorrect same-index interaction can directly corrupt their predictions. Global recognition provides complementary evidence of transferable semantics, while the particularly strong gains on spatially sensitive tasks match the intended geometric mechanism. This pattern is consistent with the intended role of prioritizing reliable local interaction.

UAV platforms also introduce viewpoint, scale, and appearance changes that vary substantially across acquisitions. The view-gap curriculum exposes syncPATC to progressively stronger affine perturbations, while the HFE blocks preserve RGB texture and IR thermal contrast before fusion. Evaluations across daytime and nighttime imagery, external datasets, and single-branch transfer settings indicate that the learned representation remains effective beyond the paired pretraining configuration. Together, the broad benchmark comparisons and matched ablations support both the competitive reach of GAAT and the contribution of its alignment modules.

\subsection{Role of UAVMeta and StateBench}
Most UAV benchmarks evaluate what a model predicts without describing the sensing conditions under which the prediction was produced. UAVMeta and StateBench extend this evaluation space by coupling paired RGB--IR perception with acquisition-state prediction. UAVMeta provides one acquisition domain for segmentation, detection, and classification, while StateBench exposes proxy scores related to camera reliability, observation scale, viewpoint stability, and maneuver complexity as explicit targets. GAAT attains the highest reported aggregate MSPA-4D score and the lowest FMCS error, indicating that its representation can predict complementary acquisition-state proxies in addition to scene content.

StateBench complements downstream task metrics with aggregate and per-state views of the acquisition process. This formulation creates a basis for condition-stratified evaluation, failure analysis, and future adaptation policies that respond to sensing quality or flight state. In this sense, UAVMeta and StateBench connect multimodal representation learning with the physical acquisition process that generates UAV imagery.

\subsection{Broader Scope and Outlook}
The present study focuses on paired RGB--IR data, which provide a representative setting for complementary UAV sensing under residual local misalignment. The alignment-first formulation is not tied to a specific downstream task, and the UAVMeta schema can accommodate SAR, depth, multispectral imagery, and other sensing streams. Extending GAAT to these modalities and to more complex geometric variation is a natural next step. The broad benchmark results and matched ablations position geometry-aware alignment as a promising basis for scalable multimodal UAV perception.

\section{Conclusion}
We presented GAAT (Geometry-Aware Alignment Transformer), an alignment-first framework that coordinates patch-center reliability estimation, geometry-guided sparse fusion, and cross-granularity contrastive supervision. Across six primary downstream tasks and multiple UAV datasets, GAAT demonstrates strong multi-task transfer, attains state-of-the-art results on multiple benchmarks, and remains competitive in other task-specific settings. We also introduced UAVMeta for paired RGB--IR pretraining and multi-task evaluation and StateBench for predicting four acquisition-state proxy scores related to camera reliability, observation scale, viewpoint stability, and flight maneuver complexity; GAAT achieves the highest reported aggregate MSPA-4D score. Although current validation centers on paired RGB--IR data and affine-supervised reliability, these results support GAAT as a broadly transferable framework for geometry-aware and acquisition-aware aerial perception.

\appendices

\section{Additional Experimental Results}
\label{sec:appendix}

\subsection{Crowd Counting on VisDrone-CC}
\label{subsec:appendix-crowd}

Crowd counting estimates the number of people or objects in densely populated scenes. We evaluate GAAT on VisDrone-CC, a UAV-based crowd counting benchmark from the VisDrone Challenge 2020 \cite{du2020visdronecc}, following prior low-altitude UAV counting protocols \cite{ptak2022uavcounting}. We report the protocol's MAE and MSE metrics, for which lower values are better, using the benchmark input stream and a CountReg regression head on top of the GAAT backbone.

Tab.~\ref{tab:crowd-visdrone} compares GAAT with the official VisDrone-CC baselines and representative density-regression, distribution-matching, transformer, and point-based counting methods \cite{li2018csrnet,ma2019bayesian,wang2020dmcount,liang2022transcrowd,song2021p2pnet}. GAAT CountReg records the lowest listed MAE (9.40) and MSE (14.18), indicating improved counting accuracy under this evaluation protocol.

\begin{table}[!h]
\caption{Crowd-counting results on VisDrone-CC. Lower MAE and MSE are better.}
\label{tab:crowd-visdrone}
\centering
\renewcommand{\arraystretch}{1.15}
\resizebox{\columnwidth}{!}{%
\begin{tabular}{l l c c}
\hline
\textbf{Protocol} & \textbf{Method} & \textbf{MAE} $\downarrow$ & \textbf{MSE} $\downarrow$ \\
\hline
\multicolumn{4}{c}{\textit{Official VisDrone-CC 2020 Challenge}} \\
\hline
Official CC2020 & FPNCC & 11.66 & 15.45 \\
Official CC2020 & BVCC & 12.36 & 17.32 \\
Official CC2020 & CFF & 13.64 & 16.59 \\
\hline
\multicolumn{4}{c}{\textit{Prior Work}} \\
\hline
Density-based & CSRNet & 14.32 & 18.67 \\
Density-based & BL & 13.89 & 18.21 \\
Density-based & DM-Count & 12.24 & 16.89 \\
Transformer-based & TransCrowd & 11.58 & 16.34 \\
Point-based & P2PNet & 11.12 & 15.78 \\
\hline
\multicolumn{4}{c}{\textit{Ours}} \\
\hline
\textbf{GAAT} & \textbf{GAAT CountReg} & \textbf{9.40} & \textbf{14.18} \\
\hline
\end{tabular}%
}
\end{table}

\subsection{Additional Semantic Segmentation Results}
\label{subsec:appendix-seg}

Additional segmentation comparisons are reported in Tabs.~\ref{tab:seg-uavid}--\ref{tab:seg-uavm}. Here, mAcc denotes mean class accuracy, aAcc denotes overall pixel accuracy, and mPA denotes mean pixel accuracy. Results are interpreted within each benchmark rather than pooled across datasets. The cited segmentation families include FCN, U-Net, PSPNet, DeepLabV3+, DANet, UPerNet, OCRNet, SETR, Segmenter, SegFormer, UNetFormer, PIDNet, and BiSeNetV2 \cite{long2015fcn,ronneberger2015unet,zhao2017pspnet,chen2018deeplabv3plus,fu2019danet,xiao2018upernet,yuan2020ocrnet,zheng2021setr,strudel2021segmenter,xie2021segformer,wang2022unetformer,xu2023pidnet,yu2021bisenetv2}; UAVid and the remote-sensing foundation-model baselines are cited separately \cite{lyu2020uavid,sun2023ringmo,diao2025ringmoaerial,szwarcman2025prithvi}.

\begin{table}[!t]
\caption{Semantic-segmentation results on UAVid. Higher mAcc and mIoU are better.}
\label{tab:seg-uavid}
\centering
\renewcommand{\arraystretch}{1.15}
\resizebox{\columnwidth}{!}{%
\begin{tabular}{@{}l l c c@{}}
\hline
\textbf{Method} & \textbf{Backbone} & \textbf{mAcc} & \textbf{mIoU} \\
\hline
DeepLabV3+ & Xception-65 & 61.60 & 48.70 \\
DANet & ResNet-101 & 61.80 & 48.80 \\
ACNet & ResNet-101 & 62.10 & 48.90 \\
OCRNet & HRNetV2-W48 & 61.90 & 48.90 \\
SETR & ViT-Large & 61.00 & 47.80 \\
SegFormer & MiT-B5 & 66.65 & 55.24 \\
ViT-Adapter & ViT-Adapter-B & 66.61 & 55.15 \\
PIDNet & PIDNet-L & 53.21 & 40.94 \\
DOFA & ViT-B & 61.82 & 49.21 \\
CSWin & CSWin & 64.00 & 51.80 \\
UAVFormer & Swin-Base & 65.10 & 53.20 \\
BiSeNetV2 & BiSeNetV2 & 71.35 & 61.64 \\
Segmenter & ViT & 64.89 & 55.48 \\
CoaT & CoaT & 76.74 & 67.55 \\
BANet & BANet & 60.87 & 43.28 \\
UNetFormer & UNetFormer & 70.69 & 60.61 \\
RingMo & RingMo + UPerNet & 71.60 & 61.90 \\

RingMo-Aerial(U) & RingMo-Aerial & 76.50 & 66.40 \\
RingMo-Aerial(A) & RingMo-Aerial & 74.30 & 64.40 \\
RingMo-Aerial(M) & RingMo-Aerial & 78.70 & 68.70 \\
\hline
\textbf{GAAT(U) (Ours)} & GAAT & \textbf{82.31} & \textbf{69.98} \\
\hline
\end{tabular}%
}%
\end{table}

On UAVid, Tab.~\ref{tab:seg-uavid} shows that GAAT(U) reaches the highest listed mAcc (82.31) and mIoU (69.98). These values exceed RingMo-Aerial(M), the strongest listed comparator on both metrics, by 3.61 and 1.28 points, respectively.

\begin{table}[!t]
\caption{Semantic-segmentation results on UDD5. Higher mIoU and aAcc are better.}
\label{tab:seg-udd5}
\centering
\renewcommand{\arraystretch}{1.15}
\resizebox{\columnwidth}{!}{%
\begin{tabular}{l l c c}
\hline
\textbf{Method} & \textbf{Backbone} & \textbf{mIoU} & \textbf{aAcc} \\
\hline
GCN & ResNet-50 & 73.17 & 88.57 \\
ENet & ENet & 71.38 & 88.14 \\
UPerNet & Swin-Base & 73.12 & 89.45 \\
DeepLabV3+ & ResNet-101 & 70.55 & 88.33 \\
SegFormer & MiT-B5 & 68.36 & 88.67 \\
ViT-Adapter & ViT-Adapter-L & 68.85 & 87.93 \\
PIDNet & PIDNet-L & 53.97 & 81.48 \\
DOFA & ViT-L & 75.46 & 89.59 \\
SatMAE++ & ViT-L & 74.89 & 89.35 \\
Prithvi-EO-2.0 & Prithvi-EO & 75.36 & 89.63 \\

\hline
\textbf{GAAT(U) (Ours)} & GAAT & \textbf{77.17} & \textbf{90.31} \\
\hline
\end{tabular}%
}
\end{table}

On UDD5, GAAT(U) obtains the highest listed mIoU (77.17) and aAcc (90.31) in Tab.~\ref{tab:seg-udd5}. The margins over the strongest listed comparator for each metric are 1.71 points in mIoU and 0.68 points in aAcc.

\begin{table}[!t]
\caption{Semantic-segmentation results on UDD6. Higher mIoU and aAcc are better.}
\label{tab:seg-udd6}
\centering
\renewcommand{\arraystretch}{1.15}
\resizebox{\columnwidth}{!}{%
\begin{tabular}{l l c c}
\hline
\textbf{Method} & \textbf{Backbone} & \textbf{mIoU} & \textbf{aAcc} \\
\hline
DeepLabV3+ & ResNet-50 & 69.01 & 85.63 \\
SegFormer & MiT-B5 & 68.47 & 86.93 \\
UPerNet & Swin-Base & 74.65 & 88.45 \\
ViT-Adapter & ViT-Adapter-L & 68.13 & 85.19 \\
PIDNet & PIDNet-L & 52.82 & 77.25 \\
DOFA & ViT-L & 74.96 & 87.92 \\
SatMAE++ & ViT-L & 74.61 & 87.82 \\
Prithvi-EO-2.0 & Prithvi-EO & 75.39 & 88.37 \\
RingMo(U) & RingMo & 75.30 & 87.40 \\
RingMo-Aerial(U) & RingMo-Aerial & 78.50 & 89.40 \\
RingMo-Aerial(M) & RingMo-Aerial & \textbf{78.90} & 89.70 \\

\hline
\textbf{GAAT(U) (Ours)} & GAAT & 78.71 & \textbf{89.95} \\
\hline
\end{tabular}%
}
\end{table}

On UDD6, Tab.~\ref{tab:seg-udd6} shows that GAAT(U) achieves the highest listed aAcc (89.95), while its mIoU of 78.71 is 0.19 points below RingMo-Aerial(M).

\begin{table}[!t]
\caption{Semantic-segmentation results on UAVM. Higher mIoU, mPA, and mF1 are better.}
\label{tab:seg-uavm}
\centering
\footnotesize
\renewcommand{\arraystretch}{1.15}
\setlength{\tabcolsep}{3pt}
\begin{tabular}{@{}l l c c c@{}}
\hline
\textbf{Type} & \textbf{Method} & \textbf{mIoU} & \textbf{mPA} & \textbf{mF1} \\
\hline
\multicolumn{5}{c}{\textit{Generic Semantic Segmentation}} \\
\hline
Generic SS & UPerNet & 57.24 & 63.71 & 64.32 \\
Generic SS & DeepLabV3+ & 62.33 & 68.99 & 67.49 \\
Generic SS & FCN & 50.03 & 55.21 & 56.87 \\
Generic SS & PSPNet & 52.17 & 58.45 & 57.41 \\
\hline
\multicolumn{5}{c}{\textit{RGB+Multispectral Fusion}} \\
\hline
RGB+MS & MFNet & 63.32 & 68.54 & 71.69 \\
RGB+MS & EGFNet & 54.56 & 64.10 & 66.55 \\
RGB+MS & RTFNet & 67.32 & 74.38 & 74.62 \\
RGB+MS & PSTNet & 57.40 & 63.79 & 64.22 \\
\hline
\multicolumn{5}{c}{\textit{Hyperspectral/Remote Sensing}} \\
\hline
HSI/RS & HyLITE & 38.98 & 44.99 & 43.85 \\
HSI/RS & FreeNet & 56.62 & 61.70 & 60.74 \\
HSI/RS & UNetFormer & 67.52 & 74.21 & 71.49 \\
HSI/RS & ClusterFormer & 62.60 & 68.43 & 70.71 \\
\hline
\multicolumn{5}{c}{\textit{Multispectral Segmentation}} \\
\hline
MSS & UAVMNet & 69.31 & 77.23 & 80.61 \\
MSS & UAVMNet & 79.50 & 85.86 & 87.95 \\
\hline
\textbf{Ours} & \textbf{GAAT(C)} & \textbf{87.08} & \textbf{93.70} & \textbf{92.75} \\
\hline
\end{tabular}
\end{table}

On UAVM, GAAT(C) records the highest listed mIoU (87.08), mPA (93.70), and mF1 (92.75) in Tab.~\ref{tab:seg-uavm}. These values exceed the strongest listed non-GAAT results by 7.58, 7.84, and 4.80 points, respectively.

\subsection{Additional Object Detection Results}
\label{subsec:appendix-det}

Tabs.~\ref{tab:det-hituav}--\ref{tab:det-indraeye-expanded-backbone-v2} report additional detection comparisons on HIT-UAV, LLVIP, UAVDT, VisDrone-DET, and IndraEye. HIT-UAV, LLVIP, UAVDT, and VisDrone are established UAV or visible--infrared detection benchmarks \cite{suo2023hituav,jia2021llvip,du2018uavdt,zhu2021visdrone}; the cited detector families include DETR, DINO, Deformable DETR, DDQ-DETR, and visible--thermal fusion baselines such as ICAFusion and DIVFusion \cite{carion2020detr,zhang2022dino,zhu2020deformabledetr,zhang2023ddq,ying2025visiblethermal,zhu2025wavemamba,shen2024icafusion,tang2023divfusion}.

\begin{table}[!t]
\caption{Object-detection results on HIT-UAV. Higher mAP, mAP$_{50}$, and mAP$_{75}$ are better.}
\label{tab:det-hituav}
\centering
\footnotesize
\renewcommand{\arraystretch}{1.15}
\begin{tabular}{@{}l l c c c@{}}
\hline
\textbf{Method} & \textbf{Backbone} & \textbf{mAP} & \textbf{mAP$_{50}$} & \textbf{mAP$_{75}$} \\
\hline
SSD-512 & VGG16 & 43.40 & 75.20 & 43.80 \\ 
Faster R-CNN & ResNet-101 & 37.50 & 66.70 & 37.70 \\
YOLOv4-tiny & CSPDarknet-tiny & 52.40 & 81.56 & 58.54 \\
RT-DETR & ResNet-50 & 26.07 & 51.50 & 22.88 \\
CSFPR-RTDETR & CSFPR-RTDETR & 42.42 & 83.10 & 40.84 \\
DDQ-DETR & Swin-Base & 51.50 & 83.10 & 57.10 \\
Faster R-CNN & ResNet-50 FPN-v2 & 54.20 & 82.23 & 60.25 \\
\hline
\textbf{GAAT(D) (Ours)} & \textbf{GAAT} & \textbf{56.40} & \textbf{87.01} & \textbf{60.48} \\
\hline
\end{tabular}
\end{table}

On HIT-UAV, GAAT(D) achieves the highest listed mAP (56.40), mAP$_{50}$ (87.01), and mAP$_{75}$ (60.48) in Tab.~\ref{tab:det-hituav}. The corresponding margins over the strongest listed comparator for each metric are 2.20, 3.91, and 0.23 points.

\begin{table}[!t]
\caption{Object-detection results on LLVIP validation set. Higher mAP and mAP$_{50}$ are better.}
\label{tab:det-llvip}
\centering
\footnotesize
\renewcommand{\arraystretch}{1.15}
\begin{tabular}{@{}l l c c@{}}
\hline
\textbf{Method} & \textbf{Backbone} & \textbf{mAP (val)} & \textbf{mAP$_{50}$ (val)} \\
\hline
YOLOv8l-IR & YOLOv8 & 62.10 & 95.20 \\
Text-IF & Transformer & 60.20 & 94.10 \\
ICAFusion & ICAFusion & 60.10 & 95.20 \\
YOLOv8l-RGB & YOLOv8 & 54.00 & 91.90 \\
DIVFusion & YOLOv5 & 52.00 & 89.80 \\
FMCAF & YOLOv11 & 59.51 & 92.08 \\
DIVFusion & YOLOv5 & 52.00 & 89.80 \\
Faster R-CNN & ResNet-50 FPN-v2 & 50.45 & 92.59 \\
\hline
\textbf{GAAT(D) (Ours)} & GAAT & \textbf{63.60} & \textbf{96.15} \\
\hline
\end{tabular}
\end{table}

On LLVIP, Tab.~\ref{tab:det-llvip} shows that GAAT(D) obtains the highest listed validation mAP (63.60) and mAP$_{50}$ (96.15). These values are 1.50 and 0.95 points above the strongest listed comparator for the respective metric.

\begin{table}[!t]
\caption{Object-detection results on UAVDT. Higher mAP, mAP$_{50}$, and mAP$_{75}$ are better.}
\label{tab:det-uavdt}
\centering
\footnotesize
\renewcommand{\arraystretch}{1.15}
\begin{tabular}{@{}l l c c c@{}}
\hline
\textbf{Method} & \textbf{Backbone} & \textbf{mAP} & \textbf{mAP$_{50}$} & \textbf{mAP$_{75}$} \\
\hline
MCR-UOD & YOLOv8 & 31.40 & 44.70 & 35.60 \\
SPAR & YOLOv8 & 30.50 & 43.90 & 34.70 \\
EVORL & ResNet-50 & 28.00 & 43.80 & 31.50 \\ 
TPH-YOLOv5 & YOLOv5 & 26.90 & 41.30 & 32.70 \\
Cascade R-CNN & ResNet-50 & 17.10 & 30.50 & 18.60 \\
Faster R-CNN & ResNet-50 & 12.10 & 23.50 & 10.80 \\ 
DDQ-DETR & Swin-Base & 39.30 & 68.80 & 41.80 \\

\hline
\textbf{GAAT(D) (Ours)} & GAAT & \textbf{54.93} & \textbf{84.26} & \textbf{66.04} \\
\hline
\end{tabular}
\end{table}

On UAVDT, GAAT(D) records the highest listed mAP (54.93), mAP$_{50}$ (84.26), and mAP$_{75}$ (66.04) in Tab.~\ref{tab:det-uavdt}. Relative to DDQ-DETR, the strongest listed comparator on all three metrics, the gains are 15.63, 15.46, and 24.24 points, respectively.

\begin{table}[!t]
\caption{Object-detection results on VisDrone-DET. Higher mAP, mAP$_{50}$, and mAP$_{75}$ are better.}
\label{tab:det-visdrone}
\centering
\footnotesize
\renewcommand{\arraystretch}{1.15}
\begin{tabular}{@{}l l c c c@{}}
\hline
\textbf{Method} & \textbf{Backbone} & \textbf{mAP} & \textbf{mAP$_{50}$} & \textbf{mAP$_{75}$} \\
\hline
Light-RCNN & ResNet-50 & 16.50 & 32.80 & 15.10 \\
CornerNet & ResNet-50 & 17.40 & 34.10 & 15.80 \\
RetinaNet & Swin-Base & 22.60 & 38.10 & 23.50 \\
Cascade R-CNN & Swin-Base & 28.40 & 45.20 & 30.10 \\
CEASC & ResNet-18 & 28.70 & 50.70 & 28.40 \\
GLSAN & ResNet-50 & 25.80 & 51.50 & 22.90 \\
QueryDet & ResNet-50 & 28.30 & 48.10 & 28.70 \\
VistrongerDet & ResNet-50 & 33.85 & 57.27 & 34.81 \\
DMNet & ResNet-101 & 29.40 & 49.30 & 30.60 \\
HRDNet & ResNet-101 & 28.30 & 49.20 & 28.10 \\
ClusDet & ResNeXt-101 & 28.40 & 53.20 & 26.40 \\
SDPDet & ResNet-101 & 34.20 & 57.80 & 34.90 \\
AMRNet & ResNet-101 & 31.70 & 52.60 & 33.00 \\
OGMN & ResNeXt-101 & 35.00 & 59.70 & 35.80 \\
DDQ-DETR & Swin-Base & 36.10 & 57.10 & 37.20 \\
RingMo-Aerial (C) & RingMo-Aerial & 31.10 & 51.80 & 32.50 \\
RingMo-Aerial (A) & RingMo-Aerial & 31.00 & 51.40 & 32.10 \\
RingMo-Aerial (D) & RingMo-Aerial & 38.60 & 63.30 & 39.40 \\
\hline
\textbf{GAAT(D) (Ours)} & GAAT & \textbf{41.14} & \textbf{66.75} & \textbf{42.64} \\
\hline
\end{tabular}
\end{table}

On VisDrone-DET, Tab.~\ref{tab:det-visdrone} shows that GAAT(D) achieves the highest listed mAP (41.14), mAP$_{50}$ (66.75), and mAP$_{75}$ (42.64). The corresponding improvements over RingMo-Aerial (D) are 2.54, 3.45, and 3.24 points.

\begin{table}[!t]
\caption{Object-detection results on IndraEye for EO and IR modalities. Higher mAP, mAP$_{50}$, and mAP$_{75}$ are better.}
\label{tab:det-indraeye-expanded-backbone-v2}
\centering
\footnotesize
\renewcommand{\arraystretch}{1.15}
\begin{tabular}{@{}l c c c c@{}}
\hline
\textbf{Method} & \textbf{Backbone} & \textbf{mAP} & \textbf{mAP$_{50}$} & \textbf{mAP$_{75}$} \\
\hline
\multicolumn{5}{c}{\textit{EO Modality}} \\
\hline
Faster R-CNN        & ResNet-50-FPN       & 40.50          & 47.60          & 32.40          \\
ReDet               & ResNet-101          & 37.80          & 43.90          & 28.70          \\
Oriented R-CNN      & ResNet-50-FPN       & \textbf{48.20} & \textbf{56.30} & \textbf{40.10} \\
YOLOv5              & CSPDarknet53        & 45.00          & 52.80          & 36.50          \\
RetinaNet           & ResNet-50-FPN       & 42.30          & 50.10          & 34.20          \\
SSD                 & VGG16               & 39.00          & 46.00          & 30.50          \\
OrientedFormer      & ResNet-50           & 39.02          & 47.60          & 30.50          \\
RS-DETR             & ResNet-50-FPN       & 7.67           & 12.32          & 8.01           \\
\hline
\textbf{GAAT(D) (Ours)} & GAAT             & 44.47          & 55.00          & 38.98          \\
\hline
\multicolumn{5}{c}{\textit{IR Modality}} \\
\hline
Faster R-CNN        & ResNet-50-FPN       & 50.10          & 57.30          & 42.80          \\
ReDet               & ResNet-101          & 37.00          & 43.80          & 29.50          \\
Oriented R-CNN      & ResNet-50-FPN       & 58.40          & 65.60          & 50.20          \\
YOLOv5              & CSPDarknet53        & 55.20          & 62.90          & 46.70          \\
RetinaNet           & ResNet-50-FPN       & 52.60          & 59.70          & 44.30          \\
SSD                 & VGG16               & 47.00          & 54.00          & 38.00          \\
OrientedFormer      & ResNet-50           & \textbf{58.57} & 64.00          & \textbf{53.10} \\
RS-DETR             & ResNet-50-FPN       & 21.59          & 33.94          & 22.86          \\
\hline
\textbf{GAAT(D) (Ours)} & GAAT             & 57.50          & \textbf{71.10} & 52.14          \\
\hline
\end{tabular}
\end{table}

The IndraEye results differ by modality. For EO, GAAT(D) remains below Oriented R-CNN on all three metrics; for IR, it achieves the highest listed mAP$_{50}$ (71.10), while its mAP (57.50) and mAP$_{75}$ (52.14) remain below the best listed values of 58.57 and 53.10 from OrientedFormer.

\subsection{UAVMeta Evaluation Protocol and Scope}
\label{subsec:appendix-uavmeta-protocol}

UAVMeta 1.0.0-rc1 exposes three supervised perception tasks: scene classification, object detection, and semantic segmentation. The canonical acquisition-disjoint split contains 1,715/572/288 temporally synchronized RGB--IR pairs for training/validation/test. All stochastic UAVMeta configurations use seeds 42, 43, and 44. Classification ignores stored label 0 (\texttt{Void}) and evaluates IDs 1--8, with validation macro-F1 used for checkpoint selection. Segmentation evaluates all seven stored classes, including background ID 0, and selects checkpoints using validation mIoU. The selected checkpoint is evaluated once on the test set; confusion matrices, per-class statistics, and per-acquisition metrics are archived with the run outputs. The three task comparisons are reported in the main text in Tabs.~\ref{tab:uavmeta-cls}, \ref{tab:ours-det}, and \ref{tab:uavmeta-seg}.

The current release contains no change masks, temporal change labels, object identities, or trajectory annotations. Change detection and multi-object tracking are therefore evaluated on the external annotated benchmarks listed in the main text, while UAVMeta 1.0.0-rc1 is used for scene classification, object detection, and semantic segmentation.

\subsection{Additional Qualitative Results}
\label{subsec:appendix-qualitative}

\begin{figure*}[!p]
  \centering
  \subfloat[UDD5.\label{fig:appendix-udd5}]{%
    \includegraphics[width=0.92\textwidth]{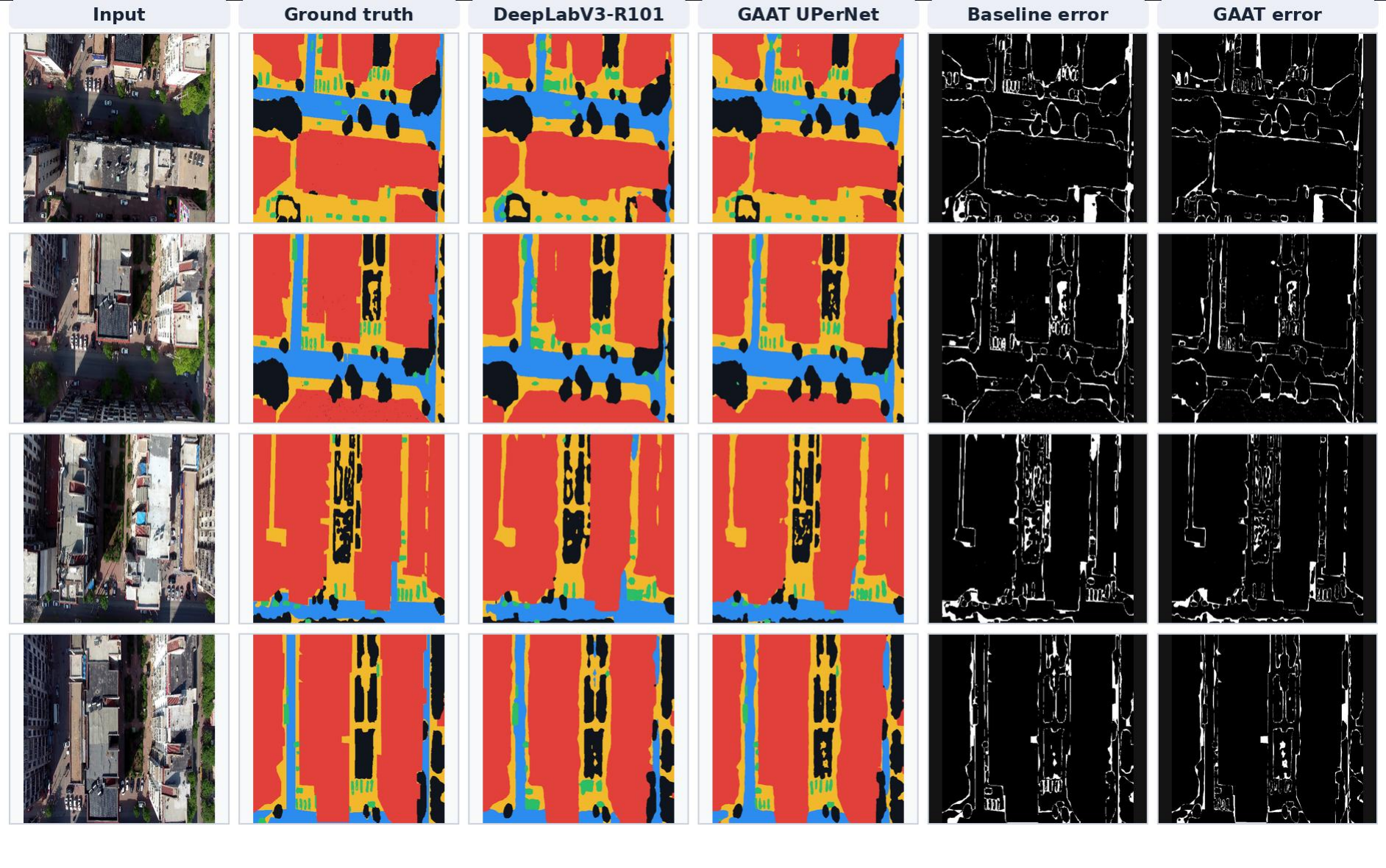}}
  \par\vspace{2pt}
  \subfloat[UAVMeta.\label{fig:appendix-uavmeta-seg}]{%
    \includegraphics[width=0.98\textwidth]{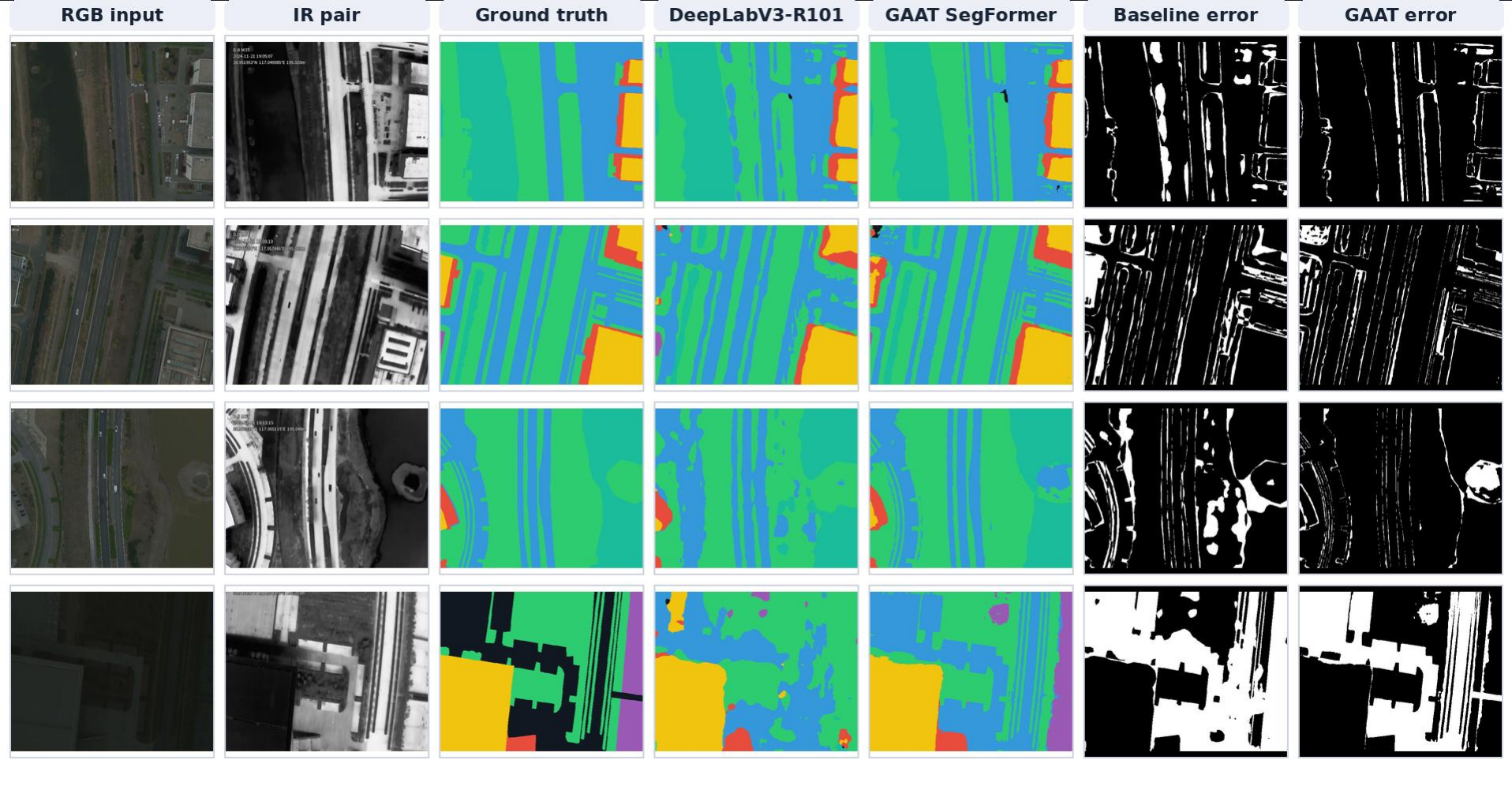}}
  \caption{Qualitative semantic-segmentation results on (a) UDD5 and (b) UAVMeta. In (a), columns show the input image, ground truth, DeepLabV3-R101 prediction, GAAT(U) prediction with UPerNet, baseline error map, and GAAT error map. In (b), columns show the RGB input, paired IR input, ground truth, DeepLabV3-R101 prediction, GAAT(C) prediction with a SegFormer-style decoder, baseline error map, and GAAT error map. Each row presents one example.}
  \label{fig:appendix-seg-qualitative}
\end{figure*}

\begin{figure*}[!p]
  \centering
  \subfloat[UAVDT.\label{fig:appendix-uavdt}]{%
    \includegraphics[width=0.49\textwidth]{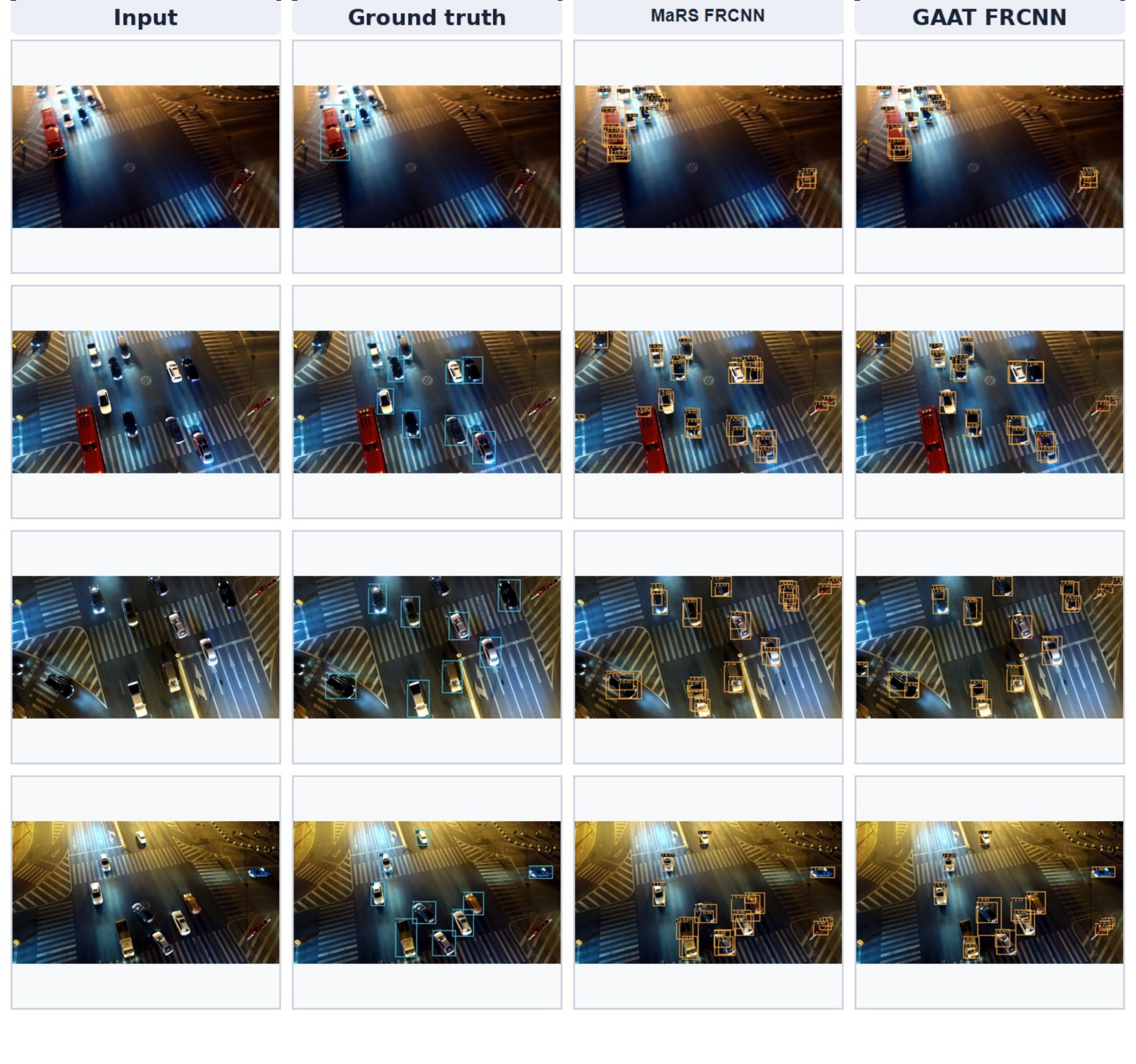}}%
  \hfill
  \subfloat[VisDrone-DET.\label{fig:appendix-visdrone}]{%
    \includegraphics[width=0.49\textwidth]{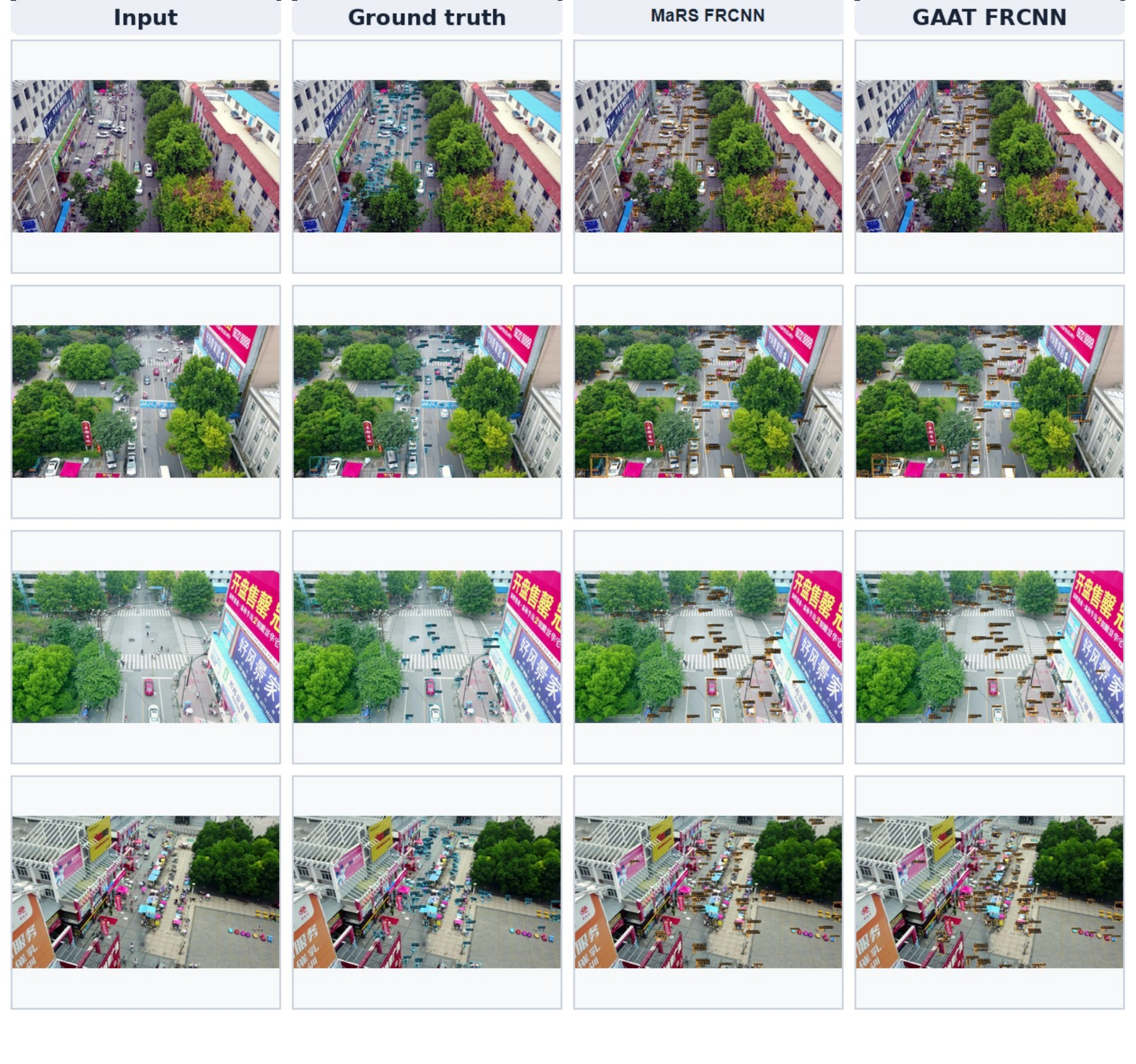}}
  \par\vspace{2pt}
  \subfloat[UAVMeta.\label{fig:appendix-uavmeta-det}]{%
    \includegraphics[width=0.78\textwidth]{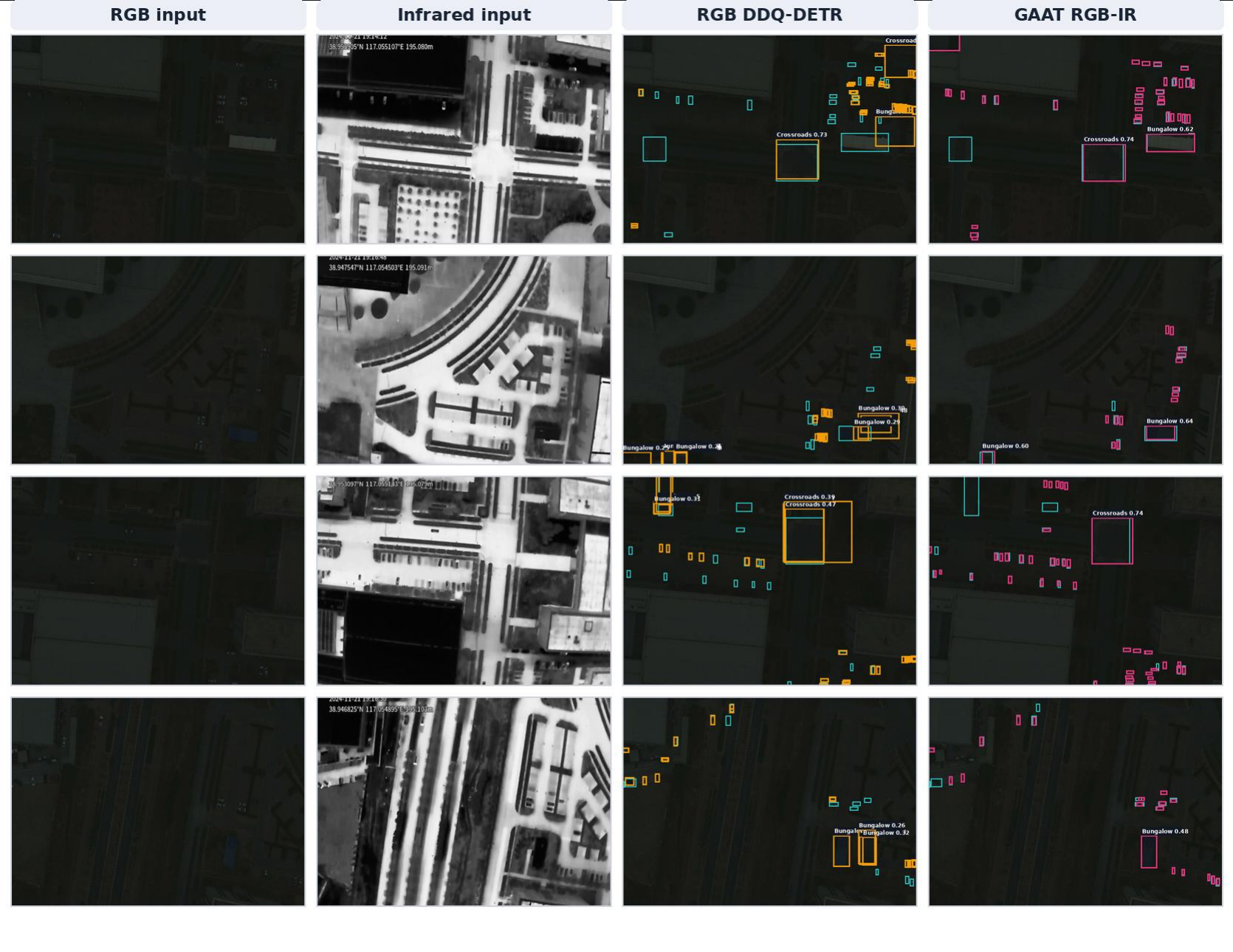}}
  \caption{Qualitative object-detection results on (a) UAVDT, (b) VisDrone-DET, and (c) UAVMeta. Panels (a) and (b) show the input image, ground-truth boxes, MaRS predictions with a Faster R-CNN head, and GAAT predictions with the same detection head. Panel (c) shows the RGB input, paired IR input, RGB-only DDQ-DETR prediction, and RGB--IR GAAT(D) prediction. Each row presents one example.}
  \label{fig:appendix-det-qualitative}
\end{figure*}

\begin{figure*}[!p]
  \centering
  \subfloat[CDD.\label{fig:appendix-cdd}]{%
    \includegraphics[width=0.88\textwidth]{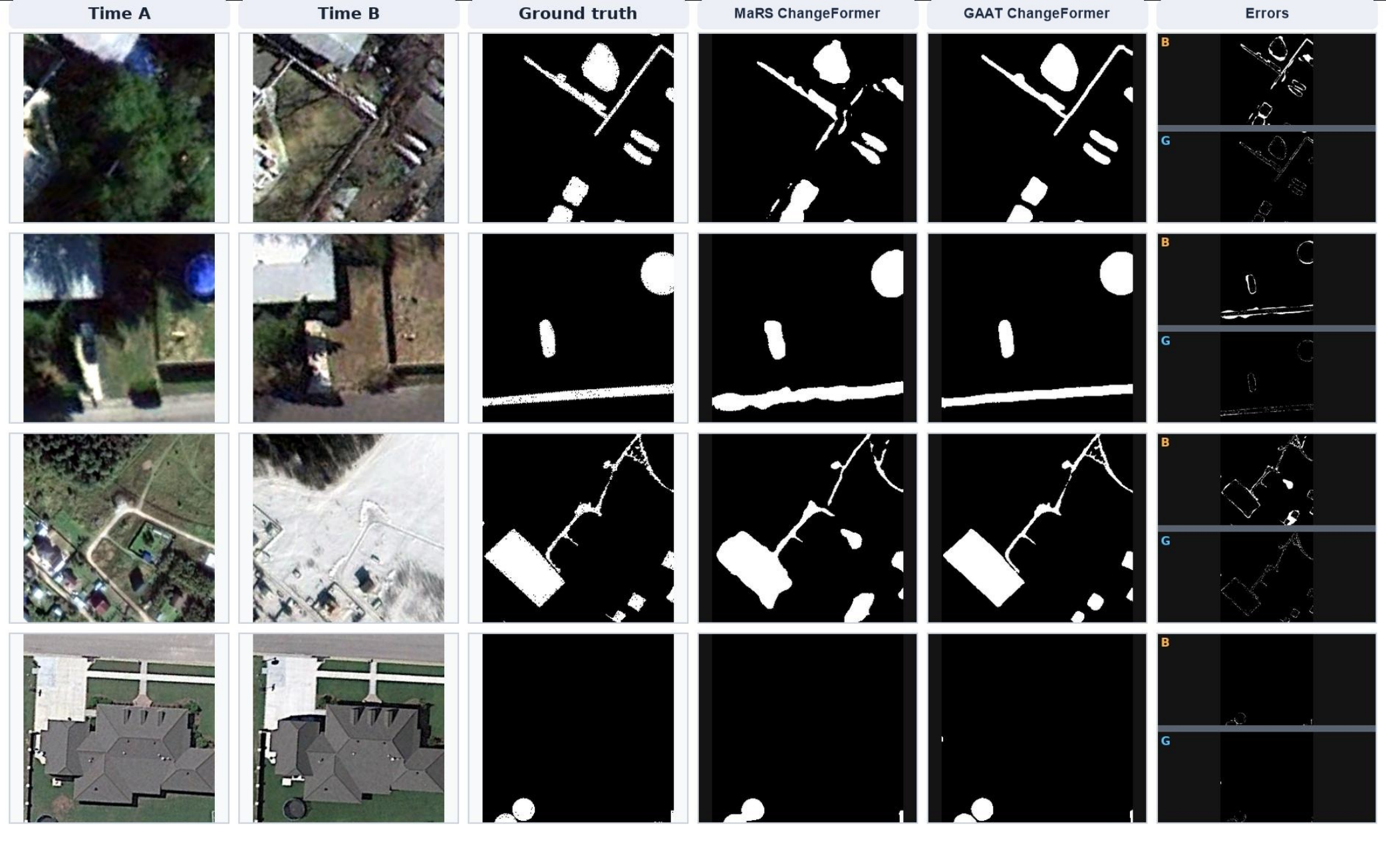}}
  \par\vspace{2pt}
  \subfloat[LEVIR-CD.\label{fig:appendix-levir-cd}]{%
    \includegraphics[width=0.88\textwidth]{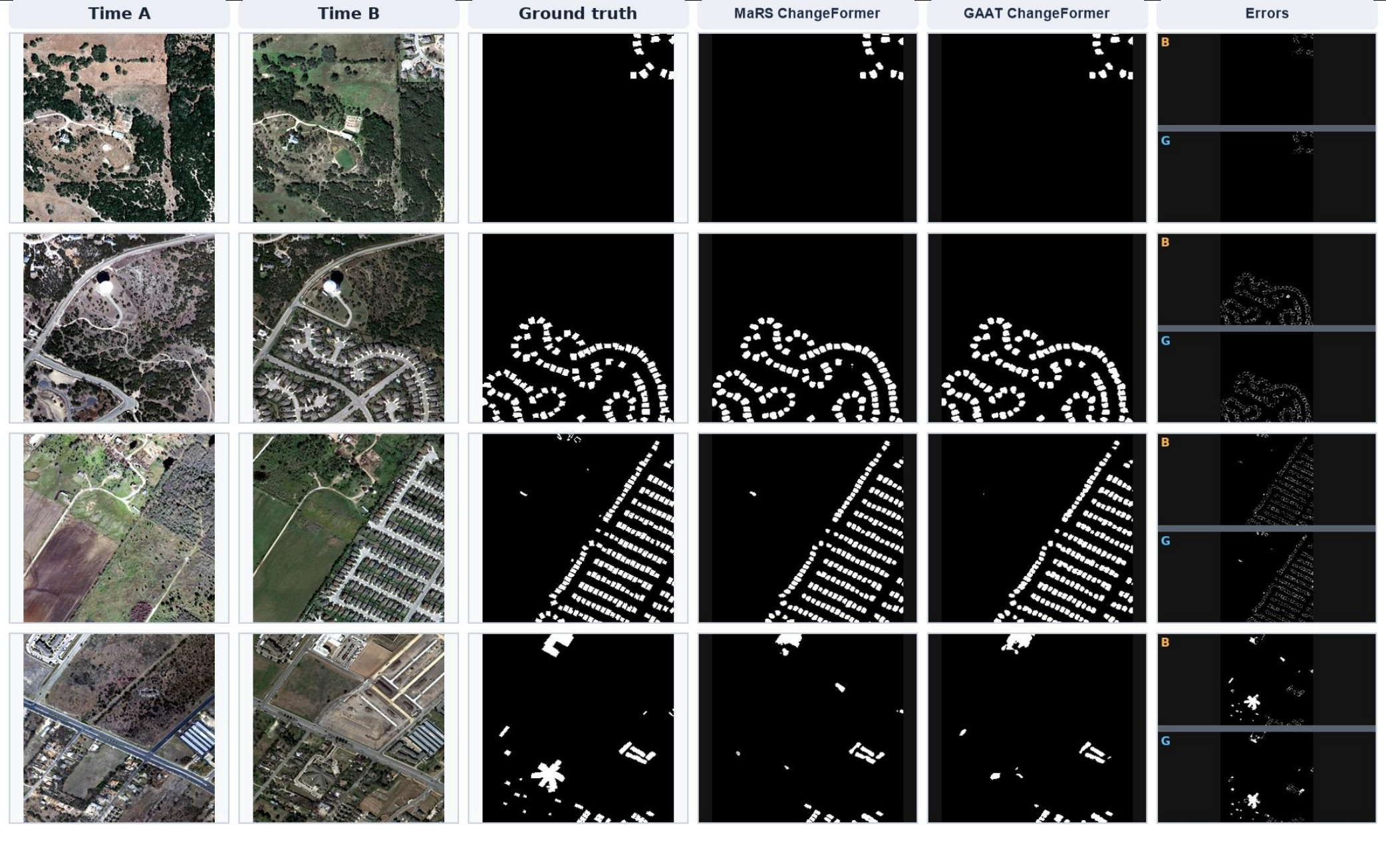}}
  \caption{Qualitative change-detection results on (a) CDD and (b) LEVIR-CD. Columns show the images at times A and B, ground-truth change mask, MaRS prediction with ChangeFormer, GAAT(CF) prediction, and error maps. In the final column, B and G denote the baseline and GAAT error maps, respectively.}
  \label{fig:appendix-cd-qualitative}
\end{figure*}

The following figures collect representative outputs after the quantitative comparisons above. Fig.~\ref{fig:appendix-seg-qualitative} covers semantic segmentation on UDD5 and UAVMeta, Fig.~\ref{fig:appendix-det-qualitative} covers object detection on UAVDT, VisDrone-DET, and UAVMeta, and Fig.~\ref{fig:appendix-cd-qualitative} covers bi-temporal change detection on CDD and LEVIR-CD. These examples are qualitative and complement the dataset-level metrics reported above; the captions specify the inputs, reference annotations, predictions, and error maps shown in each panel.

\clearpage

\clearpage
\bibliographystyle{IEEEtran}
\bibliography{references}

@inproceedings{astruc2024omnisat,
  author       = {Astruc, Guillaume and Gonthier, Nicolas and Mallet, Clement and Landrieu, Loic},
  title        = {OmniSat: Self-supervised Modality Fusion for Earth Observation},
  booktitle    = {Lecture Notes in Computer Science},
  year         = {2025},
  pages        = {409--427},
  doi          = {10.1007/978-3-031-73390-1_24},
  url          = {https://doi.org/10.1007/978-3-031-73390-1_24}
}

@inproceedings{astruc2025anysat,
  author       = {Astruc, Guillaume and Gonthier, Nicolas and Mallet, Cl{\'{e}}ment and Landrieu, Loic},
  title        = {AnySat: One Earth Observation Model for Many Resolutions, Scales, and Modalities},
  booktitle    = {2025 IEEE/CVF Conference on Computer Vision and Pattern Recognition (CVPR)},
  year         = {2025},
  pages        = {19530--19540},
  doi          = {10.1109/CVPR52734.2025.01819},
  url          = {https://doi.org/10.1109/CVPR52734.2025.01819}
}

@inproceedings{bachmann2022multimae,
  author       = {Bachmann, Roman and Mizrahi, David and Atanov, Andrei and Zamir, Amir},
  title        = {MultiMAE: Multi-modal Multi-task Masked Autoencoders},
  booktitle    = {Lecture Notes in Computer Science},
  year         = {2022},
  pages        = {348--367},
  doi          = {10.1007/978-3-031-19836-6_20},
  url          = {https://doi.org/10.1007/978-3-031-19836-6_20}
}

@inproceedings{bandara2022changeformer,
  author       = {Bandara, Wele Gedara Chaminda and Patel, Vishal M.},
  title        = {A Transformer-Based Siamese Network for Change Detection},
  booktitle    = {IGARSS 2022 - 2022 IEEE International Geoscience and Remote Sensing Symposium},
  year         = {2022},
  pages        = {207--210},
  doi          = {10.1109/igarss46834.2022.9883686},
  url          = {https://doi.org/10.1109/igarss46834.2022.9883686}
}

@inproceedings{bastani2023satlas,
  author       = {Bastani, Favyen and Wolters, Piper and Gupta, Ritwik and Ferdinando, Joe and Kembhavi, Aniruddha},
  title        = {SatlasPretrain: A Large-Scale Dataset for Remote Sensing Image Understanding},
  booktitle    = {2023 IEEE/CVF International Conference on Computer Vision (ICCV)},
  year         = {2023},
  pages        = {16772--16782},
  doi          = {10.1109/iccv51070.2023.01538},
  url          = {https://doi.org/10.1109/iccv51070.2023.01538}
}

@article{bernardin2008clearmot,
  author       = {Bernardin, Keni and Stiefelhagen, Rainer},
  title        = {Evaluating Multiple Object Tracking Performance: The CLEAR MOT Metrics},
  journal      = {EURASIP Journal on Image and Video Processing},
  year         = {2008},
  volume       = {2008},
  pages        = {1--10},
  doi          = {10.1155/2008/246309},
  url          = {https://doi.org/10.1155/2008/246309}
}

@inproceedings{bewley2016sort,
  author       = {Bewley, Alex and Ge, Zongyuan and Ott, Lionel and Ramos, Fabio and Upcroft, Ben},
  title        = {Simple online and realtime tracking},
  booktitle    = {2016 IEEE International Conference on Image Processing (ICIP)},
  year         = {2016},
  pages        = {3464--3468},
  doi          = {10.1109/ICIP.2016.7533003},
  url          = {https://doi.org/10.1109/ICIP.2016.7533003}
}

@inproceedings{cai2018cascadercnn,
  author       = {Cai, Zhaowei and Vasconcelos, Nuno},
  title        = {Cascade R-CNN: Delving Into High Quality Object Detection},
  booktitle    = {2018 IEEE/CVF Conference on Computer Vision and Pattern Recognition},
  year         = {2018},
  pages        = {6154--6162},
  doi          = {10.1109/CVPR.2018.00644},
  url          = {https://doi.org/10.1109/CVPR.2018.00644}
}

@inproceedings{cao2023ocsort,
  author       = {Cao, Jinkun and Pang, Jiangmiao and Weng, Xinshuo and Khirodkar, Rawal and Kitani, Kris},
  title        = {Observation-Centric SORT: Rethinking SORT for Robust Multi-Object Tracking},
  booktitle    = {2023 IEEE/CVF Conference on Computer Vision and Pattern Recognition (CVPR)},
  year         = {2023},
  pages        = {9686--9696},
  doi          = {10.1109/cvpr52729.2023.00934},
  url          = {https://doi.org/10.1109/cvpr52729.2023.00934}
}

@inproceedings{carion2020detr,
  author       = {Carion, Nicolas and Massa, Francisco and Synnaeve, Gabriel and Usunier, Nicolas and Kirillov, Alexander and Zagoruyko, Sergey},
  title        = {End-to-End Object Detection with Transformers},
  booktitle    = {Lecture Notes in Computer Science},
  year         = {2020},
  pages        = {213--229},
  doi          = {10.1007/978-3-030-58452-8_13},
  url          = {https://doi.org/10.1007/978-3-030-58452-8_13}
}

@inproceedings{caron2021dino,
  author       = {Caron, Mathilde and Touvron, Hugo and Misra, Ishan and Jegou, Herve and Mairal, Julien and Bojanowski, Piotr and Joulin, Armand},
  title        = {Emerging Properties in Self-Supervised Vision Transformers},
  booktitle    = {2021 IEEE/CVF International Conference on Computer Vision (ICCV)},
  year         = {2021},
  pages        = {9650--9660},
  doi          = {10.1109/ICCV48922.2021.00951},
  url          = {https://doi.org/10.1109/ICCV48922.2021.00951}
}

@inproceedings{cayedaudt2018fcsiam,
  author       = {Caye Daudt, Rodrigo and Le Saux, Bertr and Boulch, Alexandre},
  title        = {Fully Convolutional Siamese Networks for Change Detection},
  booktitle    = {2018 25th IEEE International Conference on Image Processing (ICIP)},
  year         = {2018},
  pages        = {4063--4067},
  doi          = {10.1109/ICIP.2018.8451652},
  url          = {https://doi.org/10.1109/ICIP.2018.8451652}
}

@inproceedings{chen2018deeplabv3plus,
  author       = {Chen, Liang-Chieh and Zhu, Yukun and Papandreou, George and Schroff, Florian and Adam, Hartwig},
  title        = {Encoder-Decoder with Atrous Separable Convolution for Semantic Image Segmentation},
  booktitle    = {Lecture Notes in Computer Science},
  year         = {2018},
  pages        = {833--851},
  doi          = {10.1007/978-3-030-01234-2_49},
  url          = {https://doi.org/10.1007/978-3-030-01234-2_49}
}

@misc{chen2020simclr,
  author       = {Chen, Ting and Kornblith, Simon and Norouzi, Mohammad and Hinton, Geoffrey},
  title        = {A Simple Framework for Contrastive Learning of Visual Representations},
  year         = {2020},
  eprint       = {2002.05709},
  archivePrefix= {arXiv},
  url          = {https://arxiv.org/abs/2002.05709}
}

@article{chen2021bit,
  author       = {Chen, Hao and Qi, Zipeng and Shi, Zhenwei},
  title        = {Remote Sensing Image Change Detection With Transformers},
  journal      = {IEEE Transactions on Geoscience and Remote Sensing},
  year         = {2022},
  volume       = {60},
  pages        = {1--14},
  doi          = {10.1109/tgrs.2021.3095166},
  url          = {https://doi.org/10.1109/tgrs.2021.3095166}
}

@article{chen2021levircd,
  author       = {Chen, Hao and Shi, Zhenwei},
  title        = {A Spatial-Temporal Attention-Based Method and a New Dataset for Remote Sensing Image Change Detection},
  journal      = {Remote Sensing},
  year         = {2020},
  volume       = {12},
  number       = {10},
  pages        = {1662},
  doi          = {10.3390/rs12101662},
  url          = {https://doi.org/10.3390/rs12101662}
}

@inproceedings{chen2022tensorf,
  author       = {Chen, Anpei and Xu, Zexiang and Geiger, Andreas and Yu, Jingyi and Su, Hao},
  title        = {TensoRF: Tensorial Radiance Fields},
  booktitle    = {Lecture Notes in Computer Science},
  year         = {2022},
  pages        = {333--350},
  doi          = {10.1007/978-3-031-19824-3_20},
  url          = {https://doi.org/10.1007/978-3-031-19824-3_20}
}

@inproceedings{chen2024thermal3dgs,
  author       = {Chen, Qian and Shu, Shihao and Bai, Xiangzhi},
  title        = {Thermal3D-GS: Physics-Induced 3D Gaussians for Thermal Infrared Novel-View Synthesis},
  booktitle    = {Lecture Notes in Computer Science},
  year         = {2025},
  pages        = {253--269},
  doi          = {10.1007/978-3-031-73383-3_15},
  url          = {https://doi.org/10.1007/978-3-031-73383-3_15}
}

@article{cheng2017resisc45,
  author       = {Cheng, Gong and Han, Junwei and Lu, Xiaoqiang},
  title        = {Remote Sensing Image Scene Classification: Benchmark and State of the Art},
  journal      = {Proceedings of the IEEE},
  year         = {2017},
  volume       = {105},
  number       = {10},
  pages        = {1865--1883},
  doi          = {10.1109/jproc.2017.2675998},
  url          = {https://doi.org/10.1109/jproc.2017.2675998}
}

@article{codegoni2022tinycd,
  author       = {Codegoni, Andrea and Lombardi, Gabriele and Ferrari, Alessandro},
  title        = {TINYCD: a (not so) deep learning model for change detection},
  journal      = {Neural Computing and Applications},
  year         = {2023},
  volume       = {35},
  number       = {11},
  pages        = {8471--8486},
  doi          = {10.1007/s00521-022-08122-3},
  url          = {https://doi.org/10.1007/s00521-022-08122-3}
}

@misc{cong2022satmae,
  author       = {Cong, Yezhen and Khanna, Samar and Meng, Chenlin and Liu, Patrick and Rozi, Erik and He, Yutong and Burke, Marshall and Lobell, David B. and Ermon, Stefano},
  title        = {SatMAE: Pre-training Transformers for Temporal and Multi-Spectral Satellite Imagery},
  year         = {2022},
  eprint       = {2207.08051},
  archivePrefix= {arXiv},
  url          = {https://arxiv.org/abs/2207.08051}
}

@inproceedings{dai2017dcn,
  author       = {Dai, Jifeng and Qi, Haozhi and Xiong, Yuwen and Li, Yi and Zhang, Guodong and Hu, Han and Wei, Yichen},
  title        = {Deformable Convolutional Networks},
  booktitle    = {2017 IEEE International Conference on Computer Vision (ICCV)},
  year         = {2017},
  pages        = {764--773},
  doi          = {10.1109/ICCV.2017.89},
  url          = {https://doi.org/10.1109/ICCV.2017.89}
}

@inproceedings{detone2018superpoint,
  author       = {DeTone, Daniel and Malisiewicz, Tomasz and Rabinovich, Andrew},
  title        = {SuperPoint: Self-Supervised Interest Point Detection and Description},
  booktitle    = {2018 IEEE/CVF Conference on Computer Vision and Pattern Recognition Workshops (CVPRW)},
  year         = {2018},
  pages        = {224--236},
  doi          = {10.1109/cvprw.2018.00060},
  url          = {https://doi.org/10.1109/cvprw.2018.00060}
}

@article{diao2025ringmoaerial,
  author       = {Diao, Wenhui and Yu, Haichen and Kang, Kaiyue and Ling, Tong and Liu, Di and Feng, Yingchao and Bi, Hanbo and Ren, Libo and Li, Xuexue and Mao, Yongqiang and Sun, Xian},
  title        = {RingMo-Aerial: An Aerial Remote Sensing Foundation Model With Affine Transformation Contrastive Learning},
  journal      = {IEEE Transactions on Pattern Analysis and Machine Intelligence},
  year         = {2025},
  volume       = {47},
  number       = {12},
  pages        = {10900--10913},
  doi          = {10.1109/TPAMI.2025.3602237},
  url          = {https://doi.org/10.1109/TPAMI.2025.3602237}
}

@inproceedings{ding2019roitransformer,
  author       = {Ding, Jian and Xue, Nan and Long, Yang and Xia, Gui-Song and Lu, Qikai},
  title        = {Learning RoI Transformer for Oriented Object Detection in Aerial Images},
  booktitle    = {2019 IEEE/CVF Conference on Computer Vision and Pattern Recognition (CVPR)},
  year         = {2019},
  pages        = {2849--2858},
  doi          = {10.1109/CVPR.2019.00296},
  url          = {https://doi.org/10.1109/CVPR.2019.00296}
}

@misc{dosovitskiy2021vit,
  author       = {Dosovitskiy, Alexey and Beyer, Lucas and Kolesnikov, Alexander and Weissenborn, Dirk and Zhai, Xiaohua and Unterthiner, Thomas and Dehghani, Mostafa and Minderer, Matthias and Heigold, Georg and Gelly, Sylvain and Uszkoreit, Jakob and Houlsby, Neil},
  title        = {An Image is Worth 16x16 Words: Transformers for Image Recognition at Scale},
  year         = {2020},
  eprint       = {2010.11929},
  archivePrefix= {arXiv},
  url          = {https://arxiv.org/abs/2010.11929}
}

@inproceedings{du2018uavdt,
  author       = {Du, Dawei and Qi, Yuankai and Yu, Hongyang and Yang, Yifan and Duan, Kaiwen and Li, Guorong and Zhang, Weigang and Huang, Qingming and Tian, Qi},
  title        = {The Unmanned Aerial Vehicle Benchmark: Object Detection and Tracking},
  booktitle    = {Lecture Notes in Computer Science},
  year         = {2018},
  pages        = {375--391},
  doi          = {10.1007/978-3-030-01249-6_23},
  url          = {https://doi.org/10.1007/978-3-030-01249-6_23}
}

@inproceedings{du2020visdronecc,
  author       = {Du, Dawei and Wen, Longyin and Zhu, Pengfei and Fan, Heng and Hu, Qinghua and Ling, Haibin and Shah, Mubarak and Pan, Junwen and Al-Ali, Ali and Mohamed, Amr and Imene, Bakour and Dong, Bin and Zhang, Binyu and Nesma, Bouchali Hadia and Xu, Chenfeng and Duan, Chenzhen and Castiello, Ciro and Mencar, Corrado and Liang, Dingkang and Kr{\"{u}}ger, Florian and Vessio, Gennaro and Castellano, Giovanna and Wang, Jieru and Gao, Junyu and Abualsaud, Khalid and Ding, Laihui and Zhao, Lei and Cianciotta, Marco and Saqib, Muhammad and Almaadeed, Noor and Elharrouss, Omar and Lyu, Pei and Wang, Qi and Liu, Shidong and Qiu, Shuang and Pan, Siyang and Al-Maadeed, Somaya and Khan, Sultan Daud and Khattab, Tamer and Han, Tao and Golda, Thomas and Xu, Wei and Bai, Xiang and Xu, Xiaoqing and Li, Xuelong and Zhao, Yanyun and Tian, Ye and Lin, Yingnan and Xu, Yongchao and Yao, Yuehan and Xu, Zhenyu and Zhao, Zhijian and Luo, Zhipeng and Wei, Zhiwei and Zhao, Zhiyuan},
  title        = {VisDrone-CC2020: The Vision Meets Drone Crowd Counting Challenge Results},
  booktitle    = {Lecture Notes in Computer Science},
  year         = {2020},
  pages        = {675--691},
  doi          = {10.1007/978-3-030-66823-5_41},
  url          = {https://doi.org/10.1007/978-3-030-66823-5_41}
}

@inproceedings{fan2025movingdrone,
  author       = {Fan, Yaowu and Wan, Jia and Han, Tao and Chan, Antoni B. and Ma, Andy J.},
  title        = {Video Individual Counting for Moving Drones},
  booktitle    = {Proceedings of the IEEE/CVF International Conference on Computer Vision (ICCV)},
  year         = {2025},
  pages        = {12284--12293},
  doi          = {10.1109/ICCV51701.2025.01142},
  eprint       = {2503.10701},
  archivePrefix= {arXiv},
  url          = {https://doi.org/10.1109/ICCV51701.2025.01142}
}

@article{fang2022snunetcd,
  author       = {Fang, Sheng and Li, Kaiyu and Shao, Jinyuan and Li, Zhe},
  title        = {SNUNet-CD: A Densely Connected Siamese Network for Change Detection of VHR Images},
  journal      = {IEEE Geoscience and Remote Sensing Letters},
  year         = {2022},
  volume       = {19},
  pages        = {1--5},
  doi          = {10.1109/LGRS.2021.3056416},
  url          = {https://doi.org/10.1109/LGRS.2021.3056416}
}

@article{fischler1981ransac,
  author       = {Fischler, Martin A. and Bolles, Robert C.},
  title        = {Random sample consensus},
  journal      = {Communications of the ACM},
  year         = {1981},
  volume       = {24},
  number       = {6},
  pages        = {381--395},
  doi          = {10.1145/358669.358692},
  url          = {https://doi.org/10.1145/358669.358692}
}

@inproceedings{fu2019danet,
  author       = {Fu, Jun and Liu, Jing and Tian, Haijie and Li, Yong and Bao, Yongjun and Fang, Zhiwei and Lu, Hanqing},
  title        = {Dual Attention Network for Scene Segmentation},
  booktitle    = {2019 IEEE/CVF Conference on Computer Vision and Pattern Recognition (CVPR)},
  year         = {2019},
  pages        = {3146--3154},
  doi          = {10.1109/CVPR.2019.00326},
  url          = {https://doi.org/10.1109/CVPR.2019.00326}
}

@inproceedings{fuller2023croma,
  author       = {Fuller, Anthony and Millard, Koreen and Green, James},
  title        = {CROMA: Remote Sensing Representations with Contrastive Radar-Optical Masked Autoencoders},
  booktitle    = {Advances in Neural Information Processing Systems 36},
  year         = {2023},
  pages        = {5506--5538},
  doi          = {10.52202/075280-0241},
  url          = {https://doi.org/10.52202/075280-0241}
}

@inproceedings{guo2024skysense,
  author       = {Guo, Xin and Lao, Jiangwei and Dang, Bo and Zhang, Yingying and Yu, Lei and Ru, Lixiang and Zhong, Liheng and Huang, Ziyuan and Wu, Kang and Hu, Dingxiang and He, Huimei and Wang, Jian and Chen, Jingdong and Yang, Ming and Zhang, Yongjun and Li, Yansheng},
  title        = {SkySense: A Multi-Modal Remote Sensing Foundation Model Towards Universal Interpretation for Earth Observation Imagery},
  booktitle    = {Proceedings of the IEEE/CVF Conference on Computer Vision and Pattern Recognition (CVPR)},
  year         = {2024},
  pages        = {27672--27683},
  doi          = {10.1109/CVPR52733.2024.02613},
  eprint       = {2312.10115},
  archivePrefix= {arXiv},
  url          = {https://doi.org/10.1109/CVPR52733.2024.02613}
}

@inproceedings{ha2017mfnet,
  author       = {Ha, Qishen and Watanabe, Kohei and Karasawa, Takumi and Ushiku, Yoshitaka and Harada, Tatsuya},
  title        = {MFNet: Towards real-time semantic segmentation for autonomous vehicles with multi-spectral scenes},
  booktitle    = {2017 IEEE/RSJ International Conference on Intelligent Robots and Systems (IROS)},
  year         = {2017},
  pages        = {5108--5115},
  doi          = {10.1109/IROS.2017.8206396},
  url          = {https://doi.org/10.1109/IROS.2017.8206396}
}

@article{han2022s2anet,
  author       = {Han, Jiaming and Ding, Jian and Li, Jie and Xia, Gui-Song},
  title        = {Align Deep Features for Oriented Object Detection},
  journal      = {IEEE Transactions on Geoscience and Remote Sensing},
  year         = {2022},
  volume       = {60},
  pages        = {1--11},
  doi          = {10.1109/TGRS.2021.3062048},
  url          = {https://doi.org/10.1109/TGRS.2021.3062048}
}

@inproceedings{he2016resnet,
  author       = {He, Kaiming and Zhang, Xiangyu and Ren, Shaoqing and Sun, Jian},
  title        = {Deep Residual Learning for Image Recognition},
  booktitle    = {2016 IEEE Conference on Computer Vision and Pattern Recognition (CVPR)},
  year         = {2016},
  pages        = {770--778},
  doi          = {10.1109/CVPR.2016.90},
  url          = {https://doi.org/10.1109/CVPR.2016.90}
}

@inproceedings{he2020moco,
  author       = {He, Kaiming and Fan, Haoqi and Wu, Yuxin and Xie, Saining and Girshick, Ross},
  title        = {Momentum Contrast for Unsupervised Visual Representation Learning},
  booktitle    = {2020 IEEE/CVF Conference on Computer Vision and Pattern Recognition (CVPR)},
  year         = {2020},
  pages        = {9729--9738},
  doi          = {10.1109/CVPR42600.2020.00975},
  url          = {https://doi.org/10.1109/CVPR42600.2020.00975}
}

@inproceedings{he2022mae,
  author       = {He, Kaiming and Chen, Xinlei and Xie, Saining and Li, Yanghao and Dollar, Piotr and Girshick, Ross},
  title        = {Masked Autoencoders Are Scalable Vision Learners},
  booktitle    = {2022 IEEE/CVF Conference on Computer Vision and Pattern Recognition (CVPR)},
  year         = {2022},
  pages        = {16000--16009},
  doi          = {10.1109/CVPR52688.2022.01553},
  url          = {https://doi.org/10.1109/CVPR52688.2022.01553}
}

@article{helber2019eurosat,
  author       = {Helber, Patrick and Bischke, Benjamin and Dengel, Andreas and Borth, Damian},
  title        = {EuroSAT: A Novel Dataset and Deep Learning Benchmark for Land Use and Land Cover Classification},
  journal      = {IEEE Journal of Selected Topics in Applied Earth Observations and Remote Sensing},
  year         = {2019},
  volume       = {12},
  number       = {7},
  pages        = {2217--2226},
  doi          = {10.1109/JSTARS.2019.2918242},
  url          = {https://doi.org/10.1109/JSTARS.2019.2918242}
}

@misc{hinton2015distill,
  author       = {Hinton, Geoffrey and Vinyals, Oriol and Dean, Jeff},
  title        = {Distilling the Knowledge in a Neural Network},
  year         = {2015},
  eprint       = {1503.02531},
  archivePrefix= {arXiv},
  url          = {https://arxiv.org/abs/1503.02531}
}

@inproceedings{howard2019mobilenetv3,
  author       = {Howard, Andrew and Sandler, Mark and Chen, Bo and Wang, Weijun and Chen, Liang-Chieh and Tan, Mingxing and Chu, Grace and Vasudevan, Vijay and Zhu, Yukun and Pang, Ruoming and Adam, Hartwig and Le, Quoc},
  title        = {Searching for MobileNetV3},
  booktitle    = {2019 IEEE/CVF International Conference on Computer Vision (ICCV)},
  year         = {2019},
  pages        = {1314--1324},
  doi          = {10.1109/ICCV.2019.00140},
  url          = {https://doi.org/10.1109/ICCV.2019.00140}
}

@inproceedings{jia2021llvip,
  author       = {Jia, Xinyu and Zhu, Chuang and Li, Minzhen and Tang, Wenqi and Zhou, Wenli},
  title        = {LLVIP: A Visible-infrared Paired Dataset for Low-light Vision},
  booktitle    = {2021 IEEE/CVF International Conference on Computer Vision Workshops (ICCVW)},
  year         = {2021},
  pages        = {3489--3497},
  doi          = {10.1109/iccvw54120.2021.00389},
  url          = {https://doi.org/10.1109/iccvw54120.2021.00389}
}

@article{kerbl2023gaussian,
  author       = {Kerbl, Bernhard and Kopanas, Georgios and Leimkuehler, Thomas and Drettakis, George},
  title        = {3D Gaussian Splatting for Real-Time Radiance Field Rendering},
  journal      = {ACM Transactions on Graphics},
  year         = {2023},
  volume       = {42},
  number       = {4},
  pages        = {1--14},
  doi          = {10.1145/3592433},
  url          = {https://doi.org/10.1145/3592433}
}

@inproceedings{li2018csrnet,
  author       = {Li, Yuhong and Zhang, Xiaofan and Chen, Deming},
  title        = {CSRNet: Dilated Convolutional Neural Networks for Understanding the Highly Congested Scenes},
  booktitle    = {2018 IEEE/CVF Conference on Computer Vision and Pattern Recognition},
  year         = {2018},
  pages        = {1091--1100},
  doi          = {10.1109/CVPR.2018.00120},
  url          = {https://doi.org/10.1109/CVPR.2018.00120}
}

@article{liang2022transcrowd,
  author       = {Liang, Dingkang and Chen, Xiwu and Xu, Wei and Zhou, Yu and Bai, Xiang},
  title        = {TransCrowd: weakly-supervised crowd counting with transformers},
  journal      = {Science China Information Sciences},
  year         = {2022},
  volume       = {65},
  number       = {6},
  pages        = {160104},
  doi          = {10.1007/s11432-021-3445-y},
  url          = {https://doi.org/10.1007/s11432-021-3445-y}
}

@inproceedings{lin2014coco,
  author       = {Lin, Tsung-Yi and Maire, Michael and Belongie, Serge and Hays, James and Perona, Pietro and Ramanan, Deva and Doll{\'{a}}r, Piotr and Zitnick, C. Lawrence},
  title        = {Microsoft COCO: Common Objects in Context},
  booktitle    = {Lecture Notes in Computer Science},
  year         = {2014},
  pages        = {740--755},
  doi          = {10.1007/978-3-319-10602-1_48},
  url          = {https://doi.org/10.1007/978-3-319-10602-1_48}
}

@inproceedings{lin2017focalloss,
  author       = {Lin, Tsung-Yi and Goyal, Priya and Girshick, Ross and He, Kaiming and Dollar, Piotr},
  title        = {Focal Loss for Dense Object Detection},
  booktitle    = {2017 IEEE International Conference on Computer Vision (ICCV)},
  year         = {2017},
  pages        = {2980--2988},
  doi          = {10.1109/ICCV.2017.324},
  url          = {https://doi.org/10.1109/ICCV.2017.324}
}

@inproceedings{lindenberger2023lightglue,
  author       = {Lindenberger, Philipp and Sarlin, Paul-Edouard and Pollefeys, Marc},
  title        = {LightGlue: Local Feature Matching at Light Speed},
  booktitle    = {2023 IEEE/CVF International Conference on Computer Vision (ICCV)},
  year         = {2023},
  pages        = {17627--17638},
  doi          = {10.1109/iccv51070.2023.01616},
  url          = {https://doi.org/10.1109/iccv51070.2023.01616}
}

@inproceedings{liu2016ssd,
  author       = {Liu, Wei and Anguelov, Dragomir and Erhan, Dumitru and Szegedy, Christian and Reed, Scott and Fu, Cheng-Yang and Berg, Alexander C.},
  title        = {SSD: Single Shot MultiBox Detector},
  booktitle    = {Lecture Notes in Computer Science},
  year         = {2016},
  pages        = {21--37},
  doi          = {10.1007/978-3-319-46448-0_2},
  url          = {https://doi.org/10.1007/978-3-319-46448-0_2}
}

@inproceedings{liu2021swin,
  author       = {Liu, Ze and Lin, Yutong and Cao, Yue and Hu, Han and Wei, Yixuan and Zhang, Zheng and Lin, Stephen and Guo, Baining},
  title        = {Swin Transformer: Hierarchical Vision Transformer using Shifted Windows},
  booktitle    = {2021 IEEE/CVF International Conference on Computer Vision (ICCV)},
  year         = {2021},
  pages        = {10012--10022},
  doi          = {10.1109/ICCV48922.2021.00986},
  url          = {https://doi.org/10.1109/ICCV48922.2021.00986}
}

@inproceedings{liu2022convnext,
  author       = {Liu, Zhuang and Mao, Hanzi and Wu, Chao-Yuan and Feichtenhofer, Christoph and Darrell, Trevor and Xie, Saining},
  title        = {A ConvNet for the 2020s},
  booktitle    = {2022 IEEE/CVF Conference on Computer Vision and Pattern Recognition (CVPR)},
  year         = {2022},
  pages        = {11976--11986},
  doi          = {10.1109/CVPR52688.2022.01167},
  url          = {https://doi.org/10.1109/CVPR52688.2022.01167}
}

@inproceedings{liu2022swinv2,
  author       = {Liu, Ze and Hu, Han and Lin, Yutong and Yao, Zhuliang and Xie, Zhenda and Wei, Yixuan and Ning, Jia and Cao, Yue and Zhang, Zheng and Dong, Li and Wei, Furu and Guo, Baining},
  title        = {Swin Transformer V2: Scaling Up Capacity and Resolution},
  booktitle    = {2022 IEEE/CVF Conference on Computer Vision and Pattern Recognition (CVPR)},
  year         = {2022},
  pages        = {12009--12019},
  doi          = {10.1109/CVPR52688.2022.01170},
  url          = {https://doi.org/10.1109/CVPR52688.2022.01170}
}

@inproceedings{long2015fcn,
  author       = {Long, Jonathan and Shelhamer, Evan and Darrell, Trevor},
  title        = {Fully convolutional networks for semantic segmentation},
  booktitle    = {2015 IEEE Conference on Computer Vision and Pattern Recognition (CVPR)},
  year         = {2015},
  pages        = {3431--3440},
  doi          = {10.1109/CVPR.2015.7298965},
  url          = {https://doi.org/10.1109/CVPR.2015.7298965}
}

@article{long2021millionaid,
  author       = {Long, Yang and Xia, Gui-Song and Li, Shengyang and Yang, Wen and Yang, Michael Ying and Zhu, Xiao Xiang and Zhang, Liangpei and Li, Deren},
  title        = {On Creating Benchmark Dataset for Aerial Image Interpretation: Reviews, Guidances, and Million-AID},
  journal      = {IEEE Journal of Selected Topics in Applied Earth Observations and Remote Sensing},
  year         = {2021},
  volume       = {14},
  pages        = {4205--4230},
  doi          = {10.1109/jstars.2021.3070368},
  url          = {https://doi.org/10.1109/jstars.2021.3070368}
}

@misc{loshchilov2019adamw,
  author       = {Loshchilov, Ilya and Hutter, Frank},
  title        = {Decoupled Weight Decay Regularization},
  year         = {2017},
  eprint       = {1711.05101},
  archivePrefix= {arXiv},
  url          = {https://arxiv.org/abs/1711.05101}
}

@article{lowe2004sift,
  author       = {Lowe, David G.},
  title        = {Distinctive Image Features from Scale-Invariant Keypoints},
  journal      = {International Journal of Computer Vision},
  year         = {2004},
  volume       = {60},
  number       = {2},
  pages        = {91--110},
  doi          = {10.1023/B:VISI.0000029664.99615.94},
  url          = {https://doi.org/10.1023/B:VISI.0000029664.99615.94}
}

@article{luiten2021hota,
  author       = {Luiten, Jonathon and O{\u{s}}ep, Aljo{\u{s}}a and Dendorfer, Patrick and Torr, Philip and Geiger, Andreas and Leal-Taix{\'{e}}, Laura and Leibe, Bastian},
  title        = {HOTA: A Higher Order Metric for Evaluating Multi-object Tracking},
  journal      = {International Journal of Computer Vision},
  year         = {2021},
  volume       = {129},
  number       = {2},
  pages        = {548--578},
  doi          = {10.1007/s11263-020-01375-2},
  url          = {https://doi.org/10.1007/s11263-020-01375-2}
}

@article{lyu2020uavid,
  author       = {Lyu, Ye and Vosselman, George and Xia, Gui-Song and Yilmaz, Alper and Yang, Michael Ying},
  title        = {UAVid: A semantic segmentation dataset for UAV imagery},
  journal      = {ISPRS Journal of Photogrammetry and Remote Sensing},
  year         = {2020},
  volume       = {165},
  pages        = {108--119},
  doi          = {10.1016/j.isprsjprs.2020.05.009},
  url          = {https://doi.org/10.1016/j.isprsjprs.2020.05.009}
}

@inproceedings{ma2019bayesian,
  author       = {Ma, Zhiheng and Wei, Xing and Hong, Xiaopeng and Gong, Yihong},
  title        = {Bayesian Loss for Crowd Count Estimation With Point Supervision},
  booktitle    = {2019 IEEE/CVF International Conference on Computer Vision (ICCV)},
  year         = {2019},
  pages        = {6142--6151},
  doi          = {10.1109/ICCV.2019.00624},
  url          = {https://doi.org/10.1109/ICCV.2019.00624}
}

@inproceedings{maggiolino2023deepocsort,
  author       = {Maggiolino, Gerard and Ahmad, Adnan and Cao, Jinkun and Kitani, Kris},
  title        = {Deep OC-Sort: Multi-Pedestrian Tracking by Adaptive Re-Identification},
  booktitle    = {2023 IEEE International Conference on Image Processing (ICIP)},
  year         = {2023},
  pages        = {3025--3029},
  doi          = {10.1109/ICIP49359.2023.10222576},
  url          = {https://doi.org/10.1109/ICIP49359.2023.10222576}
}

@inproceedings{manas2021seco,
  author       = {Manas, Oscar and Lacoste, Alexandre and Giro-i-Nieto, Xavier and Vazquez, David and Rodriguez, Pau},
  title        = {Seasonal Contrast: Unsupervised Pre-Training from Uncurated Remote Sensing Data},
  booktitle    = {2021 IEEE/CVF International Conference on Computer Vision (ICCV)},
  year         = {2021},
  pages        = {9414--9423},
  doi          = {10.1109/ICCV48922.2021.00928},
  url          = {https://doi.org/10.1109/ICCV48922.2021.00928}
}

@inproceedings{mildenhall2020nerf,
  author       = {Mildenhall, Ben and Srinivasan, Pratul P. and Tancik, Matthew and Barron, Jonathan T. and Ramamoorthi, Ravi and Ng, Ren},
  title        = {NeRF: Representing Scenes as Neural Radiance Fields for View Synthesis},
  booktitle    = {Lecture Notes in Computer Science},
  year         = {2020},
  pages        = {405--421},
  doi          = {10.1007/978-3-030-58452-8_24},
  url          = {https://doi.org/10.1007/978-3-030-58452-8_24}
}

@article{muhtar2023cmid,
  author       = {Muhtar, Dilxat and Zhang, Xueliang and Xiao, Pengfeng and Li, Zhenshi and Gu, Feng},
  title        = {CMID: A Unified Self-Supervised Learning Framework for Remote Sensing Image Understanding},
  journal      = {IEEE Transactions on Geoscience and Remote Sensing},
  year         = {2023},
  volume       = {61},
  pages        = {1--17},
  doi          = {10.1109/TGRS.2023.3268232},
  url          = {https://doi.org/10.1109/TGRS.2023.3268232}
}

@article{muller2022instantngp,
  author       = {M{\"{u}}ller, Thomas and Evans, Alex and Schied, Christoph and Keller, Alexander},
  title        = {Instant neural graphics primitives with a multiresolution hash encoding},
  journal      = {ACM Transactions on Graphics},
  year         = {2022},
  volume       = {41},
  number       = {4},
  pages        = {1--15},
  doi          = {10.1145/3528223.3530127},
  url          = {https://doi.org/10.1145/3528223.3530127}
}

@article{nie2025m3ot,
  author       = {Nie, Zhihao and Xue, Luyi and Fang, Zhenyu and Ren, Jinchang and Wei, Yufeng and Zheng, Jiangbin},
  title        = {M3OT: A Multi-Drone Multi-Modality dataset for Multi-Object Tracking},
  journal      = {Scientific Data},
  year         = {2025},
  volume       = {12},
  number       = {1},
  pages        = {1927},
  doi          = {10.1038/s41597-025-06204-0},
  url          = {https://doi.org/10.1038/s41597-025-06204-0}
}

@article{ouyang2025kust4k,
  author       = {Ouyang, Junlin and Wang, Qingwang and Shang, Ying and Jin, Pengcheng and Zhong, Hangwei and Zhou, Liman and Shen, Tao},
  title        = {An RGB-TIR Dataset from UAV Platform for Robust Urban Traffic Scenes Semantic Segmentation},
  journal      = {Scientific Data},
  year         = {2025},
  volume       = {12},
  number       = {1},
  pages        = {1701},
  doi          = {10.1038/s41597-025-05994-7},
  url          = {https://doi.org/10.1038/s41597-025-05994-7}
}

@article{ptak2022uavcounting,
  author       = {Ptak, Bartosz and Pieczy{\'{n}}ski, Dominik and Piechocki, Mateusz and Kraft, Marek},
  title        = {On-Board Crowd Counting and Density Estimation Using Low Altitude Unmanned Aerial Vehicles---Looking beyond Beating the Benchmark},
  journal      = {Remote Sensing},
  year         = {2022},
  volume       = {14},
  number       = {10},
  pages        = {2288},
  doi          = {10.3390/rs14102288},
  url          = {https://doi.org/10.3390/rs14102288}
}

@inproceedings{redmon2016yolo,
  author       = {Redmon, Joseph and Divvala, Santosh and Girshick, Ross and Farhadi, Ali},
  title        = {You Only Look Once: Unified, Real-Time Object Detection},
  booktitle    = {2016 IEEE Conference on Computer Vision and Pattern Recognition (CVPR)},
  year         = {2016},
  pages        = {779--788},
  doi          = {10.1109/CVPR.2016.91},
  url          = {https://doi.org/10.1109/CVPR.2016.91}
}

@inproceedings{reed2023scalemae,
  author       = {Reed, Colorado J and Gupta, Ritwik and Li, Shufan and Brockman, Sarah and Funk, Christopher and Clipp, Brian and Keutzer, Kurt and Candido, Salvatore and Uyttendaele, Matt and Darrell, Trevor},
  title        = {Scale-MAE: A Scale-Aware Masked Autoencoder for Multiscale Geospatial Representation Learning},
  booktitle    = {2023 IEEE/CVF International Conference on Computer Vision (ICCV)},
  year         = {2023},
  pages        = {4088--4099},
  doi          = {10.1109/iccv51070.2023.00378},
  url          = {https://doi.org/10.1109/iccv51070.2023.00378}
}

@article{ren2017fasterrcnn,
  author       = {Ren, Shaoqing and He, Kaiming and Girshick, Ross and Sun, Jian},
  title        = {Faster R-CNN: Towards Real-Time Object Detection with Region Proposal Networks},
  journal      = {IEEE Transactions on Pattern Analysis and Machine Intelligence},
  year         = {2017},
  volume       = {39},
  number       = {6},
  pages        = {1137--1149},
  doi          = {10.1109/TPAMI.2016.2577031},
  url          = {https://doi.org/10.1109/TPAMI.2016.2577031}
}

@inproceedings{ronneberger2015unet,
  author       = {Ronneberger, Olaf and Fischer, Philipp and Brox, Thomas},
  title        = {U-Net: Convolutional Networks for Biomedical Image Segmentation},
  booktitle    = {Lecture Notes in Computer Science},
  year         = {2015},
  pages        = {234--241},
  doi          = {10.1007/978-3-319-24574-4_28},
  url          = {https://doi.org/10.1007/978-3-319-24574-4_28}
}

@inproceedings{rublee2011orb,
  author       = {Rublee, Ethan and Rabaud, Vincent and Konolige, Kurt and Bradski, Gary},
  title        = {ORB: An efficient alternative to SIFT or SURF},
  booktitle    = {2011 International Conference on Computer Vision},
  year         = {2011},
  pages        = {2564--2571},
  doi          = {10.1109/ICCV.2011.6126544},
  url          = {https://doi.org/10.1109/ICCV.2011.6126544}
}

@inproceedings{sarlin2020superglue,
  author       = {Sarlin, Paul-Edouard and DeTone, Daniel and Malisiewicz, Tomasz and Rabinovich, Andrew},
  title        = {SuperGlue: Learning Feature Matching With Graph Neural Networks},
  booktitle    = {2020 IEEE/CVF Conference on Computer Vision and Pattern Recognition (CVPR)},
  year         = {2020},
  pages        = {4938--4947},
  doi          = {10.1109/CVPR42600.2020.00499},
  url          = {https://doi.org/10.1109/CVPR42600.2020.00499}
}

@article{shen2024icafusion,
  author       = {Shen, Jifeng and Chen, Yifei and Liu, Yue and Zuo, Xin and Fan, Heng and Yang, Wankou},
  title        = {ICAFusion: Iterative cross-attention guided feature fusion for multispectral object detection},
  journal      = {Pattern Recognition},
  year         = {2024},
  volume       = {145},
  pages        = {109913},
  doi          = {10.1016/j.patcog.2023.109913},
  url          = {https://doi.org/10.1016/j.patcog.2023.109913}
}

@article{shi2021cdd,
  author       = {Shi, Qian and Liu, Mengxi and Li, Shengchen and Liu, Xiaoping and Wang, Fei and Zhang, Liangpei},
  title        = {A Deeply Supervised Attention Metric-Based Network and an Open Aerial Image Dataset for Remote Sensing Change Detection},
  journal      = {IEEE Transactions on Geoscience and Remote Sensing},
  year         = {2022},
  volume       = {60},
  pages        = {1--16},
  doi          = {10.1109/tgrs.2021.3085870},
  url          = {https://doi.org/10.1109/tgrs.2021.3085870}
}

@inproceedings{song2021p2pnet,
  author       = {Song, Qingyu and Wang, Changan and Jiang, Zhengkai and Wang, Yabiao and Tai, Ying and Wang, Chengjie and Li, Jilin and Huang, Feiyue and Wu, Yang},
  title        = {Rethinking Counting and Localization in Crowds: A Purely Point-Based Framework},
  booktitle    = {2021 IEEE/CVF International Conference on Computer Vision (ICCV)},
  year         = {2021},
  pages        = {3365--3374},
  doi          = {10.1109/ICCV48922.2021.00335},
  url          = {https://doi.org/10.1109/ICCV48922.2021.00335}
}

@inproceedings{strudel2021segmenter,
  author       = {Strudel, Robin and Garcia, Ricardo and Laptev, Ivan and Schmid, Cordelia},
  title        = {Segmenter: Transformer for Semantic Segmentation},
  booktitle    = {2021 IEEE/CVF International Conference on Computer Vision (ICCV)},
  year         = {2021},
  pages        = {7262--7272},
  doi          = {10.1109/ICCV48922.2021.00717},
  url          = {https://doi.org/10.1109/ICCV48922.2021.00717}
}

@article{sun2019rtfnet,
  author       = {Sun, Yuxiang and Zuo, Weixun and Liu, Ming},
  title        = {RTFNet: RGB-Thermal Fusion Network for Semantic Segmentation of Urban Scenes},
  journal      = {IEEE Robotics and Automation Letters},
  year         = {2019},
  volume       = {4},
  number       = {3},
  pages        = {2576--2583},
  doi          = {10.1109/LRA.2019.2904733},
  url          = {https://doi.org/10.1109/LRA.2019.2904733}
}

@article{sun2020dronevehicle,
  author       = {Sun, Yiming and Cao, Bing and Zhu, Pengfei and Hu, Qinghua},
  title        = {Drone-Based RGB-Infrared Cross-Modality Vehicle Detection Via Uncertainty-Aware Learning},
  journal      = {IEEE Transactions on Circuits and Systems for Video Technology},
  year         = {2022},
  volume       = {32},
  number       = {10},
  pages        = {6700--6713},
  doi          = {10.1109/TCSVT.2022.3168279},
  url          = {https://doi.org/10.1109/TCSVT.2022.3168279}
}

@inproceedings{sun2021loftr,
  author       = {Sun, Jiaming and Shen, Zehong and Wang, Yuang and Bao, Hujun and Zhou, Xiaowei},
  title        = {LoFTR: Detector-Free Local Feature Matching with Transformers},
  booktitle    = {2021 IEEE/CVF Conference on Computer Vision and Pattern Recognition (CVPR)},
  year         = {2021},
  pages        = {8922--8931},
  doi          = {10.1109/cvpr46437.2021.00881},
  url          = {https://doi.org/10.1109/cvpr46437.2021.00881}
}

@article{sun2023ringmo,
  author       = {Sun, Xian and Wang, Peijin and Lu, Wanxuan and Zhu, Zicong and Lu, Xiaonan and He, Qibin and Li, Junxi and Rong, Xuee and Yang, Zhujun and Chang, Hao and He, Qinglin and Yang, Guang and Wang, Ruiping and Lu, Jiwen and Fu, Kun},
  title        = {RingMo: A Remote Sensing Foundation Model With Masked Image Modeling},
  journal      = {IEEE Transactions on Geoscience and Remote Sensing},
  year         = {2023},
  volume       = {61},
  pages        = {1--22},
  doi          = {10.1109/TGRS.2022.3194732},
  url          = {https://doi.org/10.1109/TGRS.2022.3194732}
}

@article{suo2023hituav,
  author       = {Suo, Jiashun and Wang, Tianyi and Zhang, Xingzhou and Chen, Haiyang and Zhou, Wei and Shi, Weisong},
  title        = {HIT-UAV: A high-altitude infrared thermal dataset for Unmanned Aerial Vehicle-based object detection},
  journal      = {Scientific Data},
  year         = {2023},
  volume       = {10},
  number       = {1},
  pages        = {227},
  doi          = {10.1038/s41597-023-02066-6},
  url          = {https://doi.org/10.1038/s41597-023-02066-6}
}

@article{szwarcman2025prithvi,
  author       = {Szwarcman, Daniela and Roy, Sujit and Fraccaro, Paolo and G{\'{i}}slason, {\TH}orsteinn El{\'{i}} and Blumenstiel, Benedikt and Ghosal, Rinki and de Oliveira, Pedro Henrique and de Sousa Almeida, Jo{\~a}o Lucas and Sedona, Rocco and Kang, Yanghui and Chakraborty, Srija and Wang, Sizhe and Gomes, Carlos and Kumar, Ankur and Gaur, Vishal and Truong, Myscon and Godwin, Denys and Khallaghi, Sam and Lee, Hyunho and Hsu, Chia-Yu and Asanjan, Ata Akbari and Mujeci, Besart and Shidham, Disha and Balogun, Rufai Omowunmi and Kolluru, Venkatesh and Keenan, Trevor and Arevalo, Paulo and Li, Wenwen and Alemohammad, Hamed and Olofsson, Pontus and Mayer, Timothy and Hain, Christopher and Kennedy, Robert and Zadrozny, Bianca and Bell, David and Cavallaro, Gabriele and Watson, Campbell and Maskey, Manil and Ramachandran, Rahul and Moreno, Juan Bernabe},
  title        = {Prithvi-EO-2.0: A Versatile Multitemporal Foundation Model for Earth Observation Applications},
  journal      = {IEEE Transactions on Geoscience and Remote Sensing},
  year         = {2026},
  volume       = {64},
  pages        = {1--20},
  doi          = {10.1109/tgrs.2025.3642610},
  url          = {https://doi.org/10.1109/tgrs.2025.3642610}
}

@inproceedings{tan2021efficientnetv2,
  author       = {Tan, Mingxing and Le, Quoc V.},
  title        = {EfficientNetV2: Smaller Models and Faster Training},
  booktitle    = {Proceedings of the 38th International Conference on Machine Learning},
  year         = {2021},
  volume       = {139},
  pages        = {10096--10106},
  eprint       = {2104.00298},
  archivePrefix= {arXiv},
  url          = {https://arxiv.org/abs/2104.00298}
}

@article{tang2023divfusion,
  author       = {Tang, Linfeng and Xiang, Xinyu and Zhang, Hao and Gong, Meiqi and Ma, Jiayi},
  title        = {DIVFusion: Darkness-free infrared and visible image fusion},
  journal      = {Information Fusion},
  year         = {2023},
  volume       = {91},
  pages        = {477--493},
  doi          = {10.1016/j.inffus.2022.10.034},
  url          = {https://doi.org/10.1016/j.inffus.2022.10.034}
}

@inproceedings{teed2020raft,
  author       = {Teed, Zachary and Deng, Jia},
  title        = {RAFT: Recurrent All-Pairs Field Transforms for Optical Flow},
  booktitle    = {Lecture Notes in Computer Science},
  year         = {2020},
  pages        = {402--419},
  doi          = {10.1007/978-3-030-58536-5_24},
  url          = {https://doi.org/10.1007/978-3-030-58536-5_24}
}

@inproceedings{tian2019fcos,
  author       = {Tian, Zhi and Shen, Chunhua and Chen, Hao and He, Tong},
  title        = {FCOS: Fully Convolutional One-Stage Object Detection},
  booktitle    = {2019 IEEE/CVF International Conference on Computer Vision (ICCV)},
  year         = {2019},
  pages        = {9627--9636},
  doi          = {10.1109/ICCV.2019.00972},
  url          = {https://doi.org/10.1109/ICCV.2019.00972}
}

@misc{vaswani2017transformer,
  author       = {Vaswani, Ashish and Shazeer, Noam and Parmar, Niki and Uszkoreit, Jakob and Jones, Llion and Gomez, Aidan N. and Kaiser, Lukasz and Polosukhin, Illia},
  title        = {Attention Is All You Need},
  year         = {2017},
  eprint       = {1706.03762},
  archivePrefix= {arXiv},
  url          = {https://arxiv.org/abs/1706.03762}
}

@inproceedings{wan2025sigma,
  author       = {Wan, Zifu and Zhang, Pingping and Wang, Yuhao and Yong, Silong and Stepputtis, Simon and Sycara, Katia and Xie, Yaqi},
  title        = {Sigma: Siamese Mamba Network for Multi-Modal Semantic Segmentation},
  booktitle    = {2025 IEEE/CVF Winter Conference on Applications of Computer Vision (WACV)},
  year         = {2025},
  pages        = {1734--1744},
  doi          = {10.1109/WACV61041.2025.00176},
  url          = {https://doi.org/10.1109/WACV61041.2025.00176}
}

@inproceedings{wang2020dmcount,
  author       = {Wang, Boyu and Liu, Huidong and Samaras, Dimitris and Nguyen, Minh Hoai},
  title        = {Distribution Matching for Crowd Counting},
  booktitle    = {Advances in Neural Information Processing Systems},
  year         = {2020},
  volume       = {33},
  pages        = {1595--1607},
  eprint       = {2009.13077},
  archivePrefix= {arXiv},
  url          = {https://arxiv.org/abs/2009.13077}
}

@inproceedings{wang2022tokenfusion,
  author       = {Wang, Yikai and Chen, Xinghao and Cao, Lele and Huang, Wenbing and Sun, Fuchun and Wang, Yunhe},
  title        = {Multimodal Token Fusion for Vision Transformers},
  booktitle    = {2022 IEEE/CVF Conference on Computer Vision and Pattern Recognition (CVPR)},
  year         = {2022},
  pages        = {12186--12195},
  doi          = {10.1109/CVPR52688.2022.01187},
  url          = {https://doi.org/10.1109/CVPR52688.2022.01187}
}

@article{wang2022unetformer,
  author       = {Wang, Libo and Li, Rui and Zhang, Ce and Fang, Shenghui and Duan, Chenxi and Meng, Xiaoliang and Atkinson, Peter M.},
  title        = {UNetFormer: A UNet-like transformer for efficient semantic segmentation of remote sensing urban scene imagery},
  journal      = {ISPRS Journal of Photogrammetry and Remote Sensing},
  year         = {2022},
  volume       = {190},
  pages        = {196--214},
  doi          = {10.1016/j.isprsjprs.2022.06.008},
  url          = {https://doi.org/10.1016/j.isprsjprs.2022.06.008}
}

@article{wang2023rvsa,
  author       = {Wang, Di and Zhang, Qiming and Xu, Yufei and Zhang, Jing and Du, Bo and Tao, Dacheng and Zhang, Liangpei},
  title        = {Advancing Plain Vision Transformer Toward Remote Sensing Foundation Model},
  journal      = {IEEE Transactions on Geoscience and Remote Sensing},
  year         = {2023},
  volume       = {61},
  pages        = {1--15},
  doi          = {10.1109/TGRS.2022.3222818},
  url          = {https://doi.org/10.1109/TGRS.2022.3222818}
}

@article{wang2023sgfnet,
  author       = {Wang, Yike and Li, Gongyang and Liu, Zhi},
  title        = {SGFNet: Semantic-Guided Fusion Network for RGB-Thermal Semantic Segmentation},
  journal      = {IEEE Transactions on Circuits and Systems for Video Technology},
  year         = {2023},
  volume       = {33},
  number       = {12},
  pages        = {7737--7748},
  doi          = {10.1109/TCSVT.2023.3281419},
  url          = {https://doi.org/10.1109/TCSVT.2023.3281419}
}

@inproceedings{wang2023yolov7,
  author       = {Wang, Chien-Yao and Bochkovskiy, Alexey and Liao, Hong-Yuan Mark},
  title        = {YOLOv7: Trainable Bag-of-Freebies Sets New State-of-the-Art for Real-Time Object Detectors},
  booktitle    = {2023 IEEE/CVF Conference on Computer Vision and Pattern Recognition (CVPR)},
  year         = {2023},
  pages        = {7464--7475},
  doi          = {10.1109/CVPR52729.2023.00721},
  url          = {https://doi.org/10.1109/CVPR52729.2023.00721}
}

@article{wang2024mtp,
  author       = {Wang, Di and Zhang, Jing and Xu, Minqiang and Liu, Lin and Wang, Dongsheng and Gao, Erzhong and Han, Chengxi and Guo, Haonan and Du, Bo and Tao, Dacheng and Zhang, Liangpei},
  title        = {MTP: Advancing Remote Sensing Foundation Model via Multitask Pretraining},
  journal      = {IEEE Journal of Selected Topics in Applied Earth Observations and Remote Sensing},
  year         = {2024},
  volume       = {17},
  pages        = {11632--11654},
  doi          = {10.1109/jstars.2024.3408154},
  url          = {https://doi.org/10.1109/jstars.2024.3408154}
}

@article{wang2025ringmogalaxy,
  author       = {Wang, Zhechao and Wang, Zhirui and Cheng, Peirui and Zhao, Liangjin and Tian, Pengju and Wang, Yuchao and Chen, Mingxin and Li, Xinming and Sun, Xian},
  title        = {RingMo-Galaxy: A Remote Sensing Distributed Foundation Model for Diverse Downstream Tasks},
  journal      = {IEEE Transactions on Geoscience and Remote Sensing},
  year         = {2025},
  volume       = {63},
  pages        = {1--18},
  doi          = {10.1109/TGRS.2024.3517878},
  url          = {https://doi.org/10.1109/TGRS.2024.3517878}
}

@inproceedings{wojke2017deepsort,
  author       = {Wojke, Nicolai and Bewley, Alex and Paulus, Dietrich},
  title        = {Simple online and realtime tracking with a deep association metric},
  booktitle    = {2017 IEEE International Conference on Image Processing (ICIP)},
  year         = {2017},
  pages        = {3645--3649},
  doi          = {10.1109/ICIP.2017.8296962},
  url          = {https://doi.org/10.1109/ICIP.2017.8296962}
}

@article{wu2025skysensepp,
  author       = {Wu, Kang and Zhang, Yingying and Ru, Lixiang and Dang, Bo and Lao, Jiangwei and Yu, Lei and Luo, Junwei and Zhu, Zifan and Sun, Yue and Zhang, Jiahao and Zhu, Qi and Wang, Jian and Yang, Ming and Chen, Jingdong and Zhang, Yongjun and Li, Yansheng},
  title        = {A semantic-enhanced multi-modal remote sensing foundation model for Earth observation},
  journal      = {Nature Machine Intelligence},
  year         = {2025},
  volume       = {7},
  number       = {8},
  pages        = {1235--1249},
  doi          = {10.1038/s42256-025-01078-8},
  url          = {https://doi.org/10.1038/s42256-025-01078-8}
}

@article{xia2017aid,
  author       = {Xia, Gui-Song and Hu, Jingwen and Hu, Fan and Shi, Baoguang and Bai, Xiang and Zhong, Yanfei and Zhang, Liangpei and Lu, Xiaoqiang},
  title        = {AID: A Benchmark Data Set for Performance Evaluation of Aerial Scene Classification},
  journal      = {IEEE Transactions on Geoscience and Remote Sensing},
  year         = {2017},
  volume       = {55},
  number       = {7},
  pages        = {3965--3981},
  doi          = {10.1109/TGRS.2017.2685945},
  url          = {https://doi.org/10.1109/TGRS.2017.2685945}
}

@inproceedings{xiao2018upernet,
  author       = {Xiao, Tete and Liu, Yingcheng and Zhou, Bolei and Jiang, Yuning and Sun, Jian},
  title        = {Unified Perceptual Parsing for Scene Understanding},
  booktitle    = {Lecture Notes in Computer Science},
  year         = {2018},
  pages        = {432--448},
  doi          = {10.1007/978-3-030-01228-1_26},
  url          = {https://doi.org/10.1007/978-3-030-01228-1_26}
}

@inproceedings{xie2021orientedrcnn,
  author       = {Xie, Xingxing and Cheng, Gong and Wang, Jiabao and Yao, Xiwen and Han, Junwei},
  title        = {Oriented R-CNN for Object Detection},
  booktitle    = {2021 IEEE/CVF International Conference on Computer Vision (ICCV)},
  year         = {2021},
  pages        = {3520--3529},
  doi          = {10.1109/ICCV48922.2021.00350},
  url          = {https://doi.org/10.1109/ICCV48922.2021.00350}
}

@misc{xie2021segformer,
  author       = {Xie, Enze and Wang, Wenhai and Yu, Zhiding and Anandkumar, Anima and Alvarez, Jose M. and Luo, Ping},
  title        = {SegFormer: Simple and Efficient Design for Semantic Segmentation with Transformers},
  year         = {2021},
  eprint       = {2105.15203},
  archivePrefix= {arXiv},
  url          = {https://arxiv.org/abs/2105.15203}
}

@misc{xiong2024dofa,
  author       = {Xiong, Zhitong and Wang, Yi and Zhang, Fahong and Stewart, Adam J. and Hanna, Jo{\"{e}}lle and Borth, Damian and Papoutsis, Ioannis and Saux, Bertrand Le and Camps-Valls, Gustau and Zhu, Xiao Xiang},
  title        = {Neural Plasticity-Inspired Multimodal Foundation Model for Earth Observation},
  year         = {2024},
  eprint       = {2403.15356},
  archivePrefix= {arXiv},
  url          = {https://arxiv.org/abs/2403.15356}
}

@inproceedings{xu2023pidnet,
  author       = {Xu, Jiacong and Xiong, Zixiang and Bhattacharyya, Shankar P.},
  title        = {PIDNet: A Real-time Semantic Segmentation Network Inspired by PID Controllers},
  booktitle    = {2023 IEEE/CVF Conference on Computer Vision and Pattern Recognition (CVPR)},
  year         = {2023},
  pages        = {19529--19539},
  doi          = {10.1109/cvpr52729.2023.01871},
  url          = {https://doi.org/10.1109/cvpr52729.2023.01871}
}

@misc{yang2026geometricgating,
  author       = {Yang, Jingpu and Ji, Fengxian and Lai, Zhengzhao and Wu, Juanfan and Cui, Mingxuan and Wang, Yufeng},
  title        = {Zero-Parameter Geometric Gating for Temporally Stable Low-Altitude UAV Video Semantic Segmentation},
  year         = {2026},
  eprint       = {2606.09162},
  archivePrefix= {arXiv},
  url          = {https://arxiv.org/abs/2606.09162}
}

@inproceedings{yang2026mars,
  author       = {Yang, Ruoyu and Liu, Yinhe and Yan, Heng and Zhou, Yiheng and Fu, Yihan and Luo, Han and Zhong, Yanfei},
  title        = {MaRS: A Multi-modality Very-high-resolution Remote Sensing Foundation Model with Cross-Granularity Meta-Modality Learning},
  booktitle    = {Proceedings of the AAAI Conference on Artificial Intelligence},
  year         = {2026},
  volume       = {40},
  number       = {14},
  pages        = {11685--11693},
  doi          = {10.1609/aaai.v40i14.38153},
  url          = {https://doi.org/10.1609/aaai.v40i14.38153}
}

@article{ying2025visiblethermal,
  author       = {Ying, Xinyi and Xiao, Chao and An, Wei and Li, Ruojing and He, Xu and Li, Boyang and Cao, Xu and Li, Zhaoxu and Wang, Yingqian and Hu, Mingyuan and Xu, Qingyu and Lin, Zaiping and Li, Miao and Zhou, Shilin and Liu, Li and Sheng, Weidong},
  title        = {Visible-Thermal Tiny Object Detection: A Benchmark Dataset and Baselines},
  journal      = {IEEE Transactions on Pattern Analysis and Machine Intelligence},
  year         = {2025},
  volume       = {47},
  number       = {7},
  pages        = {6088--6096},
  doi          = {10.1109/TPAMI.2025.3544621},
  url          = {https://doi.org/10.1109/TPAMI.2025.3544621}
}

@inproceedings{yu2018dla,
  author       = {Yu, Fisher and Wang, Dequan and Shelhamer, Evan and Darrell, Trevor},
  title        = {Deep Layer Aggregation},
  booktitle    = {2018 IEEE/CVF Conference on Computer Vision and Pattern Recognition},
  year         = {2018},
  pages        = {2403--2412},
  doi          = {10.1109/CVPR.2018.00255},
  url          = {https://doi.org/10.1109/CVPR.2018.00255}
}

@article{yu2021bisenetv2,
  author       = {Yu, Changqian and Gao, Changxin and Wang, Jingbo and Yu, Gang and Shen, Chunhua and Sang, Nong},
  title        = {BiSeNet V2: Bilateral Network with Guided Aggregation for Real-Time Semantic Segmentation},
  journal      = {International Journal of Computer Vision},
  year         = {2021},
  volume       = {129},
  number       = {11},
  pages        = {3051--3068},
  doi          = {10.1007/s11263-021-01515-2},
  url          = {https://doi.org/10.1007/s11263-021-01515-2}
}

@inproceedings{yuan2020ocrnet,
  author       = {Yuan, Yuhui and Chen, Xilin and Wang, Jingdong},
  title        = {Object-Contextual Representations for Semantic Segmentation},
  booktitle    = {Lecture Notes in Computer Science},
  year         = {2020},
  pages        = {173--190},
  doi          = {10.1007/978-3-030-58539-6_11},
  url          = {https://doi.org/10.1007/978-3-030-58539-6_11}
}

@inproceedings{zhang2018lpips,
  author       = {Zhang, Richard and Isola, Phillip and Efros, Alexei A. and Shechtman, Eli and Wang, Oliver},
  title        = {The Unreasonable Effectiveness of Deep Features as a Perceptual Metric},
  booktitle    = {2018 IEEE/CVF Conference on Computer Vision and Pattern Recognition},
  year         = {2018},
  pages        = {586--595},
  doi          = {10.1109/CVPR.2018.00068},
  url          = {https://doi.org/10.1109/CVPR.2018.00068}
}

@inproceedings{zhang2021bytetrack,
  author       = {Zhang, Yifu and Sun, Peize and Jiang, Yi and Yu, Dongdong and Weng, Fucheng and Yuan, Zehuan and Luo, Ping and Liu, Wenyu and Wang, Xinggang},
  title        = {ByteTrack: Multi-object Tracking by Associating Every Detection Box},
  booktitle    = {Lecture Notes in Computer Science},
  year         = {2022},
  pages        = {1--21},
  doi          = {10.1007/978-3-031-20047-2_1},
  url          = {https://doi.org/10.1007/978-3-031-20047-2_1}
}

@article{zhang2021fairmot,
  author       = {Zhang, Yifu and Wang, Chunyu and Wang, Xinggang and Zeng, Wenjun and Liu, Wenyu},
  title        = {FairMOT: On the Fairness of Detection and Re-identification in Multiple Object Tracking},
  journal      = {International Journal of Computer Vision},
  year         = {2021},
  volume       = {129},
  number       = {11},
  pages        = {3069--3087},
  doi          = {10.1007/s11263-021-01513-4},
  url          = {https://doi.org/10.1007/s11263-021-01513-4}
}

@misc{zhang2022dino,
  author       = {Zhang, Hao and Li, Feng and Liu, Shilong and Zhang, Lei and Su, Hang and Zhu, Jun and Ni, Lionel M. and Shum, Heung-Yeung},
  title        = {DINO: DETR with Improved DeNoising Anchor Boxes for End-to-End Object Detection},
  year         = {2022},
  eprint       = {2203.03605},
  archivePrefix= {arXiv},
  url          = {https://arxiv.org/abs/2203.03605}
}

@inproceedings{zhang2022vtuav,
  author       = {Zhang, Pengyu and Zhao, Jie and Wang, Dong and Lu, Huchuan and Ruan, Xiang},
  title        = {Visible-Thermal UAV Tracking: A Large-Scale Benchmark and New Baseline},
  booktitle    = {Proceedings of the IEEE/CVF Conference on Computer Vision and Pattern Recognition (CVPR)},
  year         = {2022},
  pages        = {8886--8895},
  eprint       = {2204.04120},
  archivePrefix= {arXiv},
  url          = {https://arxiv.org/abs/2204.04120}
}

@article{zhang2023cmx,
  author       = {Zhang, Jiaming and Liu, Huayao and Yang, Kailun and Hu, Xinxin and Liu, Ruiping and Stiefelhagen, Rainer},
  title        = {CMX: Cross-Modal Fusion for RGB-X Semantic Segmentation With Transformers},
  journal      = {IEEE Transactions on Intelligent Transportation Systems},
  year         = {2023},
  volume       = {24},
  number       = {12},
  pages        = {14679--14694},
  doi          = {10.1109/tits.2023.3300537},
  url          = {https://doi.org/10.1109/tits.2023.3300537}
}

@inproceedings{zhang2023ddq,
  author       = {Zhang, Shilong and Wang, Xinjiang and Wang, Jiaqi and Pang, Jiangmiao and Lyu, Chengqi and Zhang, Wenwei and Luo, Ping and Chen, Kai},
  title        = {Dense Distinct Query for End-to-End Object Detection},
  booktitle    = {2023 IEEE/CVF Conference on Computer Vision and Pattern Recognition (CVPR)},
  year         = {2023},
  pages        = {7329--7338},
  doi          = {10.1109/cvpr52729.2023.00708},
  url          = {https://doi.org/10.1109/cvpr52729.2023.00708}
}

@inproceedings{zhang2025skysensev2,
  author       = {Zhang, Yingying and Ru, Lixiang and Wu, Kang and Yu, Lei and Liang, Lei and Li, Yansheng and Chen, Jingdong},
  title        = {SkySense V2: A Unified Foundation Model for Multi-modal Remote Sensing},
  booktitle    = {Proceedings of the IEEE/CVF International Conference on Computer Vision (ICCV)},
  year         = {2025},
  pages        = {9136--9146},
  doi          = {10.1109/ICCV51701.2025.00854},
  eprint       = {2507.13812},
  archivePrefix= {arXiv},
  url          = {https://doi.org/10.1109/ICCV51701.2025.00854}
}

@inproceedings{zhao2017pspnet,
  author       = {Zhao, Hengshuang and Shi, Jianping and Qi, Xiaojuan and Wang, Xiaogang and Jia, Jiaya},
  title        = {Pyramid Scene Parsing Network},
  booktitle    = {2017 IEEE Conference on Computer Vision and Pattern Recognition (CVPR)},
  year         = {2017},
  pages        = {2881--2890},
  doi          = {10.1109/CVPR.2017.660},
  url          = {https://doi.org/10.1109/CVPR.2017.660}
}

@inproceedings{zhao2024rtdetr,
  author       = {Zhao, Yian and Lv, Wenyu and Xu, Shangliang and Wei, Jinman and Wang, Guanzhong and Dang, Qingqing and Liu, Yi and Chen, Jie},
  title        = {DETRs Beat YOLOs on Real-time Object Detection},
  booktitle    = {2024 IEEE/CVF Conference on Computer Vision and Pattern Recognition (CVPR)},
  year         = {2024},
  pages        = {16965--16974},
  doi          = {10.1109/CVPR52733.2024.01605},
  url          = {https://doi.org/10.1109/CVPR52733.2024.01605}
}

@inproceedings{zheng2021setr,
  author       = {Zheng, Sixiao and Lu, Jiachen and Zhao, Hengshuang and Zhu, Xiatian and Luo, Zekun and Wang, Yabiao and Fu, Yanwei and Feng, Jianfeng and Xiang, Tao and Torr, Philip H.S. and Zhang, Li},
  title        = {Rethinking Semantic Segmentation from a Sequence-to-Sequence Perspective with Transformers},
  booktitle    = {2021 IEEE/CVF Conference on Computer Vision and Pattern Recognition (CVPR)},
  year         = {2021},
  pages        = {6881--6890},
  doi          = {10.1109/CVPR46437.2021.00681},
  url          = {https://doi.org/10.1109/CVPR46437.2021.00681}
}

@inproceedings{zhou2020centertrack,
  author       = {Zhou, Xingyi and Koltun, Vladlen and Kr{\"{a}}henb{\"{u}}hl, Philipp},
  title        = {Tracking Objects as Points},
  booktitle    = {Lecture Notes in Computer Science},
  year         = {2020},
  pages        = {474--490},
  doi          = {10.1007/978-3-030-58548-8_28},
  url          = {https://doi.org/10.1007/978-3-030-58548-8_28}
}

@misc{zhu2020deformabledetr,
  author       = {Zhu, Xizhou and Su, Weijie and Lu, Lewei and Li, Bin and Wang, Xiaogang and Dai, Jifeng},
  title        = {Deformable DETR: Deformable Transformers for End-to-End Object Detection},
  year         = {2020},
  eprint       = {2010.04159},
  archivePrefix= {arXiv},
  url          = {https://arxiv.org/abs/2010.04159}
}

@article{zhu2021visdrone,
  author       = {Zhu, Pengfei and Wen, Longyin and Du, Dawei and Bian, Xiao and Fan, Heng and Hu, Qinghua and Ling, Haibin},
  title        = {Detection and Tracking Meet Drones Challenge},
  journal      = {IEEE Transactions on Pattern Analysis and Machine Intelligence},
  year         = {2022},
  volume       = {44},
  number       = {11},
  pages        = {7380--7399},
  doi          = {10.1109/tpami.2021.3119563},
  url          = {https://doi.org/10.1109/tpami.2021.3119563}
}

@inproceedings{zhu2025wavemamba,
  author       = {Zhu, Haodong and Dong, Wenhao and Yang, Linlin and Li, Hong and Yang, Yuguang and Ren, Yangyang and Zhu, Qingcheng and Feng, Zichao and Li, Changbai and Lin, Shaohui and Wang, Runqi and Luo, Xiaoyan and Zhang, Baochang},
  title        = {WaveMamba: Wavelet-Driven Mamba Fusion for RGB-Infrared Object Detection},
  booktitle    = {Proceedings of the IEEE/CVF International Conference on Computer Vision (ICCV)},
  year         = {2025},
  pages        = {11219--11229},
  doi          = {10.1109/ICCV51701.2025.01044},
  eprint       = {2507.18173},
  archivePrefix= {arXiv},
  url          = {https://doi.org/10.1109/ICCV51701.2025.01044}
}

\end{document}